# Crash Severity Risk Modeling Strategies under Data Imbalance

**Abdullah Al Mamun, Ph.D.**
Glenn Department of Civil Engineering
Clemson University, Clemson, South Carolina, 29634
Email: abdullm@clemson.edu

**Abyad Enan**
Glenn Department of Civil Engineering
Clemson University, Clemson, South Carolina, 29634
Email: aenan@clemson.edu

**Debbie Aisiana Indah**
Department of Engineering
South Carolina State University, Orangeburg, South Carolina, 29117
Email: dindah@scsu.edu

**Judith Mwakalonge, Ph.D.**
Professor
Department of Engineering
South Carolina State University, Orangeburg, South Carolina, 29117
Email: jmwakalo@scsu.edu

**Gurcan Comert, Ph.D.**
Associate Professor
Computational Data Science and Engineering Department
North Carolina A&T State University, Greensboro, North Carolina, 27411
Email: gcomert@ncat.edu

**Mashrur Chowdhury, Ph.D.**
Professor
Glenn Department of Civil Engineering
Clemson University, Clemson, South Carolina, 29634
Email: mac@clemson.edu

**ABSTRACT**

This study investigates crash severity risk modeling strategies for work zones involving large vehicles (i.e., trucks, buses, and vans) under crash data imbalance between low-severity (LS) and high-severity (HS) crashes. We utilized crash data involving large vehicles in South Carolina work zones from 2014 to 2018, which included four times more LS crashes than HS crashes. The objective of this study is to evaluate the crash severity prediction performance of various statistical, machine learning, and deep learning models under different feature selection and data balancing techniques. Findings highlight a disparity in LS and HS predictions, with lower accuracy for HS crashes due to class imbalance and feature overlap. Discriminative Mutual Information (DMI) yields the most effective feature set for predicting HS crashes without requiring data balancing, particularly when paired with gradient boosting models and deep neural networks such as CatBoost, NeuralNetTorch, XGBoost, and LightGBM. Data balancing techniques such as NearMiss-1 maximize HS recall when combined with DMI-selected features and certain models such as LightGBM, making them well-suited for HS crash prediction. Conversely, RandomUnderSampler, HS Class Weighting, and RandomOverSampler achieve more balanced performance, which is defined as an equitable trade-off between LS and HS metrics, especially when applied to NeuralNetTorch, NeuralNetFastAI, CatBoost, LightGBM, and Bayesian Mixed Logit (BML) using merged feature sets or models without feature selection. The insights from this study offer safety analysts guidance on selecting models, feature selection, and data balancing techniques aligned with specific safety goals, providing a robust foundation for enhancing work-zone crash severity prediction.

## INTRODUCTION

Motor vehicle traffic crashes remain a significant concern in the United States (US), with a concerning number of fatalities and injuries occurring annually on US roads. Among these incidents, crashes involving large vehicles, such as commercial motor vehicles (CMVs), heavy trucks, buses, and other oversized vehicles, are particularly alarming due to their larger size and weight, which pose considerable risks to road users and workers in work zones. The unique challenges associated with large vehicles, such as navigating narrow lanes, managing stop-and-go traffic, and performing merging maneuvers, are intensified in work zones, where temporary traffic configurations and restricted spaces further limit their maneuverability. These factors collectively contribute to a higher incidence of crashes, injuries, and fatalities within work zones (*1*).

Data from the National Highway Traffic Safety Administration (NHTSA) Fatality Analysis Reporting System (FARS) indicates an alarming upward trend in fatal work-zone crashes nationwide (*2*). As depicted in **Figure 1**, work-zone fatalities reached 821 in 2022, a 53% increase from 536 in 2013. Notably, fatal work-zone crashes involving CMVs, representing large vehicles, also show a concerning pattern, with CMV-involved fatal crashes increasing by 62% from 153 in 2013 to 248 in 2022. In South Carolina (SC), 33% of fatal crashes in work zones over the period 2013-2022 involved CMVs (*2*). These trends may have been driven, in part, by the surge in truck traffic during the COVID-19 pandemic, which increased interactions between large vehicles and other road users in work zones. Studies show that while overall traffic volumes declined during the COVID-19 pandemic due to travel restrictions, truck traffic rebounded quickly and even surpassed pre-pandemic levels in some regions, driven by increased e-commerce demand (*3*). These data highlight the urgent need for effective safety interventions, particularly targeting incidents involving large vehicles in work zones, to enhance safety for vehicle occupants, workers, and other road users.

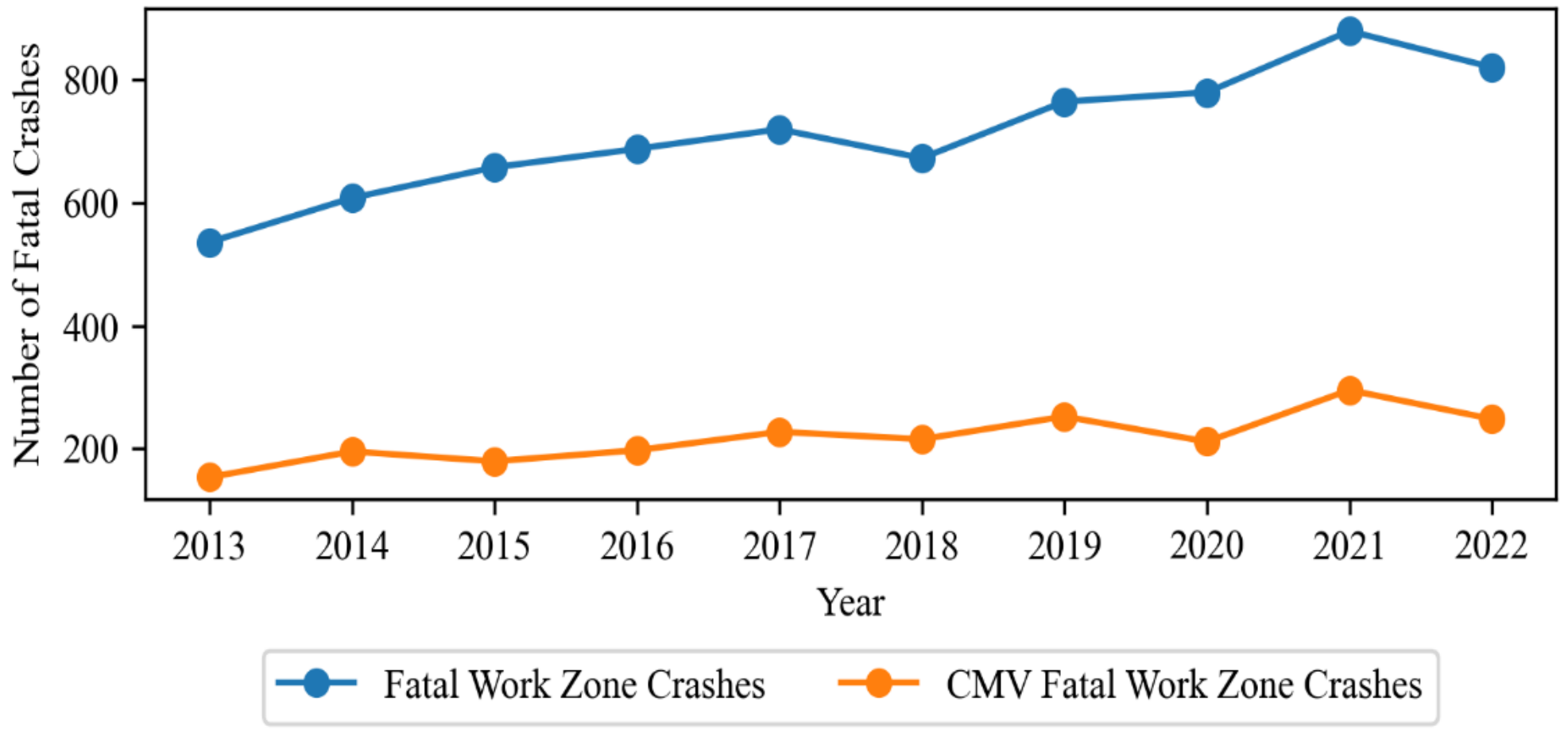

**Figure 1 Trend of US Nationwide Fatal Work-Zone Crashes and CMV-Involved Fatal Work-Zone Crashes (created from the data provided in (*2*)).**

Despite the significant risks posed by crashes involving large vehicles in work zones, there has been limited research focused on developing predictive (risk) models for crash severity specifically tailored to these incidents. This study addresses this critical gap by developing and evaluating risk models for crash severity in SC work zones, specifically focusing on large vehicles. Utilizing SC work-zone crash data from 2014 to 2018, obtained from police crash reports, this study leverages statistical, machine learning, and deep learning models to enhance our understanding and prevention of severe crashes within work zones.

Crash severity risk modeling enhances work zone safety by providing actionable insights for stakeholders. Transportation agencies can identify key factors contributing to severe crashes involving large vehicles in work zones, enabling targeted safety interventions such as stricter law enforcement and data-driven regulations such as adjusting speed limits or deploying additional warning signs during inclement weather (*4*). Vehicle manufacturers can leverage these models to develop advanced driver-assistance systems for large vehicles (*5*). Construction companies and public agencies can use model insights to optimize work zone planning and safety protocols to reduce the risk of crashes involving large vehicles. Additionally, predicting crash severity risk significantly improves emergency response efficiency by enabling the allocation of appropriate medical personnel and equipment at the scene, potentially saving lives (*6*).

Statistical, machine learning (ML), and deep learning (DL) models offer distinct advantages and disadvantages for crash severity prediction. While statistical models provide interpretable results, they require assumptions about data distribution and often rely on linear functions, potentially limiting their accuracy in capturing real-world complexities (*7*). In contrast, machine learning and deep learning models do not require pre-assumed relationships between variables. This flexibility often leads to better prediction performance, as these models can handle complex and non-linear interactions without requiring specific distributional assumptions. However, their strength in prediction comes at the cost of interpretability, making it more challenging to understand the underlying reasons behind their predictions (*8*). In recognition of these trade-offs, this study utilizes a range of statistical, machine learning, and deep learning models, with each modeling strategy contributing to its strengths.

While crash severity modeling offers actionable insights for stakeholders, predicting crash severity using real-world traffic crash data presents several challenges. One major challenge is class imbalance. Traffic crashes are dominated by non-severe events, leading to a significant disparity between the minority class (severe crashes) and the majority class (non-severe crashes). Standard classification algorithms often favor the majority class, potentially resulting in biased and inaccurate models that predict the most frequent outcome for the majority of the cases (*6*, *9*). To address the challenge of class imbalance in crash data, this study implements various data balancing techniques to create a more balanced class distribution, allowing the models to learn effectively from the less frequent yet critical severe crash data. Additionally, datasets based on police crash reports often contain numerous categorical features with potential redundancies. To improve model performance and reduce overfitting, this study employs several feature selection techniques to help identify and remove redundant features, leading to a more concise and informative feature set for crash severity prediction.

The remainder of the paper is organized as follows: The second section presents a literature review, the third section introduces the data and features utilized in this study, the fourth section describes the study's methodology, including feature selection and data balancing techniques, models, and performance metrics used to evaluate the models, the fifth section describes the analysis framework, the sixth section discusses the model performance results and provides an analysis of these results, and finally, the seventh section presents conclusions and recommendations for future study.

## LITERATURE REVIEW

This section provides a comprehensive review of crash severity prediction studies, focusing on crashes involving large vehicles, both within and outside of work zones. Among these, studies on CMV-involved crashes are particularly emphasized as they represent the largest subset of large vehicles. By examining the scope of these studies and evaluating their performance, this review highlights current knowledge gaps and presents the objectives of this study to bridge these gaps.

Khorashadi et al. analyzed CMV-involved crashes in California (1997-2000) using Multinomial Logit (MNL) models (*10*). They identified significant factors influencing crash severity, such as driving under the influence, which increased severe injury probabilities substantially more in urban areas (798%) than in rural areas (246%). Similarly, Hosseinpour et al. developed safety performance functions using zero-inflated heterogeneous Conway-Maxwell-Poisson models for Kentucky's CMV-involved crashes (2015-2019) (*11*). These models provided a better fit and accuracy compared to Negative Binomial models but focused on identifying significant predictors rather than developing predictive models. Pathivada et al. integrated crash data from the Kentucky Transportation Cabinet (KYTC) with real-time weather data for CMV-involved crashes on Interstate-65 in Kentucky (2016-2021), applying Association

Rules Mining and Correlated Mixed Logit with Heterogeneity in Means models (*12*). They found that low air temperatures and solar radiation significantly increased severe crash likelihood, emphasizing factor analysis over predictive modeling. Chen et al. used New Mexico crash data (2010-2011) to develop a Hierarchical Bayesian Random Intercept model, highlighting significant variables such as road grade, vehicle damage, and driver behavior responsible for CMV-involved crashes, without focusing on predictive capabilities (*13*).

In the context of CMV-involved work-zone crashes, Khattak and Targa used Ordered Probit models to analyze crash data from North Carolina (2000), identifying multivehicle, CMV-involved collisions as the most injurious (*1*). Osman et al. compared MNL, Nested Logit, and Generalized Ordered Logit (GORL) models using Minnesota crash data (2003-2012), finding the GORL model provided the best data fit with significant factors including rural principal arterials and adverse weather (*14*). Madarshahian et al. investigated CMV-involved work-zone crashes in SC (2014-2020) using Mixed Logit models for interstates and non-interstates, identifying factors such as dark lighting conditions, female at-fault drivers, and driving too fast for conditions as contributors to severe crashes (*15*).

While numerous studies have analyzed contributing factors to crash severity in crashes involving large vehicles, a critical gap exists in research focused on developing comprehensive risk models for crash severity. Existing studies often rely on statistical models for factor analysis, particularly regarding crashes involving large vehicles within work zones. However, while statistical models provide interpretable results, they often require assumptions about data distribution and rely on linear functions, which may limit their ability to capture the complex, non-linear relationship observed in the real-world crash data. As a result, their predictive power can sometimes be insufficient or unreliable for implementing proactive safety measures in the real-world. To bridge this gap, the authors of this study aim to develop and validate risk modelling framework for crash severity in incidents involving large vehicles within SC work zones. We employ a comprehensive methodological approach that incorporates different types of models, such as the statistical model Bayesian Mixed Logit (BML), and machine learning and deep learning models including Categorical Boosting (CatBoost), Extremely Randomized Trees (Extra Trees [ET]), Light Gradient Boosting Machine (LightGBM), Neural Network (NeuralNet), Random Forest (RF), and eXtreme Gradient Boosting (XGBoost). Additionally, we explore various data balancing techniques, including assigning weights to the minority class, oversampling of the minority class using the Synthetic Minority Oversampling Technique (SMOTE), KMeansSMOTE, Random Oversampling (RandomOverSampler), Adaptive Synthetic Sampling (ADASYN), Kernel-based SMOTE (K-SMOTE), Wasserstein Generative Adversarial Network with Gradient Penalty (WGAN-GP), and Conditional WGAN-GP, undersampling of the majority class using NearMiss-1, Random Undersampling (RandomUnderSampler), and a combination of RandomOverSampler and RandomUnderSampler, to address potential class imbalances in crash data. Furthermore, we incorporate advanced feature selection techniques such as Pearson’s correlation, Feature Importance (FI) using RF, Recursive Feature Elimination (RFE) using Logistic Regression (LR), Chi-squared test statistics (Chi2), and Discriminative Mutual Information (DMI) into our analysis for eliminating data redundancy. By implementing this comprehensive methodology, we aim to address the critical gap in predicting crash severity for large vehicles within work zones, leading to improved safety measures and targeted intervention strategies.

## DATA

The crash data from SC work zones for the years 2014-2018 are obtained from the SC Department of Transportation (SCDOT). This dataset, compiled from the SC Traffic Collision Report Form (TR-310), includes all types of trucks, buses, and vans to ensure a sufficiently large sample size for analysis. A total of 5,351 crashes are used in this study. The original dataset categorizes the target variable, ‘Crash Severity,’ into five levels: no apparent injury, possible injury, suspected minor injury, suspected major injury, and fatality. Of the 5,351 data points, the ‘no apparent injury (Low-Severity [LS])’ category constitutes the majority, with 4,217 data points (79%). To balance the dataset, all other categories are consolidated into a single ‘Injury/Fatality (High-Severity [HS])’ category. It is to be noted that LS corresponds to PDO (Property Damage Only) crashes, and HS corresponds to F+I (Fatality/Injury) crashes. **Figure 2** illustrates the frequency distribution across crash severity categories.

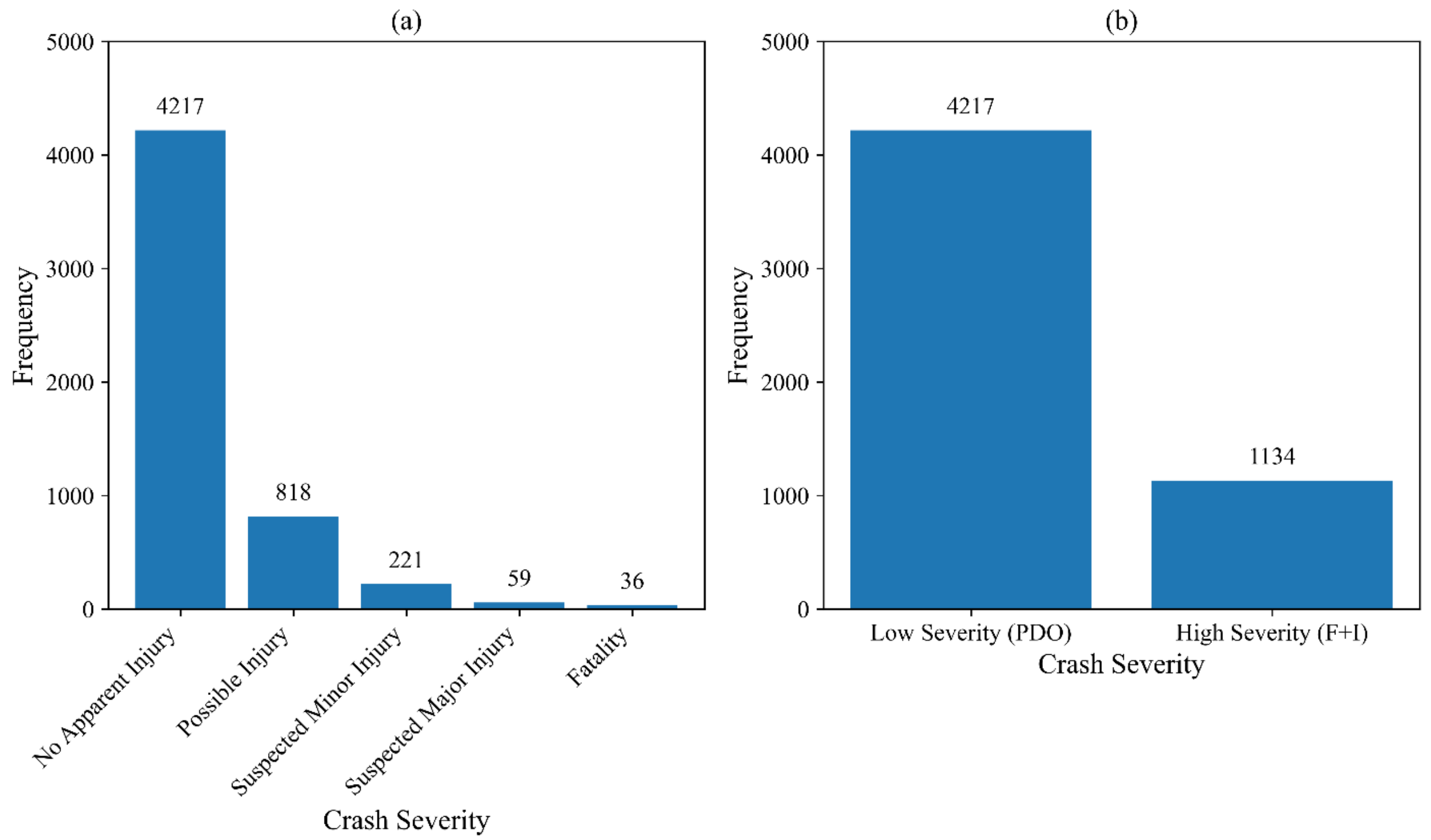

**Figure 2 Frequency Distribution of the Target Variable 'Crash Severity': (a) With Five Severity Categories and (b) With Two Categories.**

The dataset from 2014–2018 was provided to the authors by SCDOT in 2023. While the dataset covers a six-year period, data for years beyond 2018 are not available at the time of analysis, as SCDOT's dataset submission is limited to projects conducted during this timeframe. To prepare the dataset for analysis, several preprocessing steps are conducted to address data quality and ensure relevance for modeling crash severity. Feature variables (features) with substantial missing values, defined as those with more than 50% missing data, are excluded from the dataset. For features with less than 10% missing values, data imputation is performed using mode replacement to retain these variables in the analysis. Mode replacement is a broadly used technique in traffic crash data analysis, especially for categorical features (*16*). Advanced imputation methods are not considered in this study because (i) the percentage of missing data is low (below 10%) and (ii) the primary focus of the study is not on imputing missing data but on evaluating model performance under various feature selection and data balancing strategies. It is to be noted that no features have missing values between 10% and 50%, simplifying the treatment of missing data.

In addition to addressing missing values, features deemed irrelevant to crash severity prediction, such as latitude, longitude, route number, and road name, are excluded from the analysis. However, during the preprocessing step, latitude, longitude, and route number values are cross-referenced to ensure they align with each other and confirm that the reported crashes occurred within designated work zones. Continuous numerical features, such as estimated collision speed, base offset distance, and time of collision, are categorized based on existing literature and expert judgment. For instance, 'Time of Collision' is categorized into five time-intervals to account for variations in crash likelihood across different times of the day. Additionally, 'Day of Collision,' originally a categorical variable, is consolidated into two categories: weekday and weekend, to capture differences in traffic patterns. **Table 1** presents the features classified by their types that are used in this analysis. The categories within each feature, along with their distribution of LS (PDO) and HS (F+I) crashes, are detailed in **Supplemental Table 1**, providing a comprehensive view of the dataset's structure and distribution.

**TABLE 1 Feature Variables in SC Work-Zone Crash Dataset**

| Feature Type | Feature Header (*17*) | Feature Variable Name |
|---|---|---|
| Road Characteristics | RCT | Route Category |
| | RAI | Route Auxiliary |
| | LOA | Lane Number |
| | DLR | Direction of Lane |
| | ODR | Base distance Direction |
| | AHC | Road Character |
| | RSC | Road Surface Condition |
| | XWK | Crosswalk |
| | BDO | Base Distance Offset |
| | BIR | Base Route Category |
| | SIC | Second Intersection Route Category |
| | JCT | Junction Type |
| | TWAY | Trafficway |
| | TCT | Traffic Control Type |
| | SPL | Posted Speed Limit |
| Vehicle Characteristics | IBUS | School Bus Involved |
| | UTC | Unit Type |
| Driver/Occupant Characteristics | NOC | Number of Occupants/Unit |
| | DSEX | Driver Gender |
| | DLC | Driver License Class |
| | OSL | Occupant Seating Location |
| | REU | Occupant Restraint Type |
| Environmental Characteristics | ALC | Lighting Condition |
| | WCC | Weather Condition |
| Work Zone Characteristics | WZT | Work Zone Type |
| | WZL | Work Zone Location |
| | WPR | Workers Present |
| Crash Characteristics | UNT | Number of Units Involved in Collision |
| | DAY | Day of Week |
| | TIM | Time (of Collision) |
| | FHE | First Harmful Effect |
| | HEL | Harmful Event Location |
| | PRC | Primary Contributing Factor |
| | MAC | Manner of Collision |
| | CTA | Contributed to Collision |
| | ECS | Estimated Collision Speed |
| | API | Action Prior to Impact |
| | EDAM | Extent of Deformity |
| | MHE | Most Harmful Effect |
| | FDA | First Deformed Area |
| | LAI | Location After Impact |
| | SEV | Crash Severity (Target Variable in this study) |

The dataset incorporates a wide range of features to capture critical factors influencing crash severity, grouped into categories, such as Road Characteristics, Vehicle Characteristics, Driver/Occupant

Characteristics, Environmental Characteristics, Crash Characteristics, and Work Zone Characteristics. These features collectively offer a comprehensive view of the feature variables contributing to crash dynamics in work zones.

Road Characteristics account for the physical and operational attributes of the roadway. For instance, the 'Road Surface Condition' feature variable distinguishes between conditions, such as dry, wet, or icy roads. Dry road conditions dominate the dataset, comprising 89.76% of LS (PDO) crashes and 90.30% of HS (F+I) crashes. In contrast, wet road conditions are slightly less frequent in HS crashes (9.08%) compared to LS crashes (9.79%), indicating that adverse road surface conditions may not strongly correlate with severe outcomes in this dataset. Similarly, the 'Trafficway' feature variable, which classifies the type of roadway where the crash occurred, shows that crashes on two-way undivided roads account for 25.87% of LS crashes and 34.57% of HS crashes. This pattern highlights the elevated risk of severe crashes on roadways without median barriers or divisions, where opposing traffic flows are in close proximity.

Vehicle Characteristics capture information about the types and attributes of vehicles involved in crashes. For example, the 'Unit Type' feature variable identifies the category of the involved vehicle, with pickup trucks being the most frequently involved vehicles in both LS (46.76%) and HS (48.94%) crashes. Another significant feature variable, 'Occupant Restraint Type,' reveals that crashes where no restraint was used are disproportionately represented in HS crashes (6.53%) compared to LS crashes (0.95%), highlighting the role of safety measures in reducing crash severity.

Driver/Occupant Characteristics provide insights into the individuals involved in crashes. For example, the 'Driver License Class' feature variable highlights that drivers with a Commercial Driver's License (CDL [Classes A, B, and C]) are involved in 38.25% of LS crashes and 34.30% of HS crashes. The 'Driver Gender' feature variable reveals that male drivers constitute the majority of crash-involved operators, accounting for 87.43% of LS crashes and 83.16% of HS crashes, which reflects their higher representation in the larger vehicle driving workforce.

Environmental Characteristics focus on external factors, such as lighting and weather conditions. The 'Lighting Condition' feature variable shows that crashes occurring during daylight dominate both LS (77.26%) and HS (71.43%) crashes, while crashes under dark conditions with no street lighting are more prevalent in HS crashes (15.61%). The 'Weather Condition' feature variable reveals that clear weather conditions are associated with the majority of LS (86.15%) and HS (87.74%) crashes, though adverse weather, such as precipitation or fog, is slightly more common in HS crashes (8.24%) than in LS crashes (7.40%), based on the combined share of rain, fog, smoke, smog, and snow conditions.

Crash Characteristics describe the dynamics and circumstances of the crashes themselves. The 'Manner of Collision' feature variable shows that rear-end collisions account for the majority of both LS crashes (48.23%) and HS crashes (57.85%), highlighting their prominence in SC work-zone crash data. In contrast, head-on collisions, though rare overall, represent 0.81% of LS crashes but are disproportionately severe, comprising 2.47% of HS crashes. Similarly, angle collisions (e.g., Northeast/Northwest and East/West angles) are more frequently associated with severe crashes, collectively accounting for 19.37% of HS crashes compared to 13.22% of LS crashes. These patterns underscore the heightened risks posed by specific collision types in work zones. Additionally, the 'Number of Units Involved' feature variable indicates that crashes involving more than two vehicles are more prevalent in HS crashes (28.84%) compared to LS crashes (13.07%), reflecting the increased risk associated with multi-vehicle crashes in work zones.

The dataset also includes critical work zone characteristics, which provide detailed insights into the conditions and configurations of work zones. The 'Work Zone Type' feature variable categorizes the nature of work being conducted, with categories such as, Shoulder/Median Work, Lane Closure, and Intermittent/Moving Work. Crashes in 'Shoulder/Median Work' zones represent 47.93% of LS crashes and 48.85% of HS crashes, indicating that these zones contribute significantly to overall crash occurrences. Similarly, the 'Work Zone Location' feature variable identifies the specific section of the work zone where crashes occur, such as the Advanced Warning Area, Transition Area, Activity Area, and Termination Area. The ‘Activity Area,’ where active work is conducted, dominates the dataset, comprising 64.15% of LS crashes and 67.64% of HS crashes, reflecting the heightened exposure to risks in this segment. Another important feature variable, Workers Present, reveals that 48.33% of LS crashes and 48.59% of HS crashes occur in work zones with workers present, underscoring the importance of incorporating worker-related safety measures.

Furthermore, the dataset includes information on ‘Traffic Control Type,’ which categorizes the traffic management measures present in work zones, such as stop-and-go lights, officers or flaggers, pavement markings, stop signs, and work zone-specific controls. Among these, crashes in zones managed by work zone-specific controls are the most common, accounting for 47.69% of LS crashes and 40.74% of HS crashes, followed by pavement markings (only), which represent 11.45% of LS crashes and 10.14% of HS crashes. In contrast, crashes associated with officers or flaggers are less frequent, comprising 3.04% of LS crashes and 4.76% of HS crashes. However, their presence still highlights ongoing challenges in managing work zone traffic effectively. Notably, a significant proportion of crashes occurred in areas with no active traffic control measures, comprising 26.46% of LS crashes and 29.37% of HS crashes, suggesting that the absence of traffic control measures may contribute to a higher risk of severe outcomes. These findings indicate that despite the presence of various traffic control strategies, crashes continue to occur at considerable rates. This underscores the need for further evaluation of the design, implementation, and effectiveness of traffic control measures in mitigating crash severity in work zones.

These LS and HS crash distributions across feature categories are detailed in **Supplemental Table 1**, offering a comprehensive view of how these variables influence crash outcomes and providing a strong foundation for modeling crash severity in work zones.

## METHOD

To develop and evaluate crash severity prediction models for large vehicles in SC work zones, this study follows a structured framework involving data preprocessing, feature selection, data balancing, model training, and performance evaluation. The detailed flowchart of the methodological framework is presented in **Figure 3**.

The process begins with crash data provided by SCDOT, followed by data preprocessing steps including addressing missing values, excluding irrelevant features, categorizing continuous variables, and generating interaction terms to prepare the dataset for modeling. The next step is the application of feature selection techniques. If feature selection is applied, only the selected features are retained for modeling; otherwise, the full set of features is used. The cleaned dataset is then split into a training dataset (70%) and a test dataset (30%), ensuring consistency across experimental setups. It is to be noted that, the test set is held constant for each modeling scenario to enable fair performance comparison.

In this study, feature selection is performed prior to data balancing to reduce model complexity and mitigate overfitting risks, especially given the dataset's high dimensionality. This ordering ensures that informative features are selected before the class distribution is altered. However, prior research (*18*) suggests that the sequence of these steps can affect performance depending on the dataset. While the reversed order was not examined in this study, future work will explore whether applying data balancing before feature selection further enhances predictive performance.

Following the train-test split, data balancing techniques are applied to the training dataset as part of the experimental setup to assess their impact on model performance. If a data balancing technique is used, such as oversampling, undersampling, or class weighting, it is applied only to the training dataset to avoid adding bias to the evaluation. Based on whether a data balancing technique is applied, models are trained using either the original imbalanced training dataset or a balanced dataset generated through oversampling, undersampling, or class weighting. Once the models are trained, they are evaluated on the untouched test dataset and performance metrics such as balanced accuracy and class-specific precision, recall, and F1-scores are computed to assess predictive capability.

This framework provides a systematic comparison of multiple feature selection and data balancing strategies while also identifying optimal combinations of feature selection, data balancing, and modeling approaches for specific crash severity prediction objectives. This contributes to improving model performance in imbalanced and overlapping crash datasets involving large vehicles, an area that remains underexplored in current crash severity research. Details regarding the feature selection and data balancing techniques, models, and performance metrics used in this study are provided in the subsequent subsections. In addition, details on data preprocessing, the generation of training datasets combining different feature selection and data balancing techniques, and model training are provided in the Analysis section.

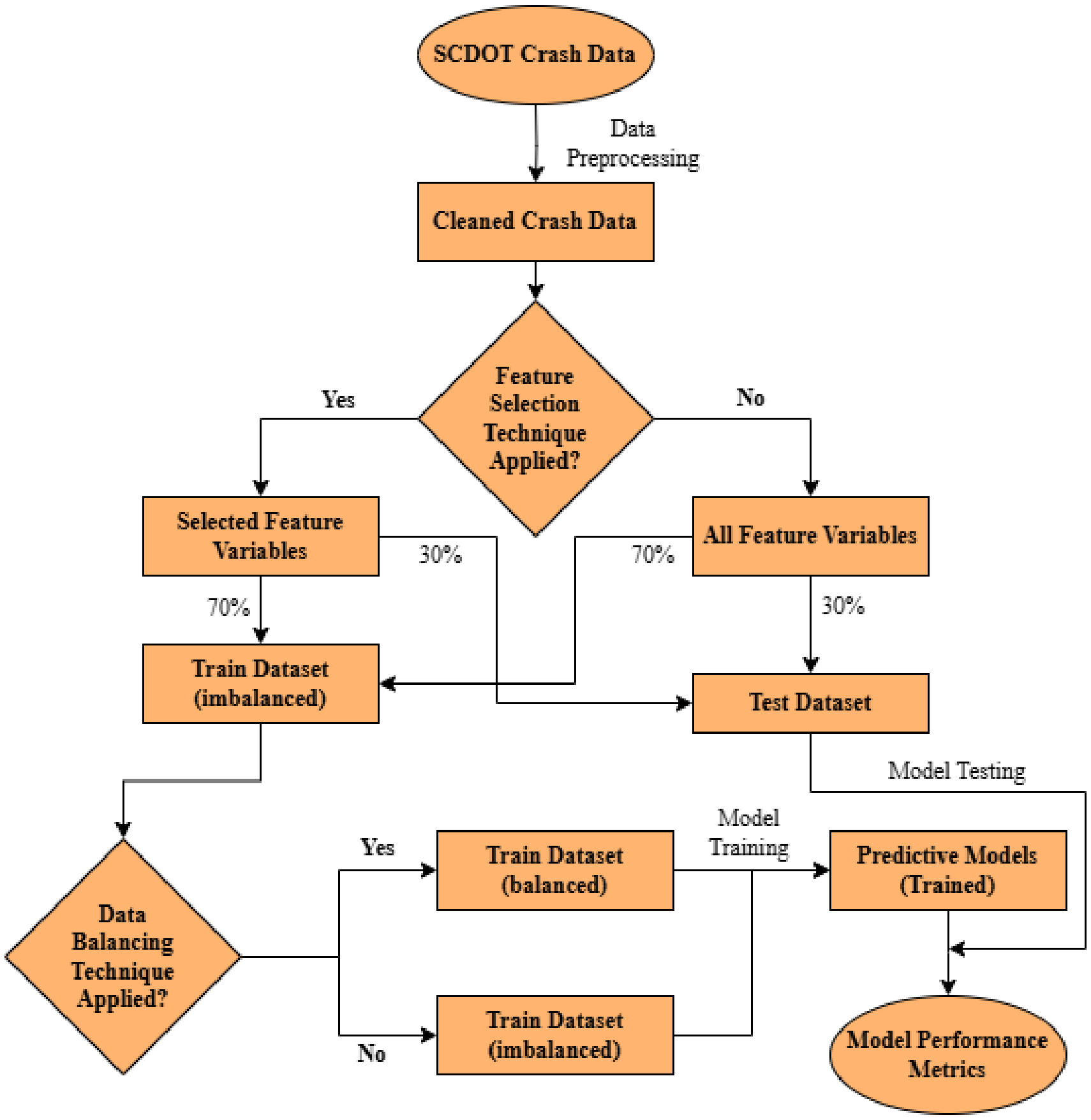

**Figure 3 Methodological Framework for Developing and Evaluating Risk Models**

## Feature Selection Techniques

Various feature selection techniques are employed in this study to identify the most relevant features contributing to the prediction of crash severity. **Table 2** presents a summary of these feature selection techniques.

**TABLE 2 Feature Selection Techniques Used in This Study**

| Feature Selection Technique | Description |
|---|---|
| Pearson's Correlation Coefficient (*19*) | Pearson's correlation coefficient measures the strength and direction of the linear relationship between a feature variable and a target variable, helping to identify the most relevant features based on their correlation with the target variable. |
| Feature Importance (FI) using Random Forest (RF) (*20, 21*) | RF models use feature importance scores, which indicate how much each feature contributes to the final prediction of the model, based on the reduction in impurity that each feature contributes. |

| Feature Selection Technique | Description |
|---|---|
| Recursive Feature Elimination (RFE) using Logistic Regression (LR) (*22*) | RFE is an iterative process that starts with all features and sequentially removes the least important features based on their contribution to a LR model's prediction of the target variable. |
| Chi-squared Statistical Test (*23*) | The Chi-squared statistical test assesses the independence between categorical features and the target variable, identifying features with significant dependence. |
| Discriminative Mutual Information (DMI) (*24*, *25*) | DMI measures the mutual information between features and the target variable, emphasizing information gain specifically for distinguishing between target variable categories. |

## Data Balancing Techniques

Crash severity data inherently suffers from class imbalance, with significantly more LS (PDO) crashes than HS (F+I) crashes. This imbalance can skew model performance towards the majority class, making it less effective at predicting the minority class of HS crashes. To mitigate this issue, various data balancing techniques are employed in this study. **Table 3** presents a summary of these data balancing techniques.

**TABLE 3 Data Balancing Techniques Used in This Study**

| Technique Category | Data Balancing Technique | Description |
|---|---|---|
| Weighting the Minority Class | Assigning Fixed Weight to Minority Class (*23*) | This technique assigns higher weights to minority class instances during model training. This ensures that these less frequent but critical instances have a greater influence on the learning process, reducing bias towards the majority class. However, this approach may not fully address the imbalance for large differences in class sizes. |
| Oversampling the Minority Class | Synthetic Minority Oversampling Technique (SMOTE) (*26*) | SMOTE generates new synthetic minority class instances by interpolating between existing ones. While this technique effectively increases the representation of the minority class, it introduces artificial data that may not accurately represent real-world scenarios, potentially leading to overfitting. |
| | KMeansSMOTE (*27*) | KMeansSMOTE is a variation of SMOTE that uses k-means clustering to identify similar minority class instances and generate new data points within those clusters. This technique captures more detailed patterns within the minority class, potentially leading to better model performance. |
| | Adaptive Synthetic Minority Oversampling Technique (ADASYN) (*28*) | ADASYN focuses on oversampling minority class instances that are more difficult for the model to learn from. It calculates the density distribution of each minority class sample and generates synthetic samples accordingly, which can improve performance for complex models by focusing on harder-to-learn instances. |
| | RandomOver-Sampler (*29*) | RandomOverSampler randomly samples with replacements from minority class instances until a desired balance is achieved. This technique increases the representation of the minority class in the training data, helping the model to learn from these instances more effectively. |
| | Kernel-based Synthetic Minority Oversampling (K-SMOTE) (*29*) | K-SMOTE estimates the probability distribution of the features for each class (minority and majority). It uses a kernel function centered on existing minority class data points, and draws samples from these smoothed distributions, creating new synthetic data points that resemble existing minority class points. |

| Technique Category | Data Balancing Technique | Description |
|---|---|---|
| | Wasserstein Generative Adversarial Network with Gradient Penalty (WGAN-GP) (*30*) | Generative Adversarial Networks (GANs) consist of two networks, a generator and a discriminator, that work in tandem to produce realistic synthetic data. The generator produces fake data points, while the discriminator attempts to distinguish between real and fake data. Through this adversarial process, the generator improves its ability to create realistic data. WGAN-GP addresses training instability and convergence issues by employing the Wasserstein distance and a gradient penalty. This technique enables the generation of realistic synthetic data points by learning complex data distributions, ensuring that the synthetic instances closely resemble real minority class instances. |
| | Conditional WGAN-GP (*31*) | Conditional WGAN-GP builds upon WGAN-GP by incorporating all labels of the target variable into the synthetic data generation process, which allows for more targeted oversampling (i.e., oversampling from the minority class). |
| Undersampling the Majority Class | NearMiss-1 (*32*) | NearMiss-1 is a targeted undersampling technique that carefully selects majority class instances based on their proximity to majority class instances (i.e., the smallest average distance to the three closest minority class instances). By focusing on the majority class samples that are closest to the minority class, NearMiss-1 effectively reduces overlap between the classes by ensuring that the selected majority class samples are those most similar to the minority class. This helps to create a clearer decision boundary, reducing confusion and improving the classifier's ability to differentiate between the two classes. |
| | RandomUnder-Sampler (*29*) | RandomUnderSampler randomly selects the majority class instances without replacement until a desired balance is achieved. This technique reduces the influence of the majority class, enabling the model to give more attention to the under-represented minority class. It is simple to implement but may discard potentially useful data. |
| Combination of Oversampling and Under-sampling | Combined RandomOver-Sampler and RandomUnder-Sampler (*29*) | This technique integrates both RandomOverSampler and RandomUnderSampler techniques by replicating existing minority class instances and randomly selecting majority class instances without replacement until a desired balance is achieved. |

## Models

For this study, several training datasets are prepared using different combinations of feature selection and data balancing techniques presented in the preceding subsections. Subsequently, a range of models are trained with each training dataset. These models are chosen to cover a diverse set of approaches, from traditional statistical methods to advanced machine learning and deep learning algorithms. **Table 4** provides a summary of these models along with their respective descriptions.

### TABLE 4 Models Used in This Study

| Model | Description |
|---|---|
| Bayesian Mixed Logit (BML) (*33*) | The BML model extends the traditional logit model (*34*) by incorporating random effects and using Bayesian inference methods. It is useful for analyzing binary response variables where outcomes are categorical and inherently binary. The model accounts for unobserved heterogeneity among individuals by allowing coefficients to vary randomly. This approach leverages Markov Chain Monte Carlo (MCMC) methods to estimate the posterior distributions of the model parameters, enhancing the model's flexibility and interpretability. |

| Model | Description |
|---|---|
| Categorical Boosting (CatBoost) (*35*) | CatBoost is a gradient boosting algorithm designed to handle categorical variables effectively. It builds an ensemble of decision trees sequentially, where each tree is constructed to correct the errors of the preceding ones. CatBoost's unique algorithmic approach allows it to process categorical variables in their raw form, reducing information loss and improving accuracy. |
| Extremely Randomized Trees (Extra Trees) (*36*) | Extra Trees introduces additional randomness in decision tree construction to enhance diversity and reduce overfitting. The model splits nodes by selecting random thresholds for each feature and choosing the best split among them, which increases the variability of the trees. It uses simple criteria such as Gini impurity and entropy for splits and combines the predictions of multiple trees through majority voting to improve robustness and accuracy. |
| Light Gradient Boosting Machine (LightGBM) (*37*) | LightGBM is an efficient gradient boosting framework that uses gradient-based one-side sampling (GOSS) and exclusive feature bundling (EFB) to reduce data processing, achieving lower losses with leaf-wise tree growth. LightGBM builds trees leaf-wise (best-first search) instead of level-wise, allowing it to achieve higher accuracy and efficiency by prioritizing informative splits. It is well-suited for large datasets and provides several variants such as LightGBMLarge, optimized for higher accuracy and handling larger datasets, and LightGBMXT, designed for classification tasks with a large number of classes. |
| NeuralNetTorch (*38*) | NeuralNetTorch, built on PyTorch, offers a flexible environment for designing custom neural networks. It simplifies aspects, such as layer definition, loss computation, and backpropagation by providing a structured and adaptable interface. NeuralNetTorch allows for easy implementation and experimentation with complex models, making it suitable for various deep learning tasks. |
| NeuralNetFastAI (*39*) | NeuralNetFastAI, built on FastAI, provides high-level abstractions for common tasks, emphasizing ease of use and productivity. It includes features such as one-cycle learning rate policies, transfer learning tools, and data augmentation pipelines. NeuralNetFastAI is particularly suited for rapid prototyping and applications where quick development is critical. |
| Random Forest (RF) (*20*) | RF is an ensemble learning technique that builds multiple decision trees to improve robustness and accuracy. Each tree is trained on a randomly selected subset of the data and features, and the final prediction is made by aggregating the predictions from all the trees through majority voting. RF uses criteria such as Gini impurity (RandomForestGini) and entropy (RandomForestEntr) to select the best splits, and it helps mitigate overfitting by averaging multiple trees, which reduces variance and improves generalization. |
| eXtreme Gradient Boosting (XGBoost) (*40*) | XGBoost is a scalable gradient boosting implementation known for handling large-scale data efficiently. It improves model accuracy by sequentially adding decision trees, where each tree corrects errors made by the previous ones. XGBoost introduces several advanced features, such as handling missing values, tree pruning, and efficient split finding, which contribute to its improved performance. It supports various objective functions, making it adaptable for classification, regression, ranking, and user-defined prediction tasks. |

## Performance Metrics

To compare the performance of the crash severity risk modeling, a collection of metrics is used. These metrics, i.e., accuracy, precision, recall, F1-score, balanced accuracy, and Receiver Operating Characteristic - Area Under the Curve (ROC AUC), are critical for evaluating different aspects of model performance, such as its overall correctness, ability to identify positive cases, and balance between precision and recall (*41*). Detailed definitions and the significance of each metric are provided in **Table 5**.

**TABLE 5 Performance Metrics**

| Performance Metrics | Description | Formula |
|---|---|---|
| Accuracy | Accuracy measures the ratio of correctly predicted instances (both true positives and true negatives) to the total number of cases evaluated. It provides an overall assessment of how often the model is correct. | $Accuracy = \frac{TP + TN}{TP + TN + FP + FN}$ <br><br> Here, TP: True Positive; TN: True Negative; FP: False Positive, and FN: False Negative |
| Precision (Specificity) | Precision evaluates the proportion of true positive predictions among the total predicted positives. It is crucial for scenarios where the cost of a false positive is high, indicating the model's reliability in classifying a case as positive. | $Precision = \frac{TP}{TP + FP}$ |
| Recall (Sensitivity) | Recall measures the percentage of true positive instances that the model correctly identifies. It is particularly important in contexts, where missing a positive (such as a High-Severity crash) is critical. | $Recall = \frac{TP}{TP + FN}$ |
| F1-Score | The F1-score represents the harmonic mean of precision and recall, offering a single metric to evaluate the trade-off between them. It is important when we need to consider both the model's precision and recall equally, especially in imbalanced datasets. | $F1\ Score = \frac{2 \times Precision \times Recall}{Precision + Recall}$ |
| Balanced Accuracy | Balanced Accuracy accounts for class imbalance by averaging the model's sensitivity (recall for the positive class) and specificity (recall for the negative class). It provides an equitable assessment of performance when classes are imbalanced by giving equal importance to both classes. This metric is especially useful in scenarios where the minority class (e.g., High-Severity crashes) is of critical interest. | $Balanced\ Accuracy = \frac{1}{2}\left(\frac{TP}{TP + FN} + \frac{TN}{TN + FP}\right)$ |
| ROC AUC | ROC AUC evaluates a model's ability to discriminate between classes across all thresholds. The AUC measures the probability that the model will correctly rank a randomly chosen positive instance above a randomly chosen negative instance. It is important for understanding the model's overall discriminative power. | ROC AUC involves plotting the true positive rate (Recall) against the false positive rate (FP / (TN + FP)) at various threshold settings and then calculating the area under the resulting curve. |

## ANALYSIS

The analysis framework for this study outlines the process used to develop and evaluate risk models for crash severity in incidents involving large vehicles within SC work zones. This framework includes data preprocessing, the development of training datasets with combinations of feature selection and data balancing techniques, the tools used to implement the models and techniques, and the methods for training the models.

### Data Preprocessing

As discussed in the DATA section, the initial step of data preprocessing involves addressing missing values, excluding irrelevant features, and categorizing continuous variables based on literature and expert judgment. Post-cleaning, forty-one (41) feature variables remain (**Table 1**), which are then one-hot-encoded to create 458 binary feature variables.

To enhance model performance and explore non-linear interactions, additional interaction terms are generated specifically from the 458 binary feature variables in the dataset. These interaction terms are selected based on their Pearson's correlation coefficient with the target variable, Crash Severity. Only interaction terms with a Pearson's correlation coefficient greater than 0.40 are retained for further analysis. The threshold of 0.40 is determined through sensitivity analysis, balancing the trade-off between including interactions and avoiding overfitting. This step results in a final dataset containing 547 feature variables, which serves as the base dataset for model training and evaluation.

**Feature Selection and Data Balancing Techniques**

A total of 24 training datasets are generated by combining different feature selection and data balancing techniques used in this study. The description of these training datasets is provided in **Table 6**.

To develop a strong and representative feature set, five distinct feature selection techniques are employed: Pearson's correlation, Feature Importance using Random Forest (FI using RF), Recursive Feature Elimination (RFE) using Logistic Regression, Chi-squared Test Statistics (Chi2), and Discriminative Mutual Information (DMI). Each technique ranks features based on different criteria, offering complementary insights into variable importance. For consistency and comparability, the top 50 ranked binary or interaction features are selected from each feature selection technique. This number is chosen based on sensitivity analysis to balance predictive power and computational efficiency. It is to be noted that the 50 selected variables correspond to the 547 binary features and interaction terms, not the raw features having multiple categories.

To enhance the robustness of the feature set, the top 50 ranked features selected by all five feature selection techniques are merged, resulting in a combined set of 124 unique features. This merging process leverages the complementary strengths of different feature selection methods, ensuring that the most informative features from each approach are included. By narrowing the feature set to these 124 variables, the study is designed to minimize the risk of the model capturing irrelevant or noisy patterns in the data, such as random fluctuations and outliers, instead of underlying trends. This approach also addresses potential sparsity that could affect parameter estimation. The top 124 variables resulting from the merging process are listed in **Supplemental Table 2**.

However, during performance evaluation, models trained using features selected exclusively by DMI (Training Dataset 7) demonstrated strong results in terms of HS recall, exceeding the performance of models built on the merged feature set. These observations suggest that DMI-selected features are particularly effective in capturing the distinguishing characteristics of HS crashes. To further explore the potential of DMI under varying data balancing techniques, five additional training datasets were introduced (Training Datasets 20 - 24 in **Table 6**). The data balancing techniques used in these training datasets are selected based on their strong performance, particularly in achieving high HS recall and balanced model performance, when previously applied to the merged feature set. The DMI-selected features included in the merged set are highlighted in bold in **Supplemental Table 2**.

**TABLE 6 Description of Training Datasets**

| Training Dataset Number | Applied Feature Selection and Data Balancing Techniques |
|---|---|
| 1 | All feature variables with no feature selection and data balancing technique |
| 2 | All feature variables with no feature selection; High-Severity class assigned a weight of 4 |
| 3 | Top 50 ranked feature variables selected using *Pearson's correlation coefficient*; No data balancing method |
| 4 | Top 50 ranked feature variables selected using *FI using RF*; No data balancing method |
| 5 | Top 50 ranked feature variables selected using *RFE using LR*; No data balancing method |
| 6 | Top 50 ranked feature variables selected using *Chi-squared Test Statistics*; No data balancing method |
| 7 | Top 50 ranked feature variables selected using *DMI*; No data balancing method |
| 8 | Feature variables from merging top 50 feature variables selected in training dataset numbers 3, 4, 5, 6, and 7; No data balancing method |
| 9 | Feature variables from merging top 50 feature variables selected in training dataset numbers 3, 4, 5, 6, and 7; High-Severity class assigned a weight of 4 |

| **Training Dataset Number** | **Applied Feature Selection and Data Balancing Techniques** |
|---|---|
| 10 | Feature variables from merging top 50 feature variables selected in training dataset numbers 3, 4, 5, 6, and 7; Oversampling of High-Severity class using *SMOTE* |
| 11 | Feature variables from merging top 50 feature variables selected in training dataset numbers 3, 4, 5, 6, and 7; Oversampling of High-Severity class using *KMeansSMOTE* |
| 12 | Feature variables from merging top 50 feature variables selected in training dataset numbers 3, 4, 5, 6, and 7; Oversampling of High-Severity class using *ADASYN* |
| 13 | Feature variables from merging top 50 feature variables selected in training dataset numbers 3, 4, 5, 6, and 7; Oversampling of High-Severity class using *RandomOverSampler* |
| 14 | Feature variables from merging top 50 feature variables selected in training dataset numbers 3, 4, 5, 6, and 7; Oversampling of High-Severity class using *Kernel-based Synthetic Minority Oversampling* |
| 15 | Feature variables from merging top 50 feature variables selected in training dataset numbers 3, 4, 5, 6, and 7; Oversampling of High-Severity class using *WGAN-GP* |
| 16 | Feature variables from merging top 50 feature variables selected in training dataset numbers 3, 4, 5, 6, and 7; Oversampling of High-Severity class using *Conditional WGAN-GP* |
| 17 | Feature variables from merging top 50 feature variables selected in training dataset numbers 3, 4, 5, 6, and 7; Undersampling of Low-Severity class using *NearMiss-1* |
| 18 | Feature variables from merging top 50 feature variables selected in training dataset numbers 3, 4, 5, 6, and 7; Undersampling of Low-Severity class using *RandomUnderSampler* |
| 19 | Feature variables from merging top 50 feature variables selected in training dataset numbers 3, 4, 5, 6, and 7; Oversampling of High-Severity class using *RandomOverSampler* and undersampling of Low-Severity class using *RandomUnderSampler* |
| 20 | Top 50 ranked feature variables selected using *DMI*; High-Severity class assigned a weight of 4 |
| 21 | Top 50 ranked feature variables selected using *DMI*; Oversampling of High-Severity class using *RandomOverSampler* |
| 22 | Top 50 ranked feature variables selected using *DMI*; Undersampling of Low-Severity class using *NearMiss-1* |
| 23 | Top 50 ranked feature variables selected using *DMI*; Undersampling of Low-Severity class using *RandomUnderSampler* |
| 24 | Top 50 ranked feature variables selected using *DMI*; Oversampling of High-Severity class using *RandomOverSampler* and undersampling of Low-Severity class using *RandomUnderSampler* |

## Implementation of Techniques

The feature selection and data balancing techniques are primarily implemented in Python using the scikit-learn (*42*) and imblearn (*43*) libraries. However, RandomOverSampler, RandomUnderSampler, combined RandomOverSampler and RandomUnderSampler, and K-SMOTE techniques are implemented using the ROSE package in R (*29*). DMI, WGAN-GP, and Conditional WGAN-GP techniques are coded and implemented in Python.

## Dataset Splitting and Model Training

The datasets for training and testing are prepared with a 70/30 split, a commonly adopted ratio in machine learning to ensure an optimal balance between model training and evaluation (*44*). Allocating 70% of the data for training provides the model with sufficient exposure to diverse patterns and relationships in the dataset while reserving 30% as an independent test dataset, enabling an unbiased evaluation of the model's generalization performance on unseen data. This split maintains a consistent ratio of LS (PDO) crashes to HS (F+I) crashes across both datasets. Specifically, the training dataset includes 2,951 LS crashes and 793 HS crashes, while the test dataset includes 1,266 LS crashes and 341 HS crashes. This approach ensures that the test dataset is large enough to reliably assess performance metrics, particularly for the minority HS (F+I) class in this study, while leaving adequate data for effective model training.

To ensure model robustness and reduce the risk of overfitting, 5-fold cross-validation is performed on the training dataset. During cross-validation, the training dataset is divided into five approximately equal parts using a stratified approach, which ensures that the proportion of LS and HS crashes in each fold reflects the overall class distribution in the dataset. In each iteration, four parts are used to train the model, while the remaining part serves as a temporary validation set to evaluate model performance. This process is repeated five times, with each part serving as the validation set once, ensuring that the model is tested on all subsets of the training data. Although 10-fold cross-validation was also tested, it produced poorer performance compared to the 5-fold approach. This outcome may be attributed to the smaller fold sizes in 10-fold cross-validation, which could have led to insufficient representation of the minority HS class in individual folds, thereby destabilizing model training.

Data balancing techniques are applied only to the training dataset to address the class imbalance, while the test dataset remains untouched to provide an unbiased evaluation of the models' performance on imbalanced data. The BML model is trained in R using the brms package (*45*), while all other machine learning (ML) and deep learning (DL) models are implemented in Python using the AutoGluon library (*46*). Each ML/DL model is trained using a 5-fold bagging ensemble strategy under the 'best quality' preset, which enables hyperparameter tuning, multi-model ensembling, and stacking, to maximize predictive performance and generalizability. A maximum training time limit of two hours is imposed per model to prevent overfitting and ensure that all models are trained within a consistent and computationally feasible framework.

The hyperparameters used for each ML and DL model are summarized in **Supplemental Table 3**. These configurations reflect the default and optimized settings automatically selected by AutoGluon's 'best quality' preset (*46*), which applies consistent hyperparameter tuning procedures across all training datasets. For ensemble-based models, the hyperparameters represent those of the best-performing model variants selected based on balanced accuracy on the validation folds (i.e., the average of recall values across both classes, which helps mitigate the influence of class imbalance by giving equal weight to minority and majority class performance). The Bayesian Mixed Logit (BML) model configuration reflects the priors and settings defined using the brms package in R.

## RESULTS AND DISCUSSION

This section presents the performance of risk models for predicting crash severity in crashes involving large vehicles within SC work zones. Given the critical importance of accurately predicting HS (F+I) crashes to enable timely and effective interventions, the primary focus is on HS-specific metrics such as HS precision, HS recall, and HS F1-score. HS precision is crucial to minimize false positives and ensure efficient resource allocation, while HS recall is essential to capture as many HS crashes as possible for timely interventions. The HS F1-score provides a balanced measure of both precision and recall for HS crashes. Additionally, the models' performance in predicting LS (PDO) crashes are assessed using LS precision, LS recall, and LS F1-score.

To further evaluate models' performance in the context of class imbalance, balanced accuracy is included as a key metric. Balanced accuracy calculates the average of recall for both HS and LS classes, offering a less biased representation of model performance when the data is skewed toward the majority class. The analysis also considers ROC AUC, which reflects the model's ability to distinguish between LS and HS crashes across all classification thresholds.

### Overall Performance of Models

The output of the risk models for crash severity related to incidents involving large vehicles within SC work zones reveal insightful trends and observations. Each of the 24 training datasets (**Table 6**) is used to train 12 different models: BML, CatBoost, Extra Trees [ET] (2 variants), LightGBM (3 variants), NeuralNet (2 variants), Random Forest [RF] (2 variants), and XGBoost, resulting in 288 model-training dataset combinations. **Figure 4** presents the summary statistics of the model performance on the test dataset for these 288 combinations, illustrating the performance metrics across all training datasets and models. Detailed performance metrics for each model-training dataset combination are presented in **Supplemental Table 4**.

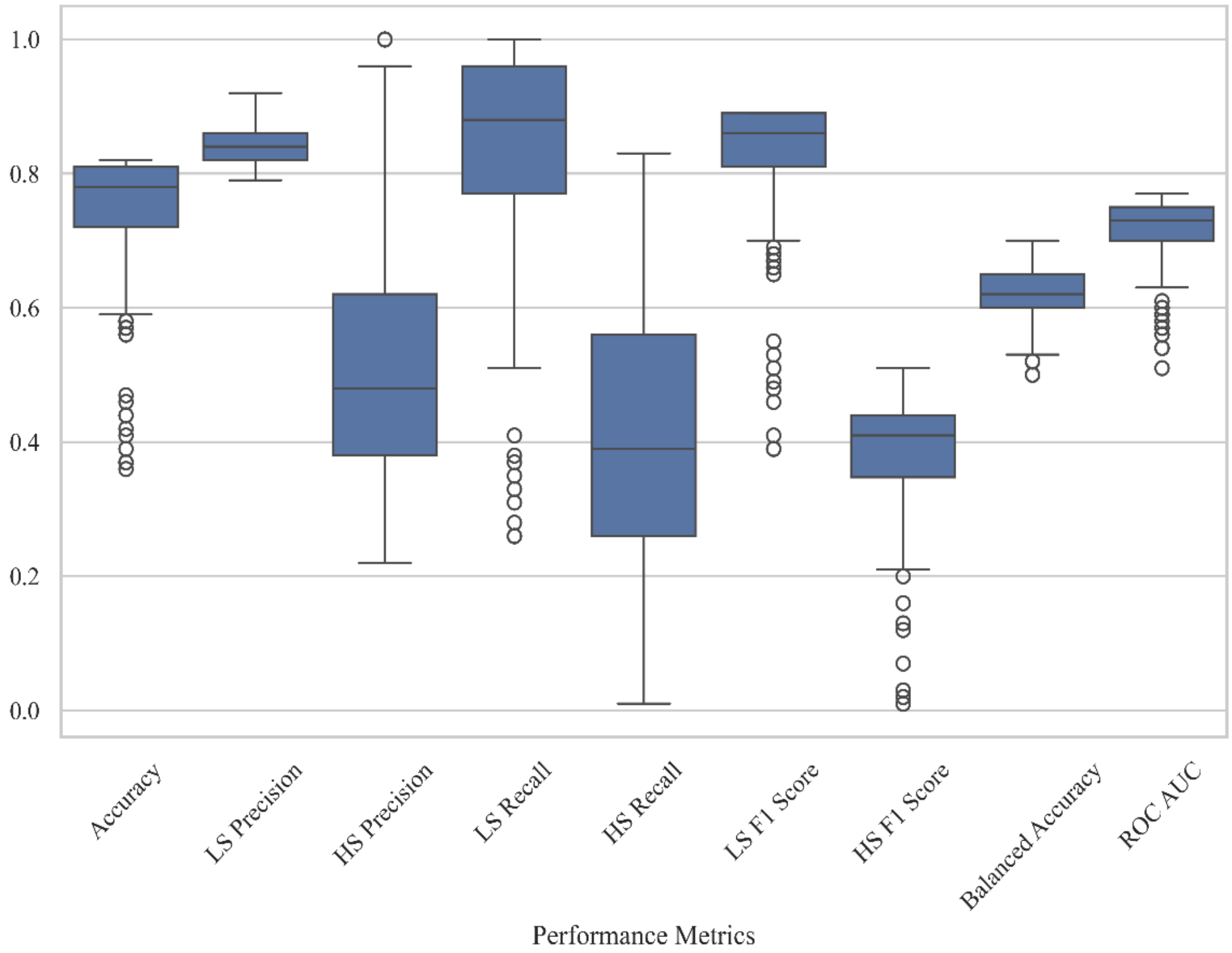

**Figure 4 Summary of Performance Metrics Across All Model-Training Dataset Combinations**

The boxplot of performance metrics (**Figure 4**) reveals a noticeable disparity between LS (PDO) and HS (F+I) crash predictions. LS precision and recall values are consistently high, indicating that models are generally effective at identifying non-severe incidents. In contrast, predicting HS crashes is more challenging due to the class imbalance in the dataset, with far more LS crashes than HS crashes, and the overlap between the two classes. This challenge is reflected in the broader variability observed in HS-related metrics.

The ranges and quartiles shown in the boxplot (**Figure 4)** offer deeper insights into the variability of model performance. HS recall exhibits the widest spread, followed by HS precision, underscoring the difficulty in accurately predicting HS crashes. In contrast, LS precision exhibits the smallest spread, indicating consistent performance across models. Metrics such as accuracy, LS F1-score, balanced accuracy, and ROC AUC also exhibit narrower spreads, reflecting the models' consistency and reliability in identifying LS crashes and maintaining strong overall classification performance. However, LS recall shows relatively higher variability compared to other LS metrics, which may arise from an inherent trade-off between optimizing HS and LS predictions. Models and techniques that prioritize improving HS recall tend to compromise on LS recall due to the overlapping feature space and imbalanced class distribution.

It is important to note that the maximum values for performance metrics are achieved through different combinations of training datasets and models rather than a single optimal model. For instance, the highest accuracy of 82% is achieved by all models considered in this study except LightGBM and LightGBMXT, each trained on different datasets. However, these models result in low HS recall values (18-33%). Similarly, the highest HS recall of 83% is achieved by the CatBoost model trained on training dataset 7 (**Table 6**), which employs Discriminative Mutual Information (DMI) as the feature selection technique without any data balancing technique. While this result highlights the model's ability to capture HS crashes, it comes at the cost of reduced overall accuracy (58%) and LS recall (52%).

The trade-offs between metrics further illustrate the complexity of optimizing predictions for HS crashes. For instance, the highest HS precision of 100% is observed in CatBoost, ExtraTreesEntr, LightGBMLarge, LightGBMXT, RandomForestEntr, RandomForestGini, and XGBoost models trained on training dataset 14 (**Table 6**), which uses oversampling of HS crashes with K-SMOTE. While this demonstrates high confidence in the identified HS crashes, the corresponding HS recall values are close to zero, indicating that these models fail to predict any HS crashes.

Similarly, the highest ROC AUC value of 0.77 is achieved by multiple models trained on different datasets, reflecting strong overall discriminative ability. However, these high ROC AUC values do not consistently translate to effective HS crash identification. For example, the LightGBM model trained on training dataset 9 (HS weight of 4) achieves both a high ROC AUC (0.77) and a relatively strong HS recall (62%), indicating a good balance between overall classification performance and minority class sensitivity. In contrast, the ExtraTreesEntr also achieves ROC AUC values of 0.77 when trained on training dataset 9, but its HS recall remains low (18%). These findings underscore that while ROC AUC is a valuable indicator of model discrimination, it should be interpreted alongside class-specific metrics such as HS recall to assess real-world effectiveness in identifying severe crashes.

In addition to ROC AUC, balanced accuracy serves as a critical metric for evaluating model robustness in the presence of class imbalance. Unlike overall accuracy, which can be inflated by the dominant LS class, balanced accuracy gives equal weight to both HS and LS predictions by averaging their respective recall scores. This makes it particularly useful for identifying models that perform well across both classes rather than just favoring the majority. In this study, several models, including NeuralNetTorch, NeuralNetFastAI, CatBoost, LightGBM, XGBoost, and BML, achieved relatively high balanced accuracy (69–70%) while maintaining moderate to strong HS recall (60–64%), indicating their ability to deliver reliable performance across imbalanced class distributions. As such, balanced accuracy complements ROC AUC by offering a more equitable measure of classification performance in safety-critical applications such as crash severity prediction.

Based on these evaluations, two selection criteria are used for further analysis:

(i) HS recall is prioritized to ensure that the model successfully identifies a high proportion of severe crashes, which is essential for enabling timely interventions and reducing crash consequences.

(ii) Balanced accuracy and ROC AUC are used to evaluate the model's overall robustness and ability to provide consistent predictions across both classes. These metrics are particularly important in the context of imbalanced datasets, where accuracy alone may be misleading. Balanced accuracy ensures equal weight is given to both HS and LS predictions, while ROC AUC assesses the model's overall discriminative power regardless of classification thresholds.

**Table 7** presents summary statistics (mean, standard deviation, and range) for three key metrics: Balanced Accuracy, ROC AUC, and HS Recall across all 12 models evaluated in this study. These metrics provide a holistic comparison of model robustness, discriminative ability, and effectiveness in identifying HS crashes.

**TABLE 7 Performance Summary of All Models Based on Balanced Accuracy, ROC AUC, and HS Recall**

| Model | Balanced Accuracy | | ROC AUC | | HS Recall | |
|---|---|---|---|---|---|---|
| | Mean ± Std | (Min, Max) | Mean ± Std | (Min, Max) | Mean ± Std | (Min, Max) |
| BML | 0.63 ± 0.04 | (0.54, 0.69) | 0.73 ± 0.04 | (0.60, 0.76) | 0.45 ± 0.19 | (0.19, 0.75) |
| CatBoost | 0.64 ± 0.05 | (0.50, 0.70) | 0.72 ± 0.06 | (0.54, 0.77) | 0.45 ± 0.20 | (0.01, 0.83) |
| ExtraTreesEntr | 0.60 ± 0.04 | (0.52, 0.66) | 0.71 ± 0.04 | (0.56, 0.76) | 0.35 ± 0.19 | (0.06, 0.77) |
| ExtraTreesGini | 0.60 ± 0.03 | (0.52, 0.66) | 0.71 ± 0.04 | (0.57, 0.77) | 0.35 ± 0.19 | (0.07, 0.77) |
| LightGBM | 0.63 ± 0.04 | (0.52, 0.70) | 0.72 ± 0.05 | (0.54, 0.77) | 0.44 ± 0.20 | (0.04, 0.80) |
| LightGBMLarge | 0.62 ± 0.04 | (0.50, 0.66) | 0.70 ± 0.06 | (0.51, 0.75) | 0.39 ± 0.19 | (0.01, 0.78) |
| LightGBMXT | 0.62 ± 0.04 | (0.50, 0.70) | 0.71 ± 0.04 | (0.59, 0.76) | 0.41 ± 0.20 | (0.01, 0.79) |
| NeuralNetFastAI | 0.65 ± 0.03 | (0.55, 0.69) | 0.73 ± 0.04 | (0.59, 0.76) | 0.50 ± 0.13 | (0.27, 0.75) |
| NeuralNetTorch | 0.64 ± 0.04 | (0.54, 0.70) | 0.73 ± 0.04 | (0.59, 0.76) | 0.44 ± 0.17 | (0.09, 0.82) |
| RandomForestEntr | 0.60 ± 0.04 | (0.50, 0.66) | 0.71 ± 0.05 | (0.57, 0.77) | 0.35 ± 0.19 | (0.01, 0.77) |
| RandomForestGini | 0.60 ± 0.04 | (0.50, 0.66) | 0.71 ± 0.05 | (0.57, 0.77) | 0.35 ± 0.19 | (0.01, 0.77) |
| XGBoost | 0.63 ± 0.05 | (0.50, 0.70) | 0.72 ± 0.05 | (0.54, 0.76) | 0.42 ± 0.20 | (0.01, 0.81) |

As shown in **Table 7**, among the 12 evaluated models, NeuralNetFastAI achieves the highest average balanced accuracy (0.65 [mean] ± 0.03 [standard deviation]), indicating strong and consistent performance across both HS and LS classes. It also records the highest mean HS recall (0.50 ± 0.13), suggesting its superior sensitivity in identifying severe crashes. NeuralNetTorch also demonstrates strong performance, achieving a balanced accuracy of 0.64 ± 0.04 and HS recall of 0.44 ± 0.17.

BML, CatBoost, and LightGBM also show competitive performance, each achieving a mean balanced accuracy of 0.63 - 0.64 and mean HS recall around 0.43 - 0.44. Their ROC AUC values (0.72 - 0.73) further reflect strong overall discriminative capability. Notably, CatBoost has the highest maximum HS recall of 83%, though its wider standard deviation (± 0.20) indicates variable performance across training datasets.

Tree-based ensemble models such as RandomForest and ExtraTrees (both Gini and Entr variants) exhibit lower average HS recall (0.35) despite achieving good ROC AUC estimates (0.71). This suggests that while they are generally good at distinguishing classes, they may underperform in capturing severe crash instances.

Overall, **Table 7** reinforces earlier observations from **Figure 4** by showing that no single model dominates across all metrics, and high ROC AUC or balanced accuracy does not always imply strong HS recall. Neural network-based models, particularly NeuralNetFastAI and NeuralNetTorch, appear more robust in detecting HS crashes while maintaining overall performance.

### Comparison Between Feature Selection Techniques

**Figures 5, 6, and 7** illustrate the impact of different feature selection techniques, including Pearson's correlation, Feature Importance (FI) using RF, Recursive Feature Elimination (RFE) using LR, Chi-squared test (Chi2), Discriminative Mutual Information (DMI), and a combined set of top-ranked features, on the performance of the following models: NeuralNetFastAI, NeuralNetTorch, BML, CatBoost, LightGBM, XGBoost, ExtraTreesGini, and RandomForestEntr.

Balanced accuracy (**Figure 5**) and ROC AUC (**Figure 6**) show consistent and similar performance across all feature selection techniques, which indicates minimal sensitivity of these metrics to the choice of feature selection. However, considerable variation is observed for HS recall (**Figure 7**), where the DMI-based feature selection outperforms other individual techniques. In particular, CatBoost, NeuralNetTorch, XGBoost, LightGBM, and NeuralNetFastAI achieve substantially higher HS recall values using DMI.

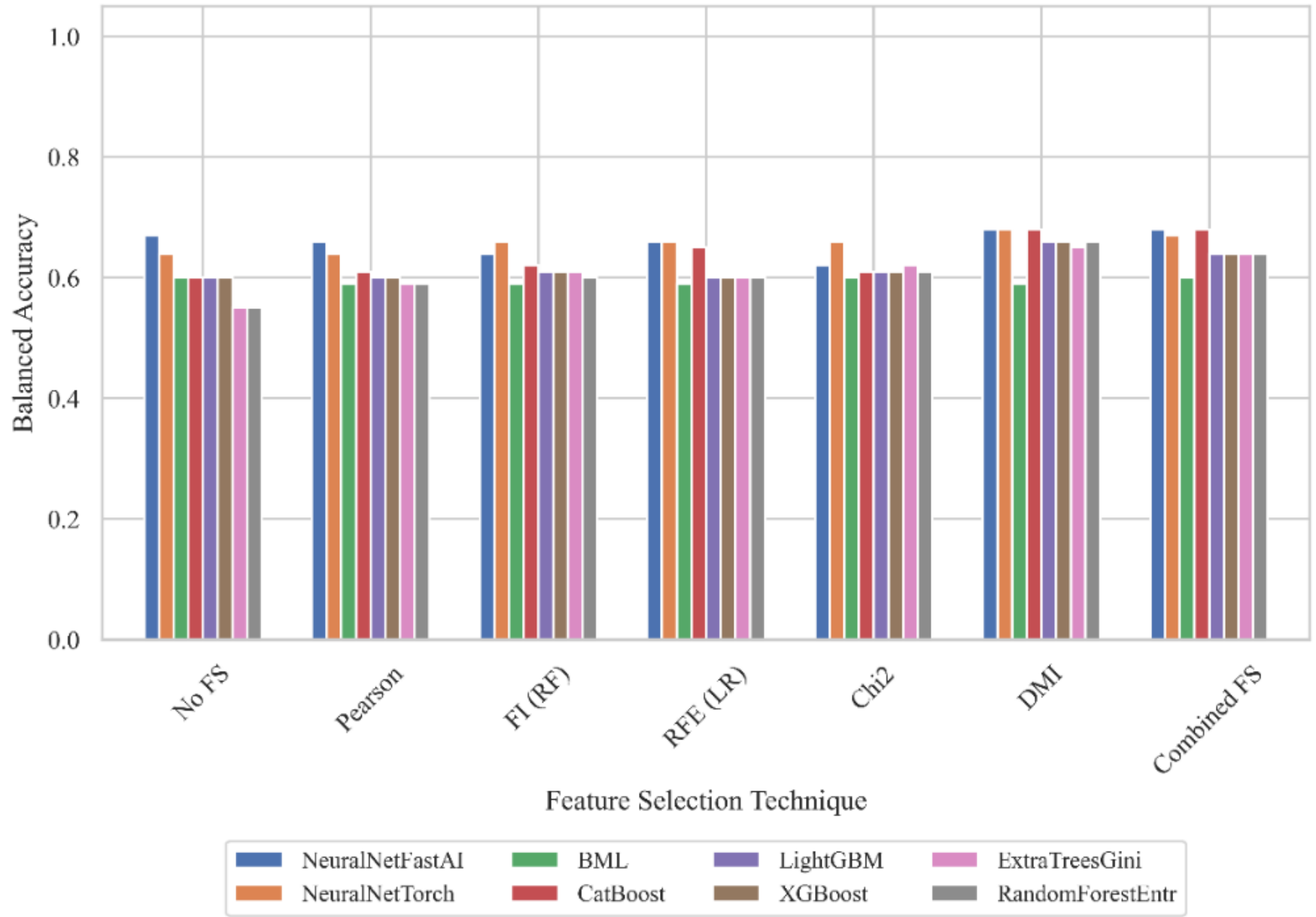

**Figure 5 Balanced Accuracy Comparison Across Feature Selection Techniques and Models**

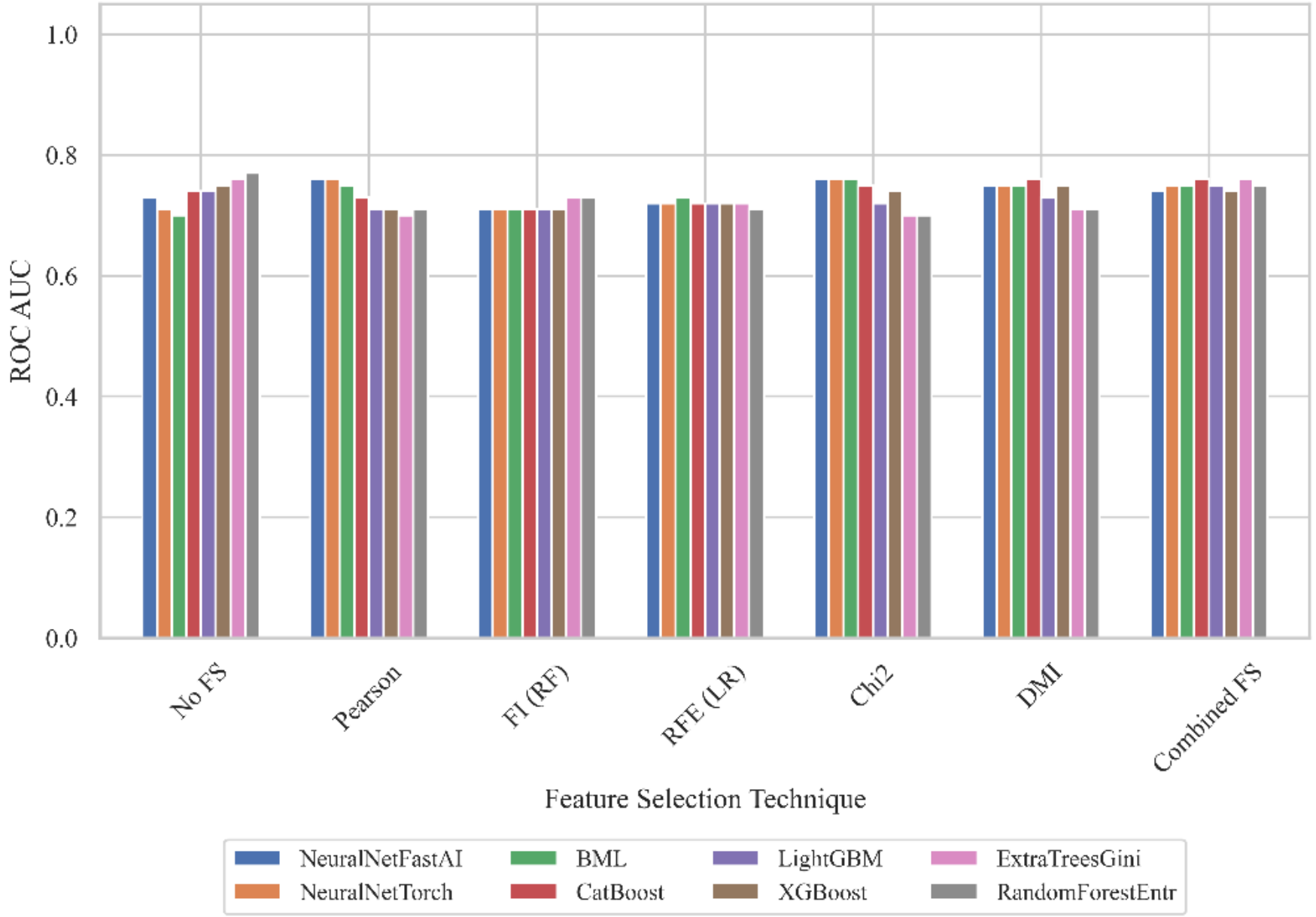

**Figure 6 ROC AUC Comparison Across Feature Selection Techniques and Models**

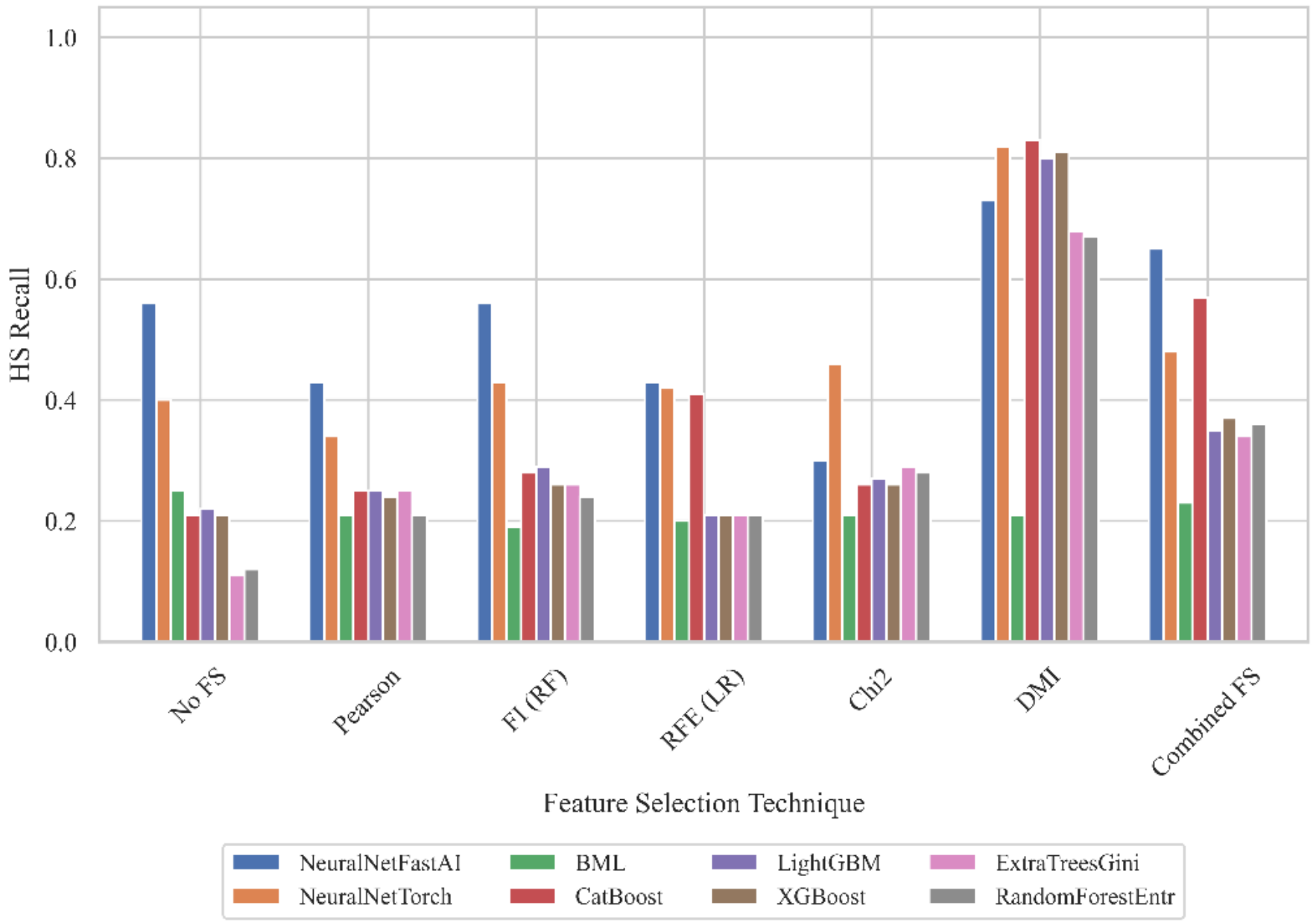

**Figure 7 HS Recall Comparison Across Feature Selection Techniques and Models**

The likely reason for the superior performance of DMI is its explicit focus on class discrimination. Unlike other techniques that rely primarily on linear relationships (e.g., Pearson's correlation or RFE using LR) or general statistical relevance (e.g., FI using RF or Chi2), DMI evaluates the discriminative power of features directly with respect to the target class. Moreover, DMI captures both linear and non-linear dependencies between features and the target labels, making it better suited for identifying complex, non-linear patterns associated with rare events such as severe crashes.

The merged feature selection approach (combining top-ranked features from all individual techniques) generally results in performance comparable to or slightly better than individual methods such as Pearson's correlation, FI using RF, RFE using LR, and Chi2. However, it does not surpass DMI alone. The inclusion of DMI-selected features within the merged set slightly boosts HS recall performance compared to most individual techniques (excluding DMI), suggesting that incorporating highly discriminative features from DMI enhances the combined feature set's capability in predicting severe crashes.

**Comparison Between Data Balancing Techniques**

**Figures 8, 9, and 10** illustrate the impact of different data balancing techniques, including class weighting (HS Weight=4), oversampling methods (SMOTE, KMeansSMOTE, ADASYN, RandomOverSampler [ROS], K-SMOTE, WGAN-GP, Conditional WGAN-GP [CWGAN-GP]), undersampling methods (NearMiss-1, RandomUnderSampler [RUS]), and combined sampling (ROS + RUS), on model performance (balanced accuracy, ROC AUC, and HS recall, respectively) using the merged feature set (124 features) derived from the top 50 features selected by each feature selection technique.

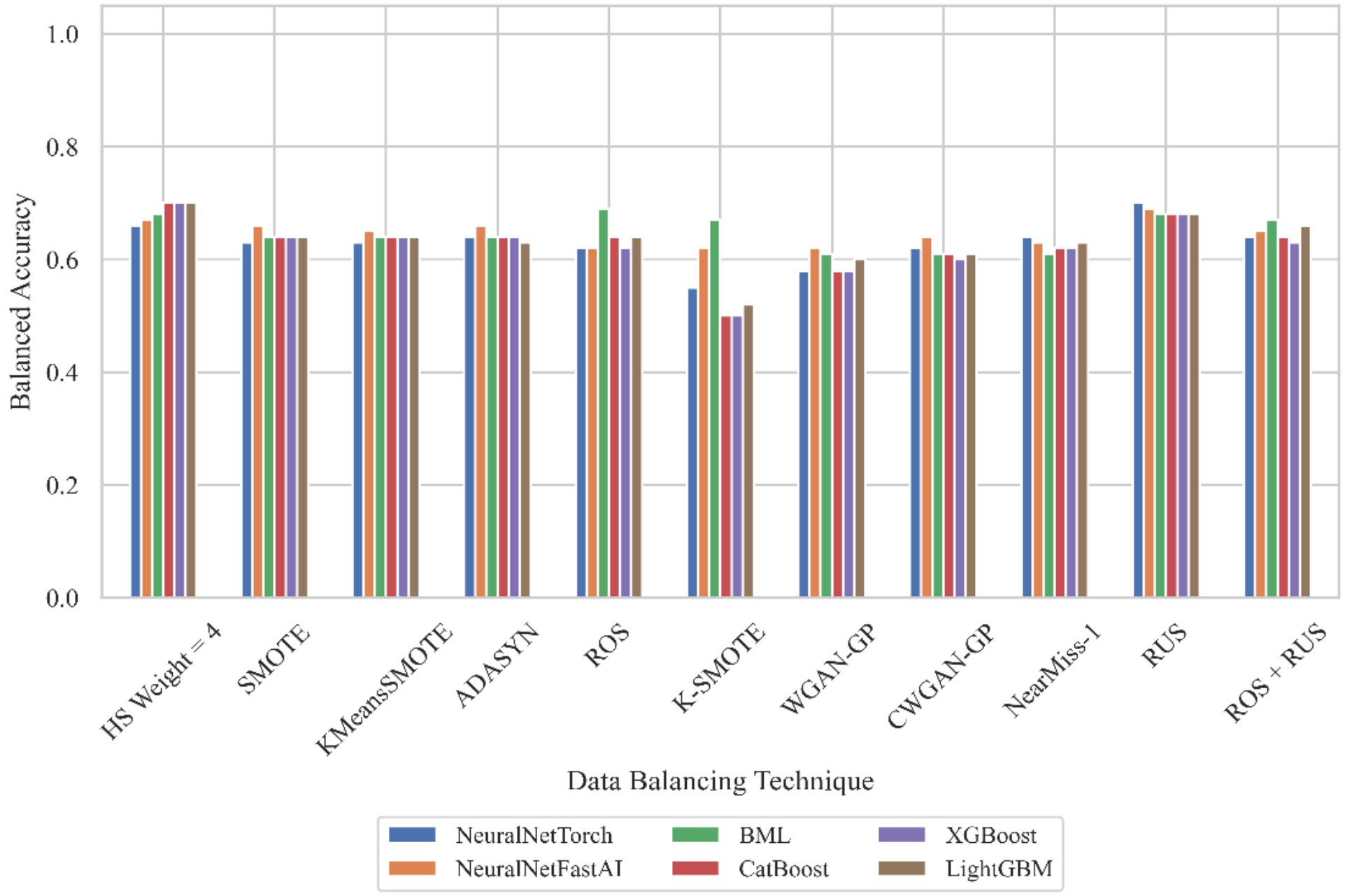

**Figure 8 Balanced Accuracy Comparison Across Data Balancing Techniques Using Merged Feature Set**

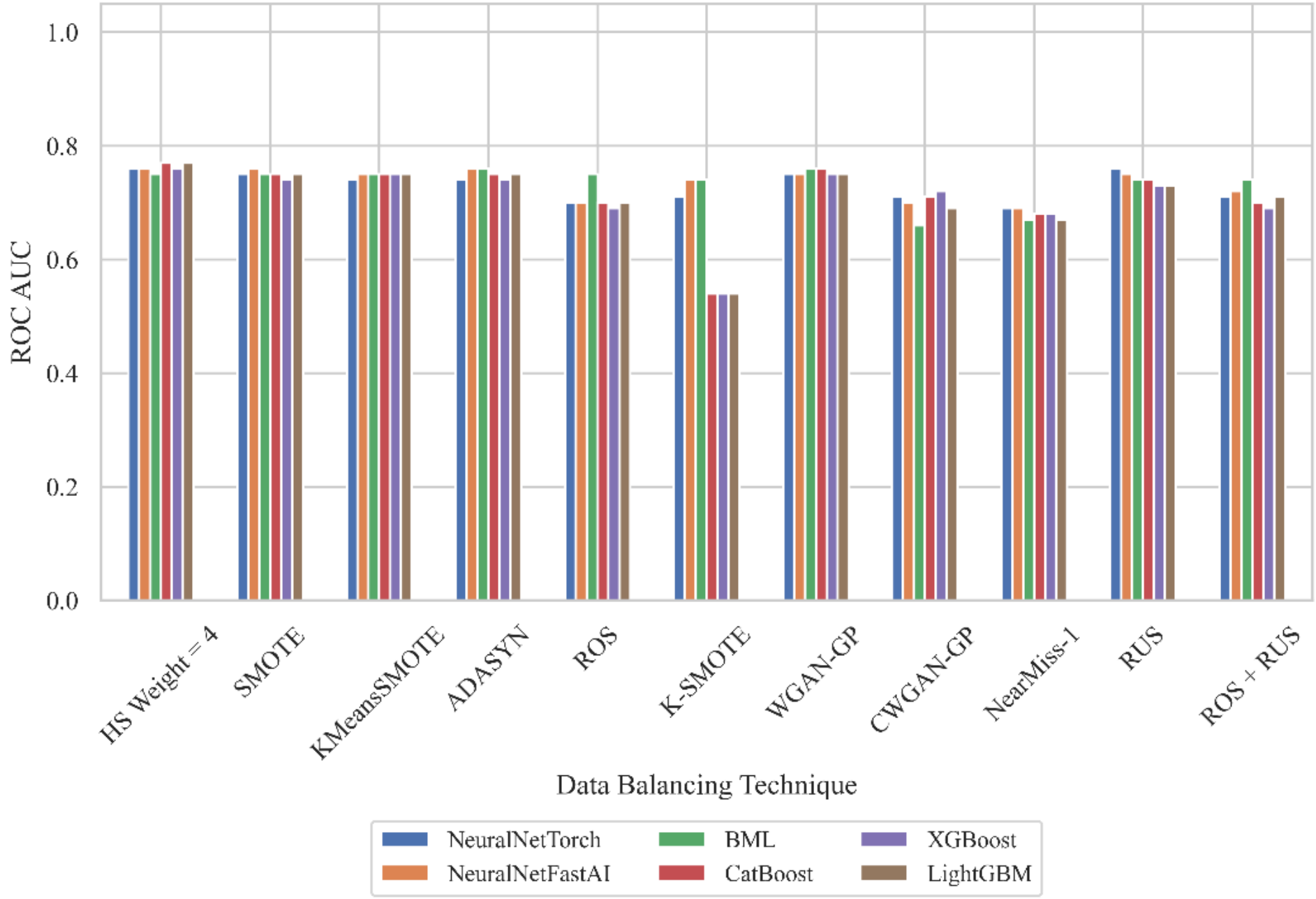

**Figure 9 ROC AUC Comparison Across Data Balancing Techniques Using Merged Feature Set**

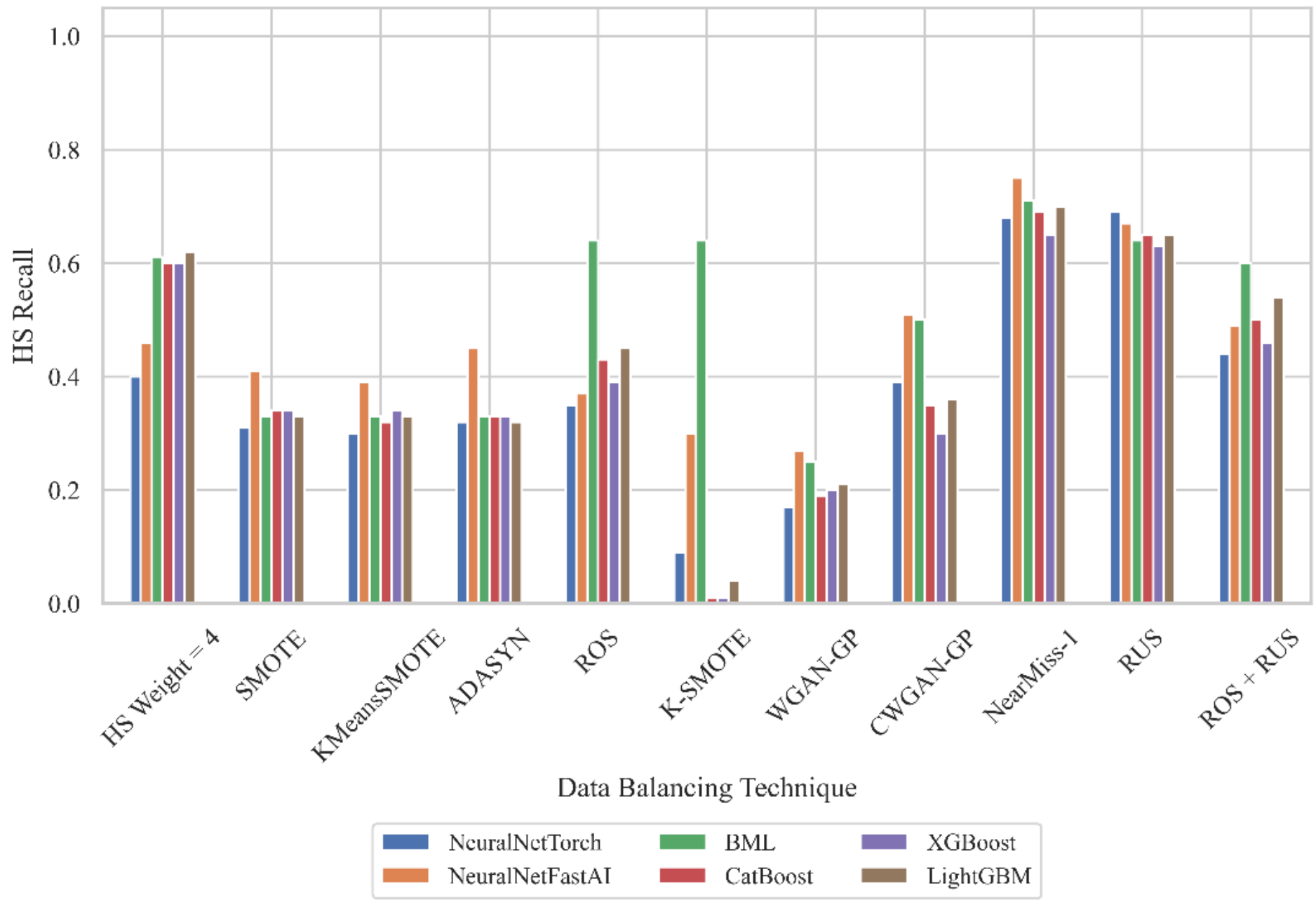

**Figure 10 HS Recall Comparison Across Data Balancing Techniques Using Merged Feature Set**

NearMiss-1 significantly enhances HS recall across most models, achieving the highest improvement among all techniques (**Figure 10**). However, this increase in HS recall is accompanied by relatively lower balanced accuracy and ROC AUC scores, indicating a trade-off between accurately detecting severe crashes and maintaining overall classification performance (**Figures 8 and 9**).

RUS provides balanced performance across all three metrics (i.e., balanced accuracy, ROC AUC, and HS recall), highlighting its ability to effectively address class imbalance without substantially compromising overall model quality. Class weighting (HS Weight=4) shows balanced performance across all metrics, performing slightly below RUS overall. It boosts HS recall while keeping balanced accuracy and ROC AUC stable. ROS and the combined method (ROS + RUS) offer slightly lower performance than RUS and class weighting but show effectiveness particularly with the statistical BML model.

Other oversampling methods, including SMOTE, KMeansSMOTE, ADASYN, K-SMOTE, WGAN-GP, and CWGAN-GP, generally demonstrate limited improvements in HS recall compared to simpler methods such as ROS and class weighting, suggesting minimal practical benefit in this context.

We further compare the best-performing data balancing techniques from the merged features analysis, i.e., NearMiss-1, RUS, HS Weight=4, ROS, and ROS+RUS, on datasets utilizing DMI feature selection to assess if DMI-selected features enhance HS recall performance further (**Figure 11**). Although NearMiss-1 shows an increase in HS recall compared to the merged feature set, data balancing techniques generally do not provide any improvements over models without any data balancing when applied specifically with DMI-selected features, with the BML model being the only exception. This indicates that the DMI-based feature selection already effectively captures the distinguishing characteristics of severe crashes.

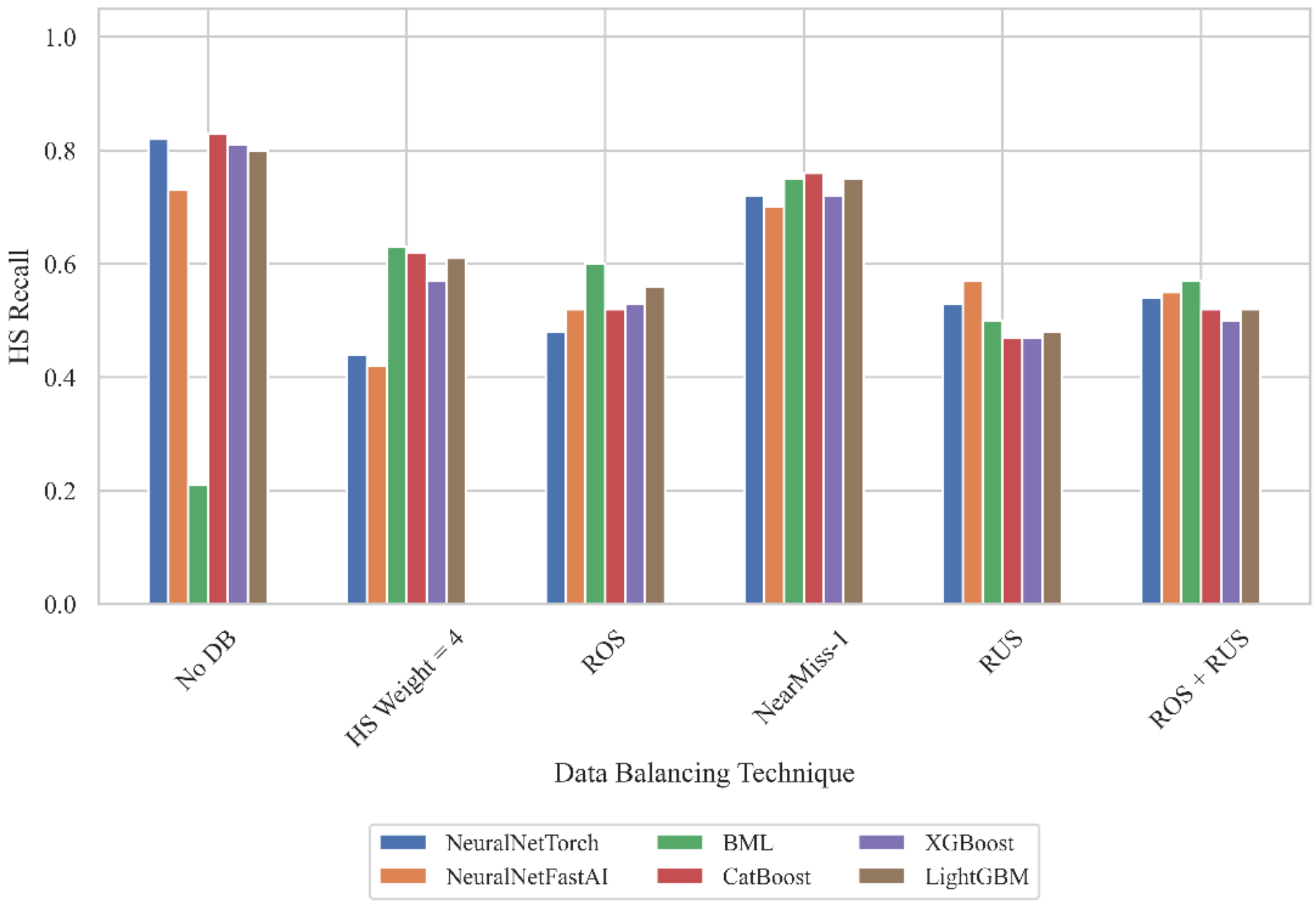

**Figure 11 HS Recall Comparison of Selected Data Balancing Techniques Using DMI Feature Set**

In summary, data balancing techniques such as RUS, class weighting (HS Weight=4), ROS, and the combined ROS + RUS method consistently demonstrate strong and balanced performance across all three key metrics (balanced accuracy, ROC AUC, and HS recall), making them the most robust choices for general applications. NearMiss-1 remains the most effective in maximizing HS recall but comes at the cost of reduced balanced accuracy and ROC AUC. Meanwhile, advanced oversampling methods such as SMOTE, KMeansSMOTE, ADASYN, and K-SMOTE perform well in terms of accuracy and overall robustness but offer limited gains in HS recall. When paired with DMI-based features, the need for data balancing is diminished, as DMI alone effectively enhances HS recall. Practitioners can use these comparative findings, supported by **Figures 8 to 11**, to align data balancing strategies with specific application goals, such as prioritizing HS crash detection, achieving balanced performance, or maintaining overall model stability.

**Best Performing Model-Training Dataset Combination**

The selection of the best-performing model-training dataset combinations depends on the specific application needs, particularly whether the priority is detection of critical HS (F+I) crashes or balanced performance between the HS and LS (PDO) classes. For HS detection optimization, we prioritize models that achieve high HS recall. By balanced performance, we refer to achieving strong and consistent results across balanced accuracy, ROC AUC, and HS recall, ensuring neither class is disproportionately prioritized in the model's predictions. **Tables 8 and 9** summarize the results for these two evaluation priorities, highlighting the most effective combinations of feature selection techniques, data balancing techniques, and model configurations based on these performance metrics.

**Table 8** highlights combinations of feature selection techniques, data balancing techniques, and models that prioritize the accurate detection of HS crashes. Notably, all combinations selected for optimal HS recall performance are based on features selected by Discriminative Mutual Information (DMI). As demonstrated earlier, DMI effectively captures discriminative patterns relevant to severe crashes, resulting in high sensitivity for HS incidents. The highest HS recall (83%) is achieved by the CatBoost model trained without any data balancing. Other strong-performing combinations without data balancing include NeuralNetTorch (82%), XGBoost (81%), and LightGBM (80%). When NearMiss-1 is applied to DMI-selected features, models such as LightGBMXT and LightGBMLarge achieve high HS recall (79% and 78%, respectively), though balanced accuracy and ROC AUC are significantly lower due to trade-offs associated with aggressive undersampling.

**TABLE 8 Selected Model, Feature Selection Technique, and Data Balancing Technique Combinations for Optimizing HS Detection**

| Feature Selection Technique | Data Balancing Technique | Model | Accuracy | LS Precision | HS Precision | LS Recall | HS Recall | LS F1 Score | HS F1 Score | Balanced Accuracy | ROC AUC |
|---|---|---|---|---|---|---|---|---|---|---|---|
| DMI | None | CatBoost | 0.58 | 0.92 | 0.32 | 0.52 | **0.83** | 0.66 | 0.46 | 0.68 | 0.76 |
| | | NeuralNetTorch | 0.60 | 0.92 | 0.33 | 0.54 | **0.82** | 0.68 | 0.47 | 0.68 | 0.75 |
| | | XGBoost | 0.58 | 0.91 | 0.31 | 0.52 | **0.81** | 0.67 | 0.45 | 0.66 | 0.75 |
| | | LightGBM | 0.59 | 0.91 | 0.31 | 0.53 | **0.80** | 0.67 | 0.45 | 0.66 | 0.73 |
| | NearMiss-1 | LightGBMXT | 0.39 | 0.83 | 0.23 | 0.28 | **0.79** | 0.41 | 0.35 | 0.54 | 0.59 |
| | | LightGBMLarge | 0.37 | 0.81 | 0.22 | 0.26 | **0.78** | 0.39 | 0.34 | 0.52 | 0.56 |

**Table 9** presents the model, feature selection, and data balancing combinations optimized for achieving balanced performance across both HS and LS crash predictions. In this context, balanced performance is defined by simultaneously attaining balanced accuracy ≥ 0.69, ROC AUC ≥ 0.75, and HS recall ≥ 0.60. These metrics collectively reflect the model's overall discriminative ability and its effectiveness in capturing minority class instances without compromising majority class performance. Most of the combinations meeting these criteria are based on the merged feature set composed of 124 features. However, the final three combinations listed in **Table 9** are generated from models trained without any feature selection technique applied, demonstrating that balanced performance is still achievable under the full feature space in certain configurations. Across these combinations, data balancing techniques such as RUS, HS class weighting (HS Weight=4), and ROS prove to be the most effective. These techniques, when used with models such as NeuralNetTorch, NeuralNetFastAI, LightGBM, LightGBMXT, CatBoost, and XGBoost, consistently result in high-performing models with balanced accuracy ranging from 0.69 to 0.70, ROC AUC between 0.75 and 0.77, and HS recall values from 0.60 to 0.69. These findings underscore the value of combining appropriate data balancing strategies with diverse or comprehensive feature sets to achieve stable and balanced crash severity prediction performance across both classes.

**TABLE 9 Selected Model and Data Balancing Technique Combinations for Balanced LS and HS Performance Metrics**

| Feature Selection Technique | Data Balancing Technique | Model | Accuracy | LS Precision | HS Precision | LS Recall | HS Recall | LS F1 Score | HS F1 Score | Balanced Accuracy | ROC AUC |
|---|---|---|---|---|---|---|---|---|---|---|---|
| Merging all feature selection techniques | RUS | NeuralNetTorch | 0.71 | 0.89 | 0.39 | **0.71** | **0.69** | 0.79 | 0.50 | **0.70** | **0.76** |
| | | NeuralNetFastAI | 0.70 | 0.89 | 0.38 | **0.71** | **0.67** | 0.79 | 0.49 | **0.69** | **0.75** |
| | HS Weight of 4 | LightGBM | 0.74 | 0.88 | 0.42 | **0.77** | **0.62** | 0.82 | 0.50 | **0.70** | **0.77** |
| | | LightGBMXT | 0.74 | 0.88 | 0.42 | **0.77** | **0.62** | 0.82 | 0.50 | **0.70** | **0.76** |
| | | CatBoost | 0.75 | 0.88 | 0.43 | **0.79** | **0.60** | 0.83 | 0.50 | **0.70** | **0.77** |
| | | XGBoost | 0.76 | 0.88 | 0.45 | **0.80** | **0.60** | 0.84 | 0.51 | **0.70** | **0.76** |
| | ROS | BML | 0.71 | 0.88 | 0.39 | **0.74** | **0.64** | 0.80 | 0.49 | **0.69** | **0.75** |
| None | HS Weight of 4 | LightGBM | 0.73 | 0.89 | 0.41 | **0.75** | **0.64** | 0.81 | 0.50 | **0.70** | **0.76** |
| | | CatBoost | 0.73 | 0.88 | 0.42 | **0.77** | **0.62** | 0.82 | 0.50 | **0.70** | **0.75** |
| | | XGBoost | 0.75 | 0.88 | 0.43 | **0.79** | **0.60** | 0.83 | 0.50 | **0.70** | **0.76** |

**Impact of Class Overlap in Crash Severity Prediction**

The relatively lower recall and F1-score for HS crashes indicate the model's difficulty in detecting all relevant HS (F+I) cases. This is attributed to the overlapping characteristics of LS (PDO) and HS incidents, where features indicative of HS crashes are also present in LS cases, potentially leading to misclassifications. This phenomenon of class overlap complicates the prediction process by making it difficult for the model to distinguish between classes that share similar feature characteristics. In our case, the model exhibits lower HS precision and higher LS precision, suggesting that many LS crashes are incorrectly classified as HS due to the presence of HS-indicative features. This leads to an increase in false positives for HS, reducing HS precision. However, this overlap does not necessarily improve HS recall, as recall depends on correctly identifying actual HS cases. The results indicate that our model encounters more instances of HS indicative features in LS crashes than vice versa, aligning with the findings of several researchers (*47*, *48*) who discuss imbalanced datasets and overlapping feature spaces that hinder accurate minority class predictions.

To investigate further, we conduct additional analysis using the balanced dataset generated after applying the NearMiss-1 undersampling technique (training dataset 17 in **Table 6**). NearMiss-1 reduces class overlap by strategically selecting majority class instances that are closest to the minority class instances in the feature space. This process effectively removes the majority class samples that are likely to be misclassified as the minority class, thereby reducing the overlap between the two classes and achieving a more balanced dataset (*32*). Therefore, unlike other analyses in this study where we separate the training and test datasets initially and then apply data balancing techniques to the training dataset, for this particular analysis, we select both training and test datasets from the balanced pool to maintain maximum class separation. This analysis results in a balanced accuracy of 76% and a ROC AUC of 0.83 with the BML model. For LS cases, the model achieves precision and recall scores of 74% and 80%, respectively, resulting in an F1-score of 77%. The HS cases yield a precision of 78%, a recall of 72%, and an F1-score of 75%, substantially improving the minority class performance metrics compared to those obtained from the best model-training dataset combinations presented in the previous subsection for balanced performance (**Table 9**).

**Real-World Applications and Scalability of Models**

The key findings from this study underscore the importance of addressing class imbalance and overlap in crash severity risk analysis, paving the way for more accurate predictions of HS crashes, which are critical for timely interventions and safety measures. By leveraging the predictive capabilities of the best-performing model-training dataset combinations outlined in **Tables 8 and 9**, stakeholders can gain actionable insights to enhance safety and operational efficiency in work zones. For instance, transportation agencies prioritizing the detection of HS crashes can adopt models utilizing DMI-selected features and data balancing techniques such as NearMiss-1, which excel in HS recall, thereby facilitating timely identification of high-risk incidents and enabling proactive safety measures.

Furthermore, for scenarios requiring balanced performance between detecting HS crashes and maintaining accuracy in LS predictions, models trained with merged features and data balancing techniques such as RandomUnderSampler (RUS) and HS-weighted approaches offer robust solutions. This approach can guide stakeholders in resource allocation, optimizing emergency response strategies, and planning work zone configurations. By implementing these tailored predictive models, transportation agencies can enhance decision-making processes, reduce crash severity, and ultimately improve road safety for large vehicles including CMVs, and other road users.

Additionally, these findings emphasize the scalability and adaptability of the proposed methodologies. While this study focuses on SC work zones, the models, techniques, and insights can be extended to other regions with appropriate calibration to local crash characteristics and data availability. Future efforts to standardize feature engineering and incorporate localized crash patterns will further enhance the transferability of these models, making them a valuable tool for improving road safety nationwide. This adaptability ensures that stakeholders across various jurisdictions can benefit from the study's outcomes.

**CONCLUSIONS**

This study aims to enhance the performance of crash severity risk models for incidents involving large vehicles (i.e., CMVs) in SC work-zone crashes by developing and comparing a range of statistical, machine learning, and deep learning models under varying feature selection and data balancing conditions. Our findings highlight the challenges associated with accurately predicting HS crashes due to class imbalance and feature overlap.

Among feature selection methods, Discriminative Mutual Information (DMI) is found to outperform other techniques, particularly in improving HS recall. DMI's ability to capture both linear and non-linear dependencies, along with its emphasis on class discrimination, contributes to its superior performance. While merging top-ranked features from multiple selection techniques results in slightly improved performance over all individual methods except DMI, models using DMI-selected features achieve superior high-severity crash detection. Conversely, for applications requiring balanced performance between HS and LS crash predictions, most top-performing combinations are derived from merged features or full feature sets without any feature selection, suggesting that comprehensive input spaces can be more effective for robust generalization across both classes.

With respect to data balancing, the NearMiss-1 technique significantly increases HS recall, especially when applied to DMI-selected features, but at the cost of reduced balanced accuracy and ROC AUC. RandomUnderSampler (RUS), HS class weighting, and RandomOverSampler (ROS) achieve strong and consistent performance across balanced accuracy, ROC AUC, and HS recall, offering potential solutions for balanced applications. It is to be noted that when DMI is used as the feature selection technique, the incremental benefits of applying data balancing techniques diminish, demonstrating that DMI alone is capable of isolating patterns relevant to HS crash detection.

In terms of model performance, several models show promising results. The CatBoost model trained with DMI-selected features achieves the highest HS recall (83%) without any data balancing. NeuralNetFastAI achieves the highest average balanced accuracy and HS recall across all model-training combinations, followed closely by NeuralNetTorch. Models such as LightGBM, LightGBMXT, and XGBoost also demonstrated strong balanced performance when paired with merged features and effective data balancing methods. The statistical Bayesian Mixed Logit (BML) model, known for its interpretability, shows improvements when paired with ROS and class weighting methods.

This study also underscores the impact of class overlap on accurately predicting HS crashes. By employing the NearMiss-1 undersampling technique to create a balanced dataset where both training and test sets are derived from the same balanced pool, we demonstrate the challenges posed by overlapping features between LS and HS incidents. The substantial improvement in HS performance metrics achieved with the balanced dataset highlights the necessity of addressing class overlap, in addition to class imbalance, to enhance model accuracy in predicting HS crashes.

Additionally, this study underscores the importance of aligning modeling strategies with specific application needs. For instance, stakeholders prioritizing the detection of HS crashes can benefit from DMI-based models and targeted undersampling methods such as NearMiss-1, while those requiring balanced performance should consider merged or full feature sets combined with RUS, HS weighting, or ROS. Furthermore, the methodologies developed in this study are scalable and adaptable to other regions, with appropriate calibration to local crash characteristics and data availability. Future efforts to standardize feature engineering and incorporate localized crash patterns will further enhance the transferability and practical utility of these models.

In this study, feature selection is applied prior to data balancing; however, future research should explore reversing this order to assess whether applying data balancing techniques first can further improve model performance. Future research should also explore incorporating external factors such as temporal changes in traffic volume and behavior, including surges in truck traffic driven by events such as the COVID-19 pandemic. Modeling such contextual influences can help capture emerging crash risks and identify dynamic patterns that might influence crash severity. Advanced approaches, such as time-series modeling, anomaly detection, or dynamic Bayesian models, could be particularly useful for capturing these effects and enhancing the predictive power of crash severity models.

To further enhance crash severity prediction models and address the persistent challenges of class imbalance and feature overlap, future research should prioritize advanced techniques such as feature engineering, cost-sensitive learning, and anomaly detection. Researchers should also focus on developing standardized methods for characterizing class overlap, exploring alternative overlap identification techniques, optimizing parameter estimation, and investigating hybrid approaches.

## ACKNOWLEDGMENTS

This work is based upon the work supported by the Federal Motor Carrier Safety Administration (FMCSA) (an agency in the USDOT). Any opinions, findings, conclusions, and recommendations expressed in this material are those of the author(s) and do not necessarily reflect the views of FMCSA, and the U.S. Government assumes no liability for the contents or use thereof.

Note that ChatGPT was used solely to check grammar and paraphrase texts. No information, text, figure, or table has been generated, nor has any kind of analysis been conducted using any Large Language Model or Generative Artificial Intelligence.

## CONFLICT OF INTEREST

The authors declare no potential conflicts of interest with respect to the research, authorship, and/or publication of this article.

## AUTHOR CONTRIBUTIONS

The authors confirm contribution to the paper as follows: study conception and design: A. Mamun, M. Chowdhury; data collection: J. Mwakalonge; analysis and interpretation of results: A. Mamun, A. Enan, D. Indah, G. Comert, J. Mwakalonge, M. Chowdhury; draft manuscript preparation: A. Mamun, M. Chowdhury, G. Comert, A. Enan, D. Indah, J. Mwakalonge. All authors reviewed the results and approved the final version of the manuscript.

## **Supplemental Table 1**

Crash Severity Distribution in South Carolina Work Zones (2014–2018) by Features

## Crash Severity Risk Modeling Strategies under Data Imbalance

*Supplemental Table 1: Crash Severity Distribution in South Carolina Work Zones (2014–2018) by Features*

| Feature Header | Feature Name | Category Code | Category Name | Low Severity (Property Damage Only) Count | Low Severity Frequency (%) | High Severity (Injury and Fatality) Count | High Severity Frequency (%) |
|---|---|---|---|---|---|---|---|
| UNT | Number of Units | 1 | One vehicle involved in crash | 263 | 6.24 | 68 | 6.00 |
| UNT | Number of Units | 2 | Two vehicles involved in crash | 3402 | 80.67 | 739 | 65.17 |
| UNT | Number of Units | 3 | Three vehicles involved in crash | 440 | 10.43 | 228 | 20.11 |
| UNT | Number of Units | 4 | Four vehicles involved in crash | 81 | 1.92 | 70 | 6.17 |
| UNT | Number of Units | 5 | Five vehicles involved in crash | 26 | 0.62 | 21 | 1.85 |
| UNT | Number of Units | 6 | Six vehicles involved in crash | 4 | 0.09 | 6 | 0.53 |
| UNT | Number of Units | 7 | Seven vehicles involved in crash | 0 | 0.00 | 1 | 0.09 |
| UNT | Number of Units | 8 | Eight vehicles involved in crash | 1 | 0.02 | 1 | 0.09 |
| DAY | Day of Week | 0 | Weekdays | 3572 | 84.70 | 952 | 83.95 |
| DAY | Day of Week | 1 | Weekend | 645 | 15.30 | 182 | 16.05 |
| RCT | Route Category | 1 | Interstate | 2452 | 58.15 | 524 | 46.21 |
| RCT | Route Category | 2 | United States (US) Primary | 562 | 13.33 | 226 | 19.93 |
| RCT | Route Category | 3 | South Carolina (SC) Primary | 579 | 13.73 | 212 | 18.69 |
| RCT | Route Category | 4 | Secondary Road | 469 | 11.12 | 151 | 13.32 |
| RCT | Route Category | 5 | County Road | 110 | 2.61 | 12 | 1.06 |
| RCT | Route Category | 7 | Ramp | 45 | 1.07 | 9 | 0.79 |
| RAI | Route Auxiliary | 0 | Main | 4111 | 97.49 | 1097 | 96.74 |
| RAI | Route Auxiliary | 2 | Alternate | 19 | 0.45 | 8 | 0.71 |
| RAI | Route Auxiliary | 5 | Spur | 3 | 0.07 | 2 | 0.18 |
| RAI | Route Auxiliary | 6 | Connection | 9 | 0.21 | 2 | 0.18 |
| RAI | Route Auxiliary | 7 | Business | 75 | 1.78 | 25 | 2.20 |
| LOA | Lane Number | 0 | Collision occurs in the center turning lane or on median crossover | 18 | 0.43 | 6 | 0.53 |
| LOA | Lane Number | 1 | Lane adjacent to the median | 2557 | 60.64 | 745 | 65.70 |
| LOA | Lane Number | 2 | Second lane from the median | 1136 | 26.94 | 279 | 24.60 |
| LOA | Lane Number | 3 | Third lane from the median | 461 | 10.93 | 91 | 8.02 |

*Supplemental Table 1: Crash Severity Distribution in South Carolina Work Zones (2014–2018) by Features*

| Feature Header | Feature Name | Category Code | Category Name | Low Severity (Property Damage Only) Count | Low Severity Frequency (%) | High Severity (Injury and Fatality) Count | High Severity Frequency (%) |
|---|---|---|---|---|---|---|---|
| LOA | Lane Number | 4 | Fourth lane from the median | 43 | 1.02 | 12 | 1.06 |
| LOA | Lane Number | 5 | Fifth lane from the median | 2 | 0.05 | 1 | 0.09 |
| DLR | Direction of Lane | 0 | North | 1259 | 29.86 | 320 | 28.22 |
| DLR | Direction of Lane | 1 | South | 1385 | 32.84 | 360 | 31.75 |
| DLR | Direction of Lane | 2 | East | 810 | 19.21 | 247 | 21.78 |
| DLR | Direction of Lane | 3 | West | 763 | 18.09 | 207 | 18.25 |
| ODR | Base Distance Direction | 0 | North | 1317 | 31.23 | 333 | 29.37 |
| ODR | Base Distance Direction | 1 | South | 1289 | 30.57 | 340 | 29.98 |
| ODR | Base Distance Direction | 2 | East | 823 | 19.52 | 225 | 19.84 |
| ODR | Base Distance Direction | 3 | West | 788 | 18.69 | 236 | 20.81 |
| ALC | Light Condition | 1 | Daylight | 3258 | 77.26 | 810 | 71.43 |
| ALC | Light Condition | 2 | Dawn | 63 | 1.49 | 12 | 1.06 |
| ALC | Light Condition | 3 | Dusk | 72 | 1.71 | 20 | 1.76 |
| ALC | Light Condition | 4 | Dark (Lighting Unspecified) | 137 | 3.25 | 42 | 3.70 |
| ALC | Light Condition | 5 | Dark (Street Lamp Lit) | 127 | 3.01 | 52 | 4.59 |
| ALC | Light Condition | 6 | Dark (Street Lamp Not Lit) | 32 | 0.76 | 21 | 1.85 |
| ALC | Light Condition | 7 | Dark (No Lights) | 528 | 12.52 | 177 | 15.61 |
| WCC | Weather Condition | 1 | Clear (No Adverse Conditions) | 3633 | 86.15 | 995 | 87.74 |
| WCC | Weather Condition | 2 | Rain | 300 | 7.11 | 74 | 6.53 |
| WCC | Weather Condition | 3 | Cloudy | 269 | 6.38 | 58 | 5.11 |
| WCC | Weather Condition | 5 | Snow | 2 | 0.05 | 2 | 0.18 |
| WCC | Weather Condition | 6 | Fog, Smog, Smoke | 10 | 0.24 | 4 | 0.35 |
| WCC | Weather Condition | 8 | Severe crosswinds | 1 | 0.02 | 1 | 0.09 |
| WCC | Weather Condition | 9 | Unknown | 2 | 0.05 | 0 | 0.00 |
| AHC | Road Character | 1 | Straight-Level | 3618 | 85.80 | 968 | 85.36 |
| AHC | Road Character | 2 | Straight-On grade | 432 | 10.24 | 100 | 8.82 |

*Supplemental Table 1: Crash Severity Distribution in South Carolina Work Zones (2014–2018) by Features*

| Feature Header | Feature Name | Category Code | Category Name | Low Severity (Property Damage Only) Count | Low Severity Frequency (%) | High Severity (Injury and Fatality) Count | High Severity Frequency (%) |
|---|---|---|---|---|---|---|---|
| AHC | Road Character | 3 | Straight-Hillcrest | 43 | 1.02 | 10 | 0.88 |
| AHC | Road Character | 4 | Curve-Level | 80 | 1.90 | 35 | 3.09 |
| AHC | Road Character | 5 | Curve-On grade | 43 | 1.02 | 20 | 1.76 |
| AHC | Road Character | 6 | Curve-Hillcrest | 1 | 0.02 | 1 | 0.09 |
| RSC | Road Surface Condition | 1 | Dry | 3785 | 89.76 | 1024 | 90.30 |
| RSC | Road Surface Condition | 2 | Wet | 413 | 9.79 | 103 | 9.08 |
| RSC | Road Surface Condition | 3 | Snow | 1 | 0.02 | 2 | 0.18 |
| RSC | Road Surface Condition | 4 | Slush | 3 | 0.07 | 0 | 0.00 |
| RSC | Road Surface Condition | 5 | Ice | 3 | 0.07 | 1 | 0.09 |
| RSC | Road Surface Condition | 6 | Contaminate | 0 | 0.00 | 1 | 0.09 |
| RSC | Road Surface Condition | 7 | Water (Standing) | 2 | 0.05 | 0 | 0.00 |
| RSC | Road Surface Condition | 8 | Other | 10 | 0.24 | 2 | 0.18 |
| RSC | Road Surface Condition | 9 | Unknown | 0 | 0.00 | 1 | 0.09 |
| FHE | First Harmful Event | 1 | Cargo/Equipment Loss or Shift | 8 | 0.19 | 1 | 0.09 |
| FHE | First Harmful Event | 2 | Cross Median/Center | 1 | 0.02 | 1 | 0.09 |
| FHE | First Harmful Event | 3 | Downhill Runaway | 1 | 0.02 | 0 | 0.00 |
| FHE | First Harmful Event | 4 | Equipment Failure | 10 | 0.24 | 0 | 0.00 |
| FHE | First Harmful Event | 6 | Immersion | 1 | 0.02 | 0 | 0.00 |
| FHE | First Harmful Event | 7 | Jackknife | 6 | 0.14 | 0 | 0.00 |
| FHE | First Harmful Event | 8 | Overturn/Rollover | 13 | 0.31 | 19 | 1.68 |
| FHE | First Harmful Event | 12 | Spill (Two-wheeled Vehicle) | 2 | 0.05 | 0 | 0.00 |
| FHE | First Harmful Event | 18 | Other Non-collision | 6 | 0.14 | 2 | 0.18 |
| FHE | First Harmful Event | 19 | Unknown Non-collision | 1 | 0.02 | 1 | 0.09 |
| FHE | First Harmful Event | 20 | Animal (Deer Only) | 6 | 0.14 | 0 | 0.00 |
| FHE | First Harmful Event | 21 | Animal (All Other Than Deer) | 6 | 0.14 | 1 | 0.09 |
| FHE | First Harmful Event | 22 | Motor Vehicle (In Transport) | 2360 | 55.96 | 610 | 53.79 |

*Supplemental Table 1: Crash Severity Distribution in South Carolina Work Zones (2014–2018) by Features*

| Feature Header | Feature Name | Category Code | Category Name | Low Severity (Property Damage Only) Count | Low Severity Frequency (%) | High Severity (Injury and Fatality) Count | High Severity Frequency (%) |
|---|---|---|---|---|---|---|---|
| FHE | First Harmful Event | 23 | Motor Vehicle (Stopped) | 1286 | 30.50 | 360 | 31.75 |
| FHE | First Harmful Event | 24 | Motor Vehicle (Other Roadway) | 1 | 0.02 | 0 | 0.00 |
| FHE | First Harmful Event | 25 | Motor Vehicle (Parked) | 106 | 2.51 | 22 | 1.94 |
| FHE | First Harmful Event | 26 | Pedalcycle | 0 | 0.00 | 2 | 0.18 |
| FHE | First Harmful Event | 27 | Pedestrian | 4 | 0.09 | 21 | 1.85 |
| FHE | First Harmful Event | 28 | Railway Vehicle | 0 | 0.00 | 1 | 0.09 |
| FHE | First Harmful Event | 29 | Moving Work Zone Maintenance Equipment | 57 | 1.35 | 12 | 1.06 |
| FHE | First Harmful Event | 38 | Other Movable Object | 60 | 1.42 | 9 | 0.79 |
| FHE | First Harmful Event | 39 | Unknown Movable Object | 2 | 0.05 | 0 | 0.00 |
| FHE | First Harmful Event | 40 | Bridge Overhead Structure | 2 | 0.05 | 0 | 0.00 |
| FHE | First Harmful Event | 43 | Bridge Rail | 3 | 0.07 | 0 | 0.00 |
| FHE | First Harmful Event | 45 | Curb | 2 | 0.05 | 1 | 0.09 |
| FHE | First Harmful Event | 46 | Ditch | 29 | 0.69 | 9 | 0.79 |
| FHE | First Harmful Event | 47 | Embankment | 8 | 0.19 | 4 | 0.35 |
| FHE | First Harmful Event | 48 | Fixed Equipment | 1 | 0.02 | 0 | 0.00 |
| FHE | First Harmful Event | 49 | Fence | 3 | 0.07 | 0 | 0.00 |
| FHE | First Harmful Event | 50 | Guardrail End | 22 | 0.52 | 4 | 0.35 |
| FHE | First Harmful Event | 51 | Guardrail Face | 31 | 0.74 | 10 | 0.88 |
| FHE | First Harmful Event | 52 | Highway Traffic Sign Post | 20 | 0.47 | 4 | 0.35 |
| FHE | First Harmful Event | 53 | Impact Attenuator/Crash Cushion | 6 | 0.14 | 1 | 0.09 |
| FHE | First Harmful Event | 54 | Light/Luminaire Support | 1 | 0.02 | 0 | 0.00 |
| FHE | First Harmful Event | 55 | Mail Box | 1 | 0.02 | 2 | 0.18 |
| FHE | First Harmful Event | 56 | Median Barrier | 64 | 1.52 | 19 | 1.68 |
| FHE | First Harmful Event | 57 | Overhead Sign Post | 3 | 0.07 | 0 | 0.00 |

*Supplemental Table 1: Crash Severity Distribution in South Carolina Work Zones (2014–2018) by Features*

| Feature Header | Feature Name | Category Code | Category Name | Low Severity (Property Damage Only) Count | Low Severity Frequency (%) | High Severity (Injury and Fatality) Count | High Severity Frequency (%) |
|---|---|---|---|---|---|---|---|
| FHE | First Harmful Event | 58 | Other Fixed Object (Post, Pole, Support, etc.) | 10 | 0.24 | 3 | 0.26 |
| FHE | First Harmful Event | 59 | Other Fixed Object (Wall, Building, Tunnel, etc.) | 7 | 0.17 | 1 | 0.09 |
| FHE | First Harmful Event | 60 | Tree | 16 | 0.38 | 8 | 0.71 |
| FHE | First Harmful Event | 61 | Utility Poll | 4 | 0.09 | 0 | 0.00 |
| FHE | First Harmful Event | 62 | Fixed Work Zone Maintenance Equipment | 30 | 0.71 | 5 | 0.44 |
| FHE | First Harmful Event | 68 | Other Fixed Object | 17 | 0.40 | 1 | 0.09 |
| HEL | Harmful Event Location | 1 | Gore | 1 | 0.02 | 0 | 0.00 |
| HEL | Harmful Event Location | 2 | Island | 1 | 0.02 | 1 | 0.09 |
| HEL | Harmful Event Location | 3 | Median | 74 | 1.75 | 22 | 1.94 |
| HEL | Harmful Event Location | 4 | Roadside | 123 | 2.92 | 37 | 3.26 |
| HEL | Harmful Event Location | 5 | Roadway | 3869 | 91.75 | 1032 | 91.01 |
| HEL | Harmful Event Location | 6 | Shoulder | 80 | 1.90 | 24 | 2.12 |
| HEL | Harmful Event Location | 7 | Sidewalk | 1 | 0.02 | 1 | 0.09 |
| HEL | Harmful Event Location | 8 | Outside Trafficway | 67 | 1.59 | 14 | 1.23 |
| HEL | Harmful Event Location | 9 | Unknown | 1 | 0.02 | 3 | 0.26 |
| XWK | Presence of Crosswalk | 1 | Yes | 27 | 0.64 | 14 | 1.23 |
| XWK | Presence of Crosswalk | 2 | No | 4184 | 99.22 | 1117 | 98.50 |
| XWK | Presence of Crosswalk | 9 | Unknown | 6 | 0.14 | 3 | 0.26 |
| PRC | Primary Contributing Factor | 1 | Dirver Disregarded Signs, Signals, etc. | 47 | 1.11 | 24 | 2.12 |
| PRC | Primary Contributing Factor | 2 | Driver Distracted/Inattention | 202 | 4.79 | 38 | 3.35 |
| PRC | Primary Contributing Factor | 3 | Driving Too Fast for Conditions | 1733 | 41.10 | 556 | 49.03 |
| PRC | Primary Contributing Factor | 4 | Exceeded Authorized Speed Limit | 5 | 0.12 | 4 | 0.35 |
| PRC | Primary Contributing Factor | 5 | Driver Failed to Yield Right of Way | 325 | 7.71 | 120 | 10.58 |

*Supplemental Table 1: Crash Severity Distribution in South Carolina Work Zones (2014–2018) by Features*

| Feature Header | Feature Name | Category Code | Category Name | Low Severity (Property Damage Only) Count | Low Severity Frequency (%) | High Severity (Injury and Fatality) Count | High Severity Frequency (%) |
|---|---|---|---|---|---|---|---|
| PRC | Primary Contributing Factor | 6 | Ran off Road | 19 | 0.45 | 4 | 0.35 |
| PRC | Primary Contributing Factor | 7 | Driver Fatigued/Asleep | 9 | 0.21 | 7 | 0.62 |
| PRC | Primary Contributing Factor | 8 | Driver Followed Too Closely | 212 | 5.03 | 56 | 4.94 |
| PRC | Primary Contributing Factor | 9 | Driver Made an Improper Turn | 59 | 1.40 | 12 | 1.06 |
| PRC | Primary Contributing Factor | 10 | Driver Medical Related | 9 | 0.21 | 13 | 1.15 |
| PRC | Primary Contributing Factor | 12 | Aggressive Operation of Vehicle by Driver | 16 | 0.38 | 16 | 1.41 |
| PRC | Primary Contributing Factor | 13 | Driver Over-correcting/Over-steering | 2 | 0.05 | 2 | 0.18 |
| PRC | Primary Contributing Factor | 14 | Driver Swerving to Avoiding Object | 1 | 0.02 | 0 | 0.00 |
| PRC | Primary Contributing Factor | 15 | Driving Wrong Side or Wrong Way | 49 | 1.16 | 25 | 2.20 |
| PRC | Primary Contributing Factor | 16 | Driver Under the Influence | 73 | 1.73 | 55 | 4.85 |
| PRC | Primary Contributing Factor | 17 | Driver Vision Obscured | 8 | 0.19 | 0 | 0.00 |
| PRC | Primary Contributing Factor | 18 | Improper Lane Usage/Change | 866 | 20.54 | 98 | 8.64 |
| PRC | Primary Contributing Factor | 19 | Driver On Cell Phone | 1 | 0.02 | 0 | 0.00 |
| PRC | Primary Contributing Factor | 20 | Driver Texting | 2 | 0.05 | 0 | 0.00 |
| PRC | Primary Contributing Factor | 28 | Other Improper Action by Driver | 308 | 7.30 | 43 | 3.79 |
| PRC | Primary Contributing Factor | 29 | Unknown Driver Factors | 127 | 3.01 | 23 | 2.03 |
| PRC | Primary Contributing Factor | 30 | Debris on Roadway | 7 | 0.17 | 2 | 0.18 |
| PRC | Primary Contributing Factor | 31 | Non-highway Work | 2 | 0.05 | 0 | 0.00 |
| PRC | Primary Contributing Factor | 32 | Obstruction in Roadway | 5 | 0.12 | 2 | 0.18 |
| PRC | Primary Contributing Factor | 33 | Road Surface Condition (i.e., Wet) | 2 | 0.05 | 0 | 0.00 |
| PRC | Primary Contributing Factor | 34 | Rut, Holes, Bumps on Roadway | 3 | 0.07 | 1 | 0.09 |
| PRC | Primary Contributing Factor | 36 | Traffic Control Device (i.e., Missing) | 1 | 0.02 | 1 | 0.09 |
| PRC | Primary Contributing Factor | 37 | Work Zone (Construction/Maintenance/Utility) | 21 | 0.50 | 5 | 0.44 |
| PRC | Primary Contributing Factor | 48 | Other Roadway Factors | 4 | 0.09 | 0 | 0.00 |

*Supplemental Table 1: Crash Severity Distribution in South Carolina Work Zones (2014–2018) by Features*

| Feature Header | Feature Name | Category Code | Category Name | Low Severity (Property Damage Only) Count | Low Severity Frequency (%) | High Severity (Injury and Fatality) Count | High Severity Frequency (%) |
|---|---|---|---|---|---|---|---|
| PRC | Primary Contributing Factor | 50 | Non-motorist Inattentive | 2 | 0.05 | 0 | 0.00 |
| PRC | Primary Contributing Factor | 51 | Non-motorist Lying and/or Illegally in Roadway | 1 | 0.02 | 3 | 0.26 |
| PRC | Primary Contributing Factor | 53 | Non-motorist Not Visible (Dark Clothing) | 0 | 0.00 | 1 | 0.09 |
| PRC | Primary Contributing Factor | 55 | Non-motorist Improper Crossing | 0 | 0.00 | 2 | 0.18 |
| PRC | Primary Contributing Factor | 58 | Other Non-motorist Involvements | 0 | 0.00 | 1 | 0.09 |
| PRC | Primary Contributing Factor | 60 | Animal in Road | 5 | 0.12 | 1 | 0.09 |
| PRC | Primary Contributing Factor | 61 | Glare | 1 | 0.02 | 0 | 0.00 |
| PRC | Primary Contributing Factor | 62 | Environmental Obstruction | 1 | 0.02 | 1 | 0.09 |
| PRC | Primary Contributing Factor | 63 | Weather Condition | 0 | 0.00 | 3 | 0.26 |
| PRC | Primary Contributing Factor | 66 | Non-motorist Under the Influence | 1 | 0.02 | 0 | 0.00 |
| PRC | Primary Contributing Factor | 68 | Other Environmental Factors | 1 | 0.02 | 0 | 0.00 |
| PRC | Primary Contributing Factor | 70 | Brakes Defect | 12 | 0.28 | 4 | 0.35 |
| PRC | Primary Contributing Factor | 71 | Steering Defect | 8 | 0.19 | 1 | 0.09 |
| PRC | Primary Contributing Factor | 72 | Vehicle Power Plant Defect | 1 | 0.02 | 0 | 0.00 |
| PRC | Primary Contributing Factor | 73 | Tires/Wheel Defect | 26 | 0.62 | 8 | 0.71 |
| PRC | Primary Contributing Factor | 78 | Truck Coupling Defect | 3 | 0.07 | 0 | 0.00 |
| PRC | Primary Contributing Factor | 79 | Cargo Defect | 28 | 0.66 | 2 | 0.18 |
| PRC | Primary Contributing Factor | 88 | Other Vehicle Defect | 6 | 0.14 | 0 | 0.00 |
| PRC | Primary Contributing Factor | 89 | Unknown Vehicle Defect | 3 | 0.07 | 0 | 0.00 |
| TIM | Time of Collision | 0 | 00:00 - 06:00 | 239 | 5.67 | 103 | 9.08 |
| TIM | Time of Collision | 1 | 06:00 - 10:00 | 826 | 19.59 | 228 | 20.11 |
| TIM | Time of Collision | 2 | 10:00 - 15:00 | 1371 | 32.51 | 340 | 29.98 |
| TIM | Time of Collision | 3 | 15:00 - 19:00 | 1150 | 27.27 | 283 | 24.96 |
| TIM | Time of Collision | 4 | 19:00 - 00:00 | 631 | 14.96 | 180 | 15.87 |

*Supplemental Table 1: Crash Severity Distribution in South Carolina Work Zones (2014–2018) by Features*

| Feature Header | Feature Name | Category Code | Category Name | Low Severity (Property Damage Only) Count | Low Severity Frequency (%) | High Severity (Injury and Fatality) Count | High Severity Frequency (%) |
|---|---|---|---|---|---|---|---|
| BDO (in Miles) | Base Distance Offset | 0 | 0 - 30 | 2857 | 67.75 | 782 | 68.96 |
| BDO | Base Distance Offset | 1 | 30 - 60 | 499 | 11.83 | 118 | 10.41 |
| BDO | Base Distance Offset | 2 | 60 - 90 | 192 | 4.55 | 49 | 4.32 |
| BDO | Base Distance Offset | 3 | 90 - 120 | 348 | 8.25 | 87 | 7.67 |
| BDO | Base Distance Offset | 4 | 120 - 150 | 69 | 1.64 | 25 | 2.20 |
| BDO | Base Distance Offset | 5 | 150 - 180 | 46 | 1.09 | 9 | 0.79 |
| BDO | Base Distance Offset | 6 | 180 - 210 | 85 | 2.02 | 26 | 2.29 |
| BDO | Base Distance Offset | 7 | 210 - 240 | 22 | 0.52 | 5 | 0.44 |
| BDO | Base Distance Offset | 8 | 240 - 270 | 10 | 0.24 | 3 | 0.26 |
| BDO | Base Distance Offset | 9 | 270 - 300 | 32 | 0.76 | 9 | 0.79 |
| BDO | Base Distance Offset | 10 | 300 - 350 | 15 | 0.36 | 4 | 0.35 |
| BDO | Base Distance Offset | 11 | 350 - 400 | 18 | 0.43 | 8 | 0.71 |
| BDO | Base Distance Offset | 12 | 400 - 450 | 2 | 0.05 | 2 | 0.18 |
| BDO | Base Distance Offset | 13 | 450 - 500 | 9 | 0.21 | 4 | 0.35 |
| BDO | Base Distance Offset | 14 | 500 - 700 | 6 | 0.14 | 3 | 0.26 |
| BDO | Base Distance Offset | 15 | 700 - 900 | 4 | 0.09 | 0 | 0.00 |
| BDO | Base Distance Offset | 16 | 900 - 1000 | 3 | 0.07 | 0 | 0.00 |
| BIR | Base Route Category | 1 | Interstate | 307 | 7.28 | 55 | 4.85 |
| BIR | Base Route Category | 2 | US Primary | 512 | 12.14 | 110 | 9.70 |
| BIR | Base Route Category | 3 | SC Primary | 830 | 19.68 | 182 | 16.05 |
| BIR | Base Route Category | 4 | Secondary Road | 1651 | 39.15 | 511 | 45.06 |
| BIR | Base Route Category | 5 | County Road | 735 | 17.43 | 238 | 20.99 |
| BIR | Base Route Category | 6 | Other | 18 | 0.43 | 5 | 0.44 |
| BIR | Base Route Category | 7 | Ramp | 164 | 3.89 | 33 | 2.91 |

*Supplemental Table 1: Crash Severity Distribution in South Carolina Work Zones (2014–2018) by Features*

| Feature Header | Feature Name | Category Code | Category Name | Low Severity (Property Damage Only) Count | Low Severity Frequency (%) | High Severity (Injury and Fatality) Count | High Severity Frequency (%) |
|---|---|---|---|---|---|---|---|
| SIC | Second Intersection Route Category | 1 | Interstate | 235 | 5.57 | 37 | 3.26 |
| SIC | Second Intersection Route Category | 2 | US Primary | 400 | 9.49 | 104 | 9.17 |
| SIC | Second Intersection Route Category | 3 | SC Primary | 517 | 12.26 | 111 | 9.79 |
| SIC | Second Intersection Route Category | 4 | Secondary Road | 1760 | 41.74 | 487 | 42.95 |
| SIC | Second Intersection Route Category | 5 | County Road | 971 | 23.03 | 308 | 27.16 |
| SIC | Second Intersection Route Category | 6 | Other | 191 | 4.53 | 51 | 4.50 |
| SIC | Second Intersection Route Category | 7 | Ramp | 143 | 3.39 | 36 | 3.17 |
| MAC | Manner of Collision | 0 | Not Collision with Motor Vehicle | 436 | 10.34 | 117 | 10.32 |
| MAC | Manner of Collision | 10 | Rear End | 2034 | 48.23 | 656 | 57.85 |
| MAC | Manner of Collision | 20 | Head On | 34 | 0.81 | 28 | 2.47 |
| MAC | Manner of Collision | 30 | Rear-to-Rear | 5 | 0.12 | 1 | 0.09 |
| MAC | Manner of Collision | 41 | Angle (Southeast/Southwest) | 109 | 2.58 | 38 | 3.35 |
| MAC | Manner of Collision | 42 | Angle (East/West) | 165 | 3.91 | 94 | 8.29 |
| MAC | Manner of Collision | 43 | Angle (Northeast/Northwest) | 284 | 6.73 | 77 | 6.79 |
| MAC | Manner of Collision | 50 | Sideswipe Same Direction | 921 | 21.84 | 91 | 8.02 |
| MAC | Manner of Collision | 60 | Sideswipe Opposite Direction | 86 | 2.04 | 22 | 1.94 |
| MAC | Manner of Collision | 70 | Backed Into | 132 | 3.13 | 6 | 0.53 |
| MAC | Manner of Collision | 99 | Unknown (Hit and Run Only) | 11 | 0.26 | 4 | 0.35 |
| JCT | Junction Type | 1 | Crossover | 14 | 0.33 | 1 | 0.09 |

*Supplemental Table 1: Crash Severity Distribution in South Carolina Work Zones (2014–2018) by Features*

| Feature Header | Feature Name | Category Code | Category Name | Low Severity (Property Damage Only) Count | Low Severity Frequency (%) | High Severity (Injury and Fatality) Count | High Severity Frequency (%) |
|---|---|---|---|---|---|---|---|
| JCT | Junction Type | 2 | Driveway | 109 | 2.58 | 51 | 4.50 |
| JCT | Junction Type | 3 | Five/More Points | 3 | 0.07 | 1 | 0.09 |
| JCT | Junction Type | 4 | Four-way Intersection | 234 | 5.55 | 88 | 7.76 |
| JCT | Junction Type | 5 | Railway Grade Crossing | 0 | 0.00 | 1 | 0.09 |
| JCT | Junction Type | 7 | Shared Use Paths or Trails | 4 | 0.09 | 2 | 0.18 |
| JCT | Junction Type | 8 | T-Intersection | 199 | 4.72 | 76 | 6.70 |
| JCT | Junction Type | 9 | Traffic Circle | 1 | 0.02 | 1 | 0.09 |
| JCT | Junction Type | 12 | Y-Intersection | 84 | 1.99 | 19 | 1.68 |
| JCT | Junction Type | 13 | Non-junction | 3565 | 84.54 | 893 | 78.75 |
| JCT | Junction Type | 99 | Unknown | 4 | 0.09 | 1 | 0.09 |
| IBUS | School Bus Involved | 1 | Yes, Directly | 34 | 0.81 | 5 | 0.44 |
| IBUS | School Bus Involved | 2 | Yes, Indirectly | 27 | 0.64 | 4 | 0.35 |
| IBUS | School Bus Involved | 3 | No | 4155 | 98.53 | 1125 | 99.21 |
| IBUS | School Bus Involved | 9 | Unknown | 1 | 0.02 | 0 | 0.00 |
| TWAY | Trafficway | 1 | Two-way, Not Divided | 1091 | 25.87 | 392 | 34.57 |
| TWAY | Trafficway | 2 | Two-way, Divided, Unprotected Median | 612 | 14.51 | 190 | 16.75 |
| TWAY | Trafficway | 3 | Two-way, Divided, Barrier | 2369 | 56.18 | 528 | 46.56 |
| TWAY | Trafficway | 4 | One-way | 134 | 3.18 | 23 | 2.03 |
| TWAY | Trafficway | 8 | Other | 11 | 0.26 | 1 | 0.09 |
| WZT | Work Zone Type | 1 | Shoulder/Median Work | 2021 | 47.93 | 554 | 48.85 |
| WZT | Work Zone Type | 2 | Lane Shift/Crossover | 427 | 10.13 | 87 | 7.67 |
| WZT | Work Zone Type | 3 | Intermittent/Moving Work | 261 | 6.19 | 99 | 8.73 |
| WZT | Work Zone Type | 4 | Lane Closure | 1239 | 29.38 | 303 | 26.72 |
| WZT | Work Zone Type | 8 | Other | 242 | 5.74 | 86 | 7.58 |
| WZT | Work Zone Type | 9 | Unknown | 27 | 0.64 | 5 | 0.44 |
| WZL | Work Zone Location | 1 | Before 1st Sign | 171 | 4.06 | 54 | 4.76 |

## Crash Severity Risk Modeling Strategies under Data Imbalance

*Supplemental Table 1: Crash Severity Distribution in South Carolina Work Zones (2014–2018) by Features*

| Feature Header | Feature Name | Category Code | Category Name | Low Severity (Property Damage Only) Count | Low Severity Frequency (%) | High Severity (Injury and Fatality) Count | High Severity Frequency (%) |
|---|---|---|---|---|---|---|---|
| WZL | Work Zone Location | 2 | Advanced Warning Area | 499 | 11.83 | 136 | 11.99 |
| WZL | Work Zone Location | 3 | Transition Area | 729 | 17.29 | 143 | 12.61 |
| WZL | Work Zone Location | 4 | Activity Area | 2705 | 64.15 | 767 | 67.64 |
| WZL | Work Zone Location | 5 | Termination Area | 113 | 2.68 | 34 | 3.00 |
| WPR | Workers Present | 1 | Yes | 2038 | 48.33 | 551 | 48.59 |
| WPR | Workers Present | 2 | No | 2179 | 51.67 | 583 | 51.41 |
| TCT | Traffic Control Type | 1 | Stop and Go Light | 205 | 4.86 | 70 | 6.17 |
| TCT | Traffic Control Type | 2 | Flashing Traffic Signal | 11 | 0.26 | 3 | 0.26 |
| TCT | Traffic Control Type | 11 | RR (Crossbucks, Lights, and Gates) | 2 | 0.05 | 1 | 0.09 |
| TCT | Traffic Control Type | 13 | Rail Road (RR) Crossbucks Only | 3 | 0.07 | 1 | 0.09 |
| TCT | Traffic Control Type | 21 | Officer or Flagman | 128 | 3.04 | 54 | 4.76 |
| TCT | Traffic Control Type | 22 | Oncoming Emergency Vehicle | 3 | 0.07 | 1 | 0.09 |
| TCT | Traffic Control Type | 31 | Pavement Markings (only) | 483 | 11.45 | 115 | 10.14 |
| TCT | Traffic Control Type | 41 | Stop Sign | 141 | 3.34 | 62 | 5.47 |
| TCT | Traffic Control Type | 42 | School Zone Sign | 1 | 0.02 | 1 | 0.09 |
| TCT | Traffic Control Type | 43 | Yield Sign | 58 | 1.38 | 13 | 1.15 |
| TCT | Traffic Control Type | 44 | Work Zone | 2011 | 47.69 | 462 | 40.74 |
| TCT | Traffic Control Type | 45 | Other Warning Signs | 44 | 1.04 | 15 | 1.32 |
| TCT | Traffic Control Type | 51 | Flashing Beacon | 5 | 0.12 | 2 | 0.18 |
| TCT | Traffic Control Type | 98 | None | 1116 | 26.46 | 333 | 29.37 |
| TCT | Traffic Control Type | 99 | Unknown | 6 | 0.14 | 1 | 0.09 |
| DSEX | Driver Gender | 0 | Male | 3687 | 87.43 | 943 | 83.16 |
| DSEX | Driver Gender | 1 | Female | 530 | 12.57 | 191 | 16.84 |
| DLC | Driver License Class | 0 | D | 2452 | 58.15 | 712 | 62.79 |
| DLC | Driver License Class | 1 | C | 353 | 8.37 | 92 | 8.11 |
| DLC | Driver License Class | 2 | B | 148 | 3.51 | 32 | 2.82 |

*Supplemental Table 1: Crash Severity Distribution in South Carolina Work Zones (2014–2018) by Features*

| Feature Header | Feature Name | Category Code | Category Name | Low Severity (Property Damage Only) Count | Low Severity Frequency (%) | High Severity (Injury and Fatality) Count | High Severity Frequency (%) |
|---|---|---|---|---|---|---|---|
| DLC | Driver License Class | 3 | A | 1112 | 26.37 | 265 | 23.37 |
| DLC | Driver License Class | 4 | F | 7 | 0.17 | 0 | 0.00 |
| DLC | Driver License Class | 5 | E | 80 | 1.90 | 12 | 1.06 |
| DLC | Driver License Class | 6 | N | 23 | 0.55 | 5 | 0.44 |
| DLC | Driver License Class | 7 | M | 9 | 0.21 | 1 | 0.09 |
| DLC | Driver License Class | 8 | G | 2 | 0.05 | 1 | 0.09 |
| DLC | Driver License Class | 9 | I | 12 | 0.28 | 8 | 0.71 |
| DLC | Driver License Class | 10 | R | 4 | 0.09 | 2 | 0.18 |
| DLC | Driver License Class | 11 | U | 10 | 0.24 | 3 | 0.26 |
| DLC | Driver License Class | 12 | Not Available | 1 | 0.02 | 0 | 0.00 |
| DLC | Driver License Class | 13 | O | 1 | 0.02 | 0 | 0.00 |
| DLC | Driver License Class | 14 | Not Available | 1 | 0.02 | 0 | 0.00 |
| DLC | Driver License Class | 15 | Not Available | 1 | 0.02 | 0 | 0.00 |
| DLC | Driver License Class | 16 | Not Available | 1 | 0.02 | 0 | 0.00 |
| DLC | Driver License Class | 17 | S | 0 | 0.00 | 1 | 0.09 |
| CTA | Contributed to Collision | 0 | No | 2155 | 51.10 | 616 | 54.32 |
| CTA | Contributed to Collision | 1 | Yes | 2062 | 48.90 | 518 | 45.68 |
| ECS (in miles per hour [MPH]) | Estimated Collision Speed | 1 | 0 - 10 | 1594 | 37.80 | 449 | 39.59 |
| ECS | Estimated Collision Speed | 2 | 10 - 25 | 720 | 17.07 | 121 | 10.67 |
| ECS | Estimated Collision Speed | 3 | 25 - 45 | 993 | 23.55 | 319 | 28.13 |
| ECS | Estimated Collision Speed | 4 | 45 - 55 | 382 | 9.06 | 106 | 9.35 |
| ECS | Estimated Collision Speed | 5 | 55 - 70 | 522 | 12.38 | 134 | 11.82 |
| ECS | Estimated Collision Speed | 6 | 70 - 100 | 6 | 0.14 | 5 | 0.44 |

## Crash Severity Risk Modeling Strategies under Data Imbalance

*Supplemental Table 1: Crash Severity Distribution in South Carolina Work Zones (2014–2018) by Features*

| Feature Header | Feature Name | Category Code | Category Name | Low Severity (Property Damage Only) Count | Low Severity Frequency (%) | High Severity (Injury and Fatality) Count | High Severity Frequency (%) |
|---|---|---|---|---|---|---|---|
| SPL (in MPH) | Posted Speed Limit | 1 | 0 - 10 | 9 | 0.21 | 1 | 0.09 |
| SPL | Posted Speed Limit | 2 | 0 - 20 | 18 | 0.43 | 2 | 0.18 |
| SPL | Posted Speed Limit | 3 | 20 - 30 | 196 | 4.65 | 43 | 3.79 |
| SPL | Posted Speed Limit | 4 | 30 - 40 | 764 | 18.12 | 227 | 20.02 |
| SPL | Posted Speed Limit | 5 | 40 - 50 | 1086 | 25.75 | 346 | 30.51 |
| SPL | Posted Speed Limit | 6 | 50 - 60 | 1884 | 44.68 | 435 | 38.36 |
| SPL | Posted Speed Limit | 7 | 60 - 70 | 260 | 6.17 | 80 | 7.05 |
| NOC | Number of Occupants/Unit | 1 | 1 Occupant/Unit | 3443 | 81.65 | 879 | 77.51 |
| NOC | Number of Occupants/Unit | 2 | 2 Occupants/Unit | 516 | 12.24 | 177 | 15.61 |
| NOC | Number of Occupants/Unit | 3 | 3 Occupants/Unit | 130 | 3.08 | 40 | 3.53 |
| NOC | Number of Occupants/Unit | 4 | 4 Occupants/Unit | 68 | 1.61 | 18 | 1.59 |
| NOC | Number of Occupants/Unit | 5 | 5 Occupants/Unit | 32 | 0.76 | 8 | 0.71 |
| NOC | Number of Occupants/Unit | 6 | 6 Occupants/Unit | 20 | 0.47 | 8 | 0.71 |
| NOC | Number of Occupants/Unit | 7 | 7 Occupants/Unit | 2 | 0.05 | 2 | 0.18 |
| NOC | Number of Occupants/Unit | 8 | 8 Occupants/Unit | 4 | 0.09 | 1 | 0.09 |
| NOC | Number of Occupants/Unit | 9 | 9 Occupants/Unit | 1 | 0.02 | 0 | 0.00 |
| NOC | Number of Occupants/Unit | 13 | 13 Occupants/Unit | 0 | 0.00 | 1 | 0.09 |
| NOC | Number of Occupants/Unit | 22 | 22 Occupants/Unit | 1 | 0.02 | 0 | 0.00 |
| UTC | Unit Type | 12 | Pickup Truck | 1972 | 46.76 | 555 | 48.94 |
| UTC | Unit Type | 13 | Truck Tractor | 951 | 22.55 | 209 | 18.43 |
| UTC | Unit Type | 14 | Other Truck | 434 | 10.29 | 119 | 10.49 |
| UTC | Unit Type | 15 | Full Size Van | 284 | 6.73 | 64 | 5.64 |
| UTC | Unit Type | 16 | Mini Van | 508 | 12.05 | 181 | 15.96 |
| UTC | Unit Type | 38 | Animal Drawn Vehicle | 2 | 0.05 | 0 | 0.00 |
| UTC | Unit Type | 39 | Animal (Ridden) | 1 | 0.02 | 0 | 0.00 |

*Supplemental Table 1: Crash Severity Distribution in South Carolina Work Zones (2014–2018) by Features*

| Feature Header | Feature Name | Category Code | Category Name | Low Severity (Property Damage Only) Count | Low Severity Frequency (%) | High Severity (Injury and Fatality) Count | High Severity Frequency (%) |
|---|---|---|---|---|---|---|---|
| UTC | Unit Type | 61 | School Bus | 30 | 0.71 | 3 | 0.26 |
| UTC | Unit Type | 62 | Passenger Bus | 35 | 0.83 | 3 | 0.26 |
| API | Action Prior to Impact | 1 | Backing | 115 | 2.73 | 7 | 0.62 |
| API | Action Prior to Impact | 2 | Changing Lanes | 419 | 9.94 | 49 | 4.32 |
| API | Action Prior to Impact | 3 | Entering Traffic Lane | 61 | 1.45 | 18 | 1.59 |
| API | Action Prior to Impact | 4 | Leaving Traffic Lane | 68 | 1.61 | 22 | 1.94 |
| API | Action Prior to Impact | 5 | Making U-turn | 10 | 0.24 | 2 | 0.18 |
| API | Action Prior to Impact | 6 | Movements Essentially Straight Ahead | 2144 | 50.84 | 608 | 53.62 |
| API | Action Prior to Impact | 7 | Overtaking/Passing | 19 | 0.45 | 8 | 0.71 |
| API | Action Prior to Impact | 8 | Parked | 106 | 2.51 | 29 | 2.56 |
| API | Action Prior to Impact | 9 | Slowing or Stopped in Traffic | 1072 | 25.42 | 320 | 28.22 |
| API | Action Prior to Impact | 10 | Turning Left | 102 | 2.42 | 55 | 4.85 |
| API | Action Prior to Impact | 11 | Turning Right | 56 | 1.33 | 12 | 1.06 |
| API | Action Prior to Impact | 88 | Other | 4 | 0.09 | 0 | 0.00 |
| API | Action Prior to Impact | 99 | Unknown | 41 | 0.97 | 4 | 0.35 |
| EDAM | Extent of Deformity | 0 | None/Minor | 1421 | 33.70 | 203 | 17.90 |
| EDAM | Extent of Deformity | 2 | Functional Damage | 2067 | 49.02 | 461 | 40.65 |
| EDAM | Extent of Deformity | 3 | Disabling Damage | 550 | 13.04 | 285 | 25.13 |
| EDAM | Extent of Deformity | 4 | Severe/Totaled | 106 | 2.51 | 172 | 15.17 |
| EDAM | Extent of Deformity | 5 | Not Applicable | 3 | 0.07 | 2 | 0.18 |
| EDAM | Extent of Deformity | 9 | Unknown | 70 | 1.66 | 11 | 0.97 |
| MHE | Most Harmful Event | 1 | Cargo/Equipment Loss or Shift | 33 | 0.78 | 2 | 0.18 |
| MHE | Most Harmful Event | 2 | Cross Median/Center | 4 | 0.09 | 2 | 0.18 |
| MHE | Most Harmful Event | 3 | Downhill Runaway | 1 | 0.02 | 0 | 0.00 |
| MHE | Most Harmful Event | 4 | Equipment Failure | 12 | 0.28 | 1 | 0.09 |
| MHE | Most Harmful Event | 5 | Fire/Explosion | 2 | 0.05 | 3 | 0.26 |

*Supplemental Table 1: Crash Severity Distribution in South Carolina Work Zones (2014–2018) by Features*

| Feature Header | Feature Name | Category Code | Category Name | Low Severity (Property Damage Only) Count | Low Severity Frequency (%) | High Severity (Injury and Fatality) Count | High Severity Frequency (%) |
|---|---|---|---|---|---|---|---|
| MHE | Most Harmful Event | 7 | Jackknife | 2 | 0.05 | 0 | 0.00 |
| MHE | Most Harmful Event | 8 | Overturn/Rollover | 34 | 0.81 | 41 | 3.62 |
| MHE | Most Harmful Event | 12 | Spill (Two-wheeled Vehicle) | 1 | 0.02 | 0 | 0.00 |
| MHE | Most Harmful Event | 18 | Other Non-collision | 48 | 1.14 | 5 | 0.44 |
| MHE | Most Harmful Event | 19 | Unknown Non-collision | 2 | 0.05 | 0 | 0.00 |
| MHE | Most Harmful Event | 20 | Animal (Deer Only) | 5 | 0.12 | 0 | 0.00 |
| MHE | Most Harmful Event | 21 | Animal (All Other Than Deer) | 0 | 0.00 | 1 | 0.09 |
| MHE | Most Harmful Event | 22 | Motor Vehicle (In Transport) | 2995 | 71.02 | 781 | 68.87 |
| MHE | Most Harmful Event | 23 | Motor Vehicle (Stopped) | 722 | 17.12 | 203 | 17.90 |
| MHE | Most Harmful Event | 25 | Motor Vehicle (Parked) | 51 | 1.21 | 8 | 0.71 |
| MHE | Most Harmful Event | 26 | Pedalcycle | 0 | 0.00 | 2 | 0.18 |
| MHE | Most Harmful Event | 27 | Pedestrian | 5 | 0.12 | 21 | 1.85 |
| MHE | Most Harmful Event | 28 | Railway Vehicle | 0 | 0.00 | 1 | 0.09 |
| MHE | Most Harmful Event | 29 | Moving Work Zone Maintenance Equipment | 40 | 0.95 | 9 | 0.79 |
| MHE | Most Harmful Event | 38 | Other Movable Object | 27 | 0.64 | 5 | 0.44 |
| MHE | Most Harmful Event | 39 | Unknown Movable Object | 2 | 0.05 | 0 | 0.00 |
| MHE | Most Harmful Event | 40 | Bridge Overhead Structure | 2 | 0.05 | 0 | 0.00 |
| MHE | Most Harmful Event | 42 | Bridge Pier or Abutment | 2 | 0.05 | 0 | 0.00 |
| MHE | Most Harmful Event | 43 | Bridge Rail | 2 | 0.05 | 0 | 0.00 |
| MHE | Most Harmful Event | 45 | Curb | 1 | 0.02 | 1 | 0.09 |
| MHE | Most Harmful Event | 46 | Ditch | 28 | 0.66 | 1 | 0.09 |
| MHE | Most Harmful Event | 47 | Embankment | 6 | 0.14 | 2 | 0.18 |
| MHE | Most Harmful Event | 48 | Fixed Equipment | 3 | 0.07 | 0 | 0.00 |
| MHE | Most Harmful Event | 49 | Fence | 4 | 0.09 | 0 | 0.00 |
| MHE | Most Harmful Event | 50 | Guardrail End | 16 | 0.38 | 4 | 0.35 |

*Supplemental Table 1: Crash Severity Distribution in South Carolina Work Zones (2014–2018) by Features*

| Feature Header | Feature Name | Category Code | Category Name | Low Severity (Property Damage Only) Count | Low Severity Frequency (%) | High Severity (Injury and Fatality) Count | High Severity Frequency (%) |
|---|---|---|---|---|---|---|---|
| MHE | Most Harmful Event | 51 | Guardrail Face | 23 | 0.55 | 5 | 0.44 |
| MHE | Most Harmful Event | 52 | Highway Traffic Sign Post | 15 | 0.36 | 2 | 0.18 |
| MHE | Most Harmful Event | 53 | Impact Attenuator/Crash Cushion | 4 | 0.09 | 1 | 0.09 |
| MHE | Most Harmful Event | 54 | Light/Luminaire Support | 1 | 0.02 | 0 | 0.00 |
| MHE | Most Harmful Event | 55 | Mail Box | 1 | 0.02 | 1 | 0.09 |
| MHE | Most Harmful Event | 56 | Median Barrier | 51 | 1.21 | 15 | 1.32 |
| MHE | Most Harmful Event | 57 | Overhead Sign Post | 2 | 0.05 | 0 | 0.00 |
| MHE | Most Harmful Event | 58 | Other Fixed Object (Post, Pole, Support, etc.) | 9 | 0.21 | 1 | 0.09 |
| MHE | Most Harmful Event | 59 | Other Fixed Object (Wall, Building, Tunnel, etc.) | 4 | 0.09 | 0 | 0.00 |
| MHE | Most Harmful Event | 60 | Tree | 16 | 0.38 | 10 | 0.88 |
| MHE | Most Harmful Event | 61 | Utility Poll | 6 | 0.14 | 2 | 0.18 |
| MHE | Most Harmful Event | 62 | Fixed Work Zone Maintenance Equipment | 21 | 0.50 | 3 | 0.26 |
| MHE | Most Harmful Event | 68 | Other Fixed Object | 14 | 0.33 | 1 | 0.09 |
| FDA | First Deformed Area | 1 | Car/Truck Front | 999 | 23.69 | 371 | 32.72 |
| FDA | First Deformed Area | 2 | Car Front-Right Corner | 343 | 8.13 | 79 | 6.97 |
| FDA | First Deformed Area | 3 | Car/Truck Front-Right Side | 272 | 6.45 | 29 | 2.56 |
| FDA | First Deformed Area | 4 | Car Rear-Right Side | 100 | 2.37 | 20 | 1.76 |
| FDA | First Deformed Area | 5 | Car/Truck Rear-Right Corner | 79 | 1.87 | 17 | 1.50 |
| FDA | First Deformed Area | 6 | Car Rear/Truck Coupler with First Trailer | 86 | 2.04 | 22 | 1.94 |
| FDA | First Deformed Area | 7 | Car/Truck Rear-Left Corner | 918 | 21.77 | 260 | 22.93 |
| FDA | First Deformed Area | 8 | Car Rear-Left Side | 103 | 2.44 | 23 | 2.03 |
| FDA | First Deformed Area | 9 | Car/Truck Front-Left Side | 65 | 1.54 | 21 | 1.85 |
| FDA | First Deformed Area | 10 | Truck Front-Left Corner | 89 | 2.11 | 23 | 2.03 |

*Supplemental Table 1: Crash Severity Distribution in South Carolina Work Zones (2014–2018) by Features*

| Feature Header | Feature Name | Category Code | Category Name | Low Severity (Property Damage Only) Count | Low Severity Frequency (%) | High Severity (Injury and Fatality) Count | High Severity Frequency (%) |
|---|---|---|---|---|---|---|---|
| FDA | First Deformed Area | 11 | Car/Truck Hood | 175 | 4.15 | 32 | 2.82 |
| FDA | First Deformed Area | 12 | Car/Truck Front Windshield | 251 | 5.95 | 76 | 6.70 |
| FDA | First Deformed Area | 13 | Car/Truck Roof | 7 | 0.17 | 0 | 0.00 |
| FDA | First Deformed Area | 14 | Car Rear Windshield | 2 | 0.05 | 1 | 0.09 |
| FDA | First Deformed Area | 15 | Car Trunk | 2 | 0.05 | 3 | 0.26 |
| FDA | First Deformed Area | 17 | Not Available | 0 | 0.00 | 4 | 0.35 |
| FDA | First Deformed Area | 23 | Not Available | 1 | 0.02 | 0 | 0.00 |
| FDA | First Deformed Area | 30 | Not Available | 1 | 0.02 | 0 | 0.00 |
| FDA | First Deformed Area | 32 | Truck Front-Right Side of the First Trailer | 3 | 0.07 | 0 | 0.00 |
| FDA | First Deformed Area | 33 | Truck Middle-Left Side of the First Trailer | 22 | 0.52 | 4 | 0.35 |
| FDA | First Deformed Area | 34 | Truck Rear-Right Side of the First Trailer | 35 | 0.83 | 5 | 0.44 |
| FDA | First Deformed Area | 35 | Truck Rear-Right Corner of the First Trailer | 54 | 1.28 | 15 | 1.32 |
| FDA | First Deformed Area | 36 | Trailer Coupler (Connects First and Second Trailers) | 8 | 0.19 | 2 | 0.18 |
| FDA | First Deformed Area | 37 | Truck Rear-Left Corner of the First Trailer | 44 | 1.04 | 15 | 1.32 |
| FDA | First Deformed Area | 38 | Truck Rear-Left Side of the First Trailer | 23 | 0.55 | 6 | 0.53 |
| FDA | First Deformed Area | 39 | Truck Middle-Left Side of the First Trailer | 64 | 1.52 | 4 | 0.35 |
| FDA | First Deformed Area | 40 | Truck Front-Left Side of the First Trailer | 37 | 0.88 | 5 | 0.44 |
| FDA | First Deformed Area | 41 | Truck Front-Left Corner of the First Trailer | 21 | 0.50 | 4 | 0.35 |
| FDA | First Deformed Area | 42 | Truck First Trailer Roof | 3 | 0.07 | 0 | 0.00 |
| FDA | First Deformed Area | 43 | Not Available | 6 | 0.14 | 0 | 0.00 |

*Supplemental Table 1: Crash Severity Distribution in South Carolina Work Zones (2014–2018) by Features*

| Feature Header | Feature Name | Category Code | Category Name | Low Severity (Property Damage Only) Count | Low Severity Frequency (%) | High Severity (Injury and Fatality) Count | High Severity Frequency (%) |
|---|---|---|---|---|---|---|---|
| FDA | First Deformed Area | 50 | Not Available | 1 | 0.02 | 0 | 0.00 |
| FDA | First Deformed Area | 54 | Truck Rear-Right Side of the Second Trailer | 3 | 0.07 | 0 | 0.00 |
| FDA | First Deformed Area | 55 | Truck Rear-Right Corner of the Second Trailer | 13 | 0.31 | 1 | 0.09 |
| FDA | First Deformed Area | 56 | Truck Rear | 3 | 0.07 | 1 | 0.09 |
| FDA | First Deformed Area | 57 | Truck Rear-Left Corner of the Second Trailer | 65 | 1.54 | 28 | 2.47 |
| FDA | First Deformed Area | 58 | Truck Rear-Left Side of the Second Trailer | 10 | 0.24 | 1 | 0.09 |
| FDA | First Deformed Area | 59 | Truck Middle-Left Side of the Second Trailer | 9 | 0.21 | 0 | 0.00 |
| FDA | First Deformed Area | 60 | Truck Front-Left Side of the Second Trailer | 2 | 0.05 | 0 | 0.00 |
| FDA | First Deformed Area | 81 | None | 183 | 4.34 | 26 | 2.29 |
| FDA | First Deformed Area | 82 | Not Available | 1 | 0.02 | 0 | 0.00 |
| FDA | First Deformed Area | 92 | Rollover | 10 | 0.24 | 18 | 1.59 |
| FDA | First Deformed Area | 93 | Total (All Areas) | 1 | 0.02 | 3 | 0.26 |
| FDA | First Deformed Area | 94 | Under Carriage | 18 | 0.43 | 1 | 0.09 |
| FDA | First Deformed Area | 98 | Other | 23 | 0.55 | 5 | 0.44 |
| FDA | First Deformed Area | 99 | Unknown | 62 | 1.47 | 9 | 0.79 |
| OSL | Occupant Seating Location | 1 | Driver | 3579 | 84.87 | 888 | 78.31 |
| OSL | Occupant Seating Location | 2 | Front Row Middle | 6 | 0.14 | 1 | 0.09 |
| OSL | Occupant Seating Location | 3 | Front Row Right | 391 | 9.27 | 140 | 12.35 |
| OSL | Occupant Seating Location | 4 | Middle Row Left | 70 | 1.66 | 19 | 1.68 |
| OSL | Occupant Seating Location | 5 | Middle Row Middle | 26 | 0.62 | 9 | 0.79 |

*Supplemental Table 1: Crash Severity Distribution in South Carolina Work Zones (2014–2018) by Features*

| Feature Header | Feature Name | Category Code | Category Name | Low Severity (Property Damage Only) Count | Low Severity Frequency (%) | High Severity (Injury and Fatality) Count | High Severity Frequency (%) |
|---|---|---|---|---|---|---|---|
| OSL | Occupant Seating Location | 6 | Middle Row Right | 91 | 2.16 | 26 | 2.29 |
| OSL | Occupant Seating Location | 7 | Last Row Left | 7 | 0.17 | 2 | 0.18 |
| OSL | Occupant Seating Location | 8 | Last Row Middle | 5 | 0.12 | 2 | 0.18 |
| OSL | Occupant Seating Location | 9 | Last Row Right | 7 | 0.17 | 3 | 0.26 |
| OSL | Occupant Seating Location | 20 | Pedestrian | 2 | 0.05 | 34 | 3.00 |
| OSL | Occupant Seating Location | 30 | Trailing Unit | 0 | 0.00 | 3 | 0.26 |
| OSL | Occupant Seating Location | 40 | Bus or Van (4th Row or Higher) | 6 | 0.14 | 1 | 0.09 |
| OSL | Occupant Seating Location | 50 | Other Enclosed Area (Non-trailing) | 0 | 0.00 | 1 | 0.09 |
| OSL | Occupant Seating Location | 51 | Other Unenclosed Area (Non-trailing) | 0 | 0.00 | 2 | 0.18 |
| OSL | Occupant Seating Location | 60 | Sleeper of Cab | 4 | 0.09 | 0 | 0.00 |
| OSL | Occupant Seating Location | 70 | Riding on Unit Exterior | 0 | 0.00 | 2 | 0.18 |
| OSL | Occupant Seating Location | 99 | Unknown | 23 | 0.55 | 1 | 0.09 |
| REU | Occupant Restraint Type | 0 | None Used | 40 | 0.95 | 74 | 6.53 |
| REU | Occupant Restraint Type | 11 | Shoulder Belt Only | 13 | 0.31 | 4 | 0.35 |
| REU | Occupant Restraint Type | 12 | Lap Belt Only | 15 | 0.36 | 2 | 0.18 |
| REU | Occupant Restraint Type | 13 | Shoulder and Lap Belt | 3870 | 91.77 | 984 | 86.77 |
| REU | Occupant Restraint Type | 21 | Child Safety Seat | 60 | 1.42 | 15 | 1.32 |
| REU | Occupant Restraint Type | 31 | Helmet (for Pedestrian and Motor/Pedalcycle) | 2 | 0.05 | 8 | 0.71 |
| REU | Occupant Restraint Type | 41 | Protective Pads (for Pedestrian and Motor/Pedalcycle) | 0 | 0.00 | 1 | 0.09 |
| REU | Occupant Restraint Type | 51 | Reflective Clothing (for Pedestrian and Motor/Pedalcycle) | 0 | 0.00 | 15 | 1.32 |
| REU | Occupant Restraint Type | 61 | Lighting (for Pedestrian and Motor/Pedalcycle) | 0 | 0.00 | 1 | 0.09 |
| REU | Occupant Restraint Type | 88 | Other | 7 | 0.17 | 4 | 0.35 |

*Supplemental Table 1: Crash Severity Distribution in South Carolina Work Zones (2014–2018) by Features*

| Feature Header | Feature Name | Category Code | Category Name | Low Severity (Property Damage Only) Count | Low Severity Frequency (%) | High Severity (Injury and Fatality) Count | High Severity Frequency (%) |
|---|---|---|---|---|---|---|---|
| REU | Occupant Restraint Type | 99 | Unknown | 210 | 4.98 | 26 | 2.29 |
| LAI | Location After Impact | 1 | Not Trapped | 4182 | 99.17 | 1034 | 91.18 |
| LAI | Location After Impact | 2 | Extricated (Mechanical Means) | 1 | 0.02 | 50 | 4.41 |
| LAI | Location After Impact | 3 | Freed (Non-mechanical) | 3 | 0.07 | 19 | 1.68 |
| LAI | Location After Impact | 4 | Not Applicable | 7 | 0.17 | 28 | 2.47 |
| LAI | Location After Impact | 9 | Unknown | 24 | 0.57 | 3 | 0.26 |

## **Supplemental Table 2**

Top 124 Unique Features Selected Through Merging of Five Feature Selection Techniques

# Crash Severity Risk Modeling Strategies under Data Imbalance

## *Supplemental Table 2: Top 124 Unique Features Selected Through Merging of Five Feature Selection Techniques*

| | | | | |
|---|---|---|---|---|
| TIM_3 | PRC_3_x_MAC_10 | FDA_17 | **MAC_50** | FDA_13 |
| MHE_26 | **LAI_3** | FHE_22 | **API_1** | **FHE_27** |
| **TWAY_1** | **UNT_4** | MHE_46 | **UNT_2** | **TWAY_3** |
| DLR_3 | OSL_99 | WZL_4 | REU_12 | **RCT_1** |
| **MAC_42** | **OSL_1** | **FHE_27_x_LAI_4** | **FHE_8** | **FHE_8_x_FDA_92** |
| REU_31 | ODR_1 | PRC_3 | **REU_51** | DLR_2 |
| JCT_1 | OSL_51 | **EDAM_4** | FHE_55 | **EDAM_3** |
| DSEX_0 | RCT_3_x_TWAY_1 | **MAC_50_x_API_2** | SPL_6 | **MHE_8** |
| ECS_3 | **MHE_27** | OSL_60 | **FDA_92** | OSL_70 |
| **REU_0** | WPR_1 | **PRC_18_x_API_2** | **REU_13** | WZT_1 |
| API_10 | **FHE_27_x_REU_51** | LOA_2 | DLC_4 | PRC_73 |
| RCT_1_x_SPL_4 | **RCT_1_x_JCT_13** | **OSL_20** | **LAI_1** | BIR_4 |
| OSL_30 | ODR_3 | HEL_9 | ODR_0 | **LAI_2** |
| **FDA_3** | PRC_10 | LAI_9 | FHE_1 | **FDA_1** |
| FDA_36 | FHE_4_x_PRC_73 | **LAI_4** | TIM_1 | **API_2** |
| SIC_4 | **RCT_1_x_SPL_6** | FHE_26 | **MAC_70** | NOC_1 |
| TCT_98 | RSC_3 | **PRC_18** | TIM_2 | BDO_0 |
| UTC_12 | **PRC_18_x_MAC_50** | WCC_5_x_RSC_3 | PRC_79_x_MHE_1 | **ECS_2** |
| ODR_2 | **TWAY_3_x_SPL_6** | PRC_63 | SPL_5 | **RCT_1_x_TWAY_3** |
| FDA_93 | **MAC_70_x_API_1** | LOA_1 | ALC_1 | WPR_2 |
| FHE_4 | **EDAM_0** | PRC_12 | MAC_0 | MHE_22 |
| WZT_4 | **EDAM_2** | BDO_14 | **MAC_10** | TCT_44 |
| **PRC_28_x_MAC_70** | PRC_13 | MAC_20 | **PRC_16** | DLC_0 |
| DLR_0 | REU_99 | **UNT_3** | **FHE_27_x_OSL_20** | DLR_1 |
| API_6_x_FDA_1 | PRC_63_x_FDA_15 | UNT_1_x_MAC_0 | **RCT_2** | |

Notes for Supplemental Table 2:

1. The five feature selection techniques include Pearson's correlation, Feature Importance using Random Forest (FI using RF), Recursive Feature Elimination (RFE) using Logistic Regression, Chi-squared Test Statistics, and Discriminative Mutual Information (DMI).
2. Binary variables are presented as FeatureHeader_CategoryCode. For example, MAC_10 refers to the "Manner of Collision" feature with the category code 10.
3. Interaction terms are presented as Feature1Header_CategoryCode_x_Feature2Header_CategoryCode. For example, PRC_3_x_MAC_10 represents the interaction between "Primary Contributing Factor" with its corresponding category code 3 and "Manner of Collision" with its corresponding category code 10.
4. The Feature Header and its corresponding Category Codes can be found in Supplemental Table 1 for reference.
5. The **bold** feature variables are derived from applying DMI, which has been identified as the most effective feature selection technique for achieving high-severity recall in this study.

**Supplemental Table 3**

Key Hyperparameter Settings for Statistical, Machine Learning (ML), and Deep Learning (DL) Models Used in This Study

***Supplemental Table 3: Key Hyperparameter Settings for Statistical, ML, and DL Models Used in This Study***

| **Model** | **Key Hyperparameters** |
|---|---|
| Bayesian Mixed Logit (BML) | Prior Distribution: Uniform(-5, 5) for all coefficients<br>Number of Chains: 2<br>Number of Iterations: 2000<br>Likelihood Function: Bernoulli (logit)<br>Adaptive Sampling: adapt_delta = 0.99 |
| CatBoost | Tree Depth: 6, 8<br>Grow Policy: SymmetricTree / Depthwise<br>L2 Leaf Regularization: 2.15 – 2.80<br>Learning Rate: 0.03 – 0.07<br>Maximum CTR Complexity: 2 – 4<br>One-Hot Encoding Threshold: 3 – 10 |
| Extra Trees (ET) | Splitting Criterion: Gini / Entropy (Entr variant)<br>Applicable Problem Types: Binary, Multiclass |
| LightGBM | Extra Trees Enabled: True (XT variant)<br>Learning Rate: 0.03<br>Number of Leaves: 128<br>Feature Fraction: 0.9<br>Minimum Samples per Leaf: 3 (Large variant) |
| NeuralNetFastAI | Batch Size: 256, 2048<br>Embedding Dropout: 0.05 – 0.54<br>Number of Training Epochs: 29, 43<br>Hidden Layer Structure: [800, 400], [200, 100]<br>Learning Rate: 0.015 – 0.090<br>Dropout Probability (ps): 0.10 – 0.24 |
| NeuralNetTorch | Activation Function: ELU<br>Dropout Probability: 0.10 – 0.12<br>Hidden Layer Size: 108, 213<br>Number of Hidden Layers: 4<br>Learning Rate: 0.0010 – 0.0027<br>Batch Normalization: True<br>Weight Decay: 1e-12 to 1e-10 |

***Supplemental Table 3: Key Hyperparameter Settings for Statistical, ML, and DL Models Used in This Study***

| **Model** | **Key Hyperparameters** |
|---|---|
| Random Forest (RF) | Splitting Criterion: Gini / Entropy (Entr variant)<br>Applicable Problem Types: Binary, Multiclass |
| XGBoost | Column Sampling (by Tree): 0.66 – 0.69<br>Enable Categorical: False<br>Learning Rate: 0.018 – 0.088<br>Maximum Tree Depth: 5 – 10<br>Minimum Child Weight: 0.60 – 0.63 |

# **Supplemental Table 4**

Summary of Model Results for Different Training Datasets

***Supplemental Table 4: Summary of Model Results for Different Training Datasets***

| Training Dataset Number | Feature Selection Technique | Data Balancing Technique | Model | Accuarcy | LS Precision | HS Precision | LS Recall | HS Recall | LS F1 Score | HS F1 Score | Balanced Accuracy | ROC AUC |
|---|---|---|---|---|---|---|---|---|---|---|---|---|
| 1 | None | None | BML | 0.79 | 0.82 | 0.54 | 0.94 | 0.25 | 0.88 | 0.34 | 0.60 | 0.70 |
| | | | CatBoost | 0.82 | 0.82 | 0.75 | 0.98 | 0.21 | 0.89 | 0.33 | 0.60 | 0.74 |
| | | | ExtraTreesEntr | 0.81 | 0.81 | 0.83 | 0.99 | 0.12 | 0.89 | 0.21 | 0.55 | 0.76 |
| | | | ExtraTreesGini | 0.81 | 0.81 | 0.80 | 0.99 | 0.11 | 0.89 | 0.20 | 0.55 | 0.76 |
| | | | LightGBM | 0.81 | 0.82 | 0.67 | 0.97 | 0.22 | 0.89 | 0.33 | 0.60 | 0.74 |
| | | | LightGBMLarge | 0.82 | 0.82 | 0.74 | 0.98 | 0.21 | 0.89 | 0.32 | 0.60 | 0.75 |
| | | | LightGBMXT | 0.81 | 0.82 | 0.67 | 0.97 | 0.22 | 0.89 | 0.33 | 0.60 | 0.74 |
| | | | NeuralNetFastAI | 0.73 | 0.87 | 0.41 | 0.78 | 0.56 | 0.82 | 0.47 | 0.67 | 0.73 |
| | | | NeuralNetTorch | 0.78 | 0.85 | 0.48 | 0.88 | 0.40 | 0.86 | 0.44 | 0.64 | 0.71 |
| | | | RandomForestEntr | 0.81 | 0.81 | 0.83 | 0.99 | 0.12 | 0.89 | 0.21 | 0.55 | 0.77 |
| | | | RandomForestGini | 0.81 | 0.81 | 0.83 | 0.99 | 0.13 | 0.89 | 0.23 | 0.56 | 0.76 |
| | | | XGBoost | 0.82 | 0.82 | 0.72 | 0.98 | 0.21 | 0.89 | 0.33 | 0.60 | 0.75 |
| 2 | None | High Severity Weight = 4 | BML | 0.70 | 0.86 | 0.37 | 0.74 | 0.55 | 0.80 | 0.44 | 0.65 | 0.69 |
| | | | CatBoost | 0.73 | 0.88 | 0.42 | 0.77 | 0.62 | 0.82 | 0.50 | 0.70 | 0.75 |
| | | | ExtraTreesEntr | 0.81 | 0.82 | 0.70 | 0.98 | 0.18 | 0.89 | 0.29 | 0.58 | 0.76 |
| | | | ExtraTreesGini | 0.81 | 0.82 | 0.75 | 0.98 | 0.17 | 0.89 | 0.28 | 0.57 | 0.76 |
| | | | LightGBM | 0.73 | 0.89 | 0.41 | 0.75 | 0.64 | 0.81 | 0.50 | 0.70 | 0.76 |
| | | | LightGBMLarge | 0.78 | 0.85 | 0.49 | 0.88 | 0.44 | 0.87 | 0.47 | 0.66 | 0.75 |
| | | | LightGBMXT | 0.75 | 0.88 | 0.43 | 0.79 | 0.58 | 0.83 | 0.50 | 0.68 | 0.76 |
| | | | NeuralNetFastAI | 0.74 | 0.87 | 0.42 | 0.79 | 0.56 | 0.83 | 0.48 | 0.68 | 0.73 |
| | | | NeuralNetTorch | 0.79 | 0.86 | 0.50 | 0.88 | 0.46 | 0.87 | 0.48 | 0.67 | 0.74 |
| | | | RandomForestEntr | 0.81 | 0.82 | 0.71 | 0.98 | 0.18 | 0.89 | 0.28 | 0.58 | 0.76 |
| | | | RandomForestGini | 0.81 | 0.82 | 0.68 | 0.97 | 0.20 | 0.89 | 0.31 | 0.58 | 0.77 |
| | | | XGBoost | 0.75 | 0.88 | 0.43 | 0.79 | 0.60 | 0.83 | 0.50 | 0.70 | 0.76 |

*Supplemental Table 4: Summary of Model Results for Different Training Datasets*

| Training Dataset Number | Feature Selection Technique | Data Balancing Technique | Model | Accuarcy | LS Precision | HS Precision | LS Recall | HS Recall | LS F1 Score | HS F1 Score | Balanced Accuracy | ROC AUC |
|---|---|---|---|---|---|---|---|---|---|---|---|---|
| 3 | Pearson's Correlation | None | BML | 0.82 | 0.82 | 0.75 | 0.98 | 0.21 | 0.89 | 0.32 | 0.59 | 0.75 |
| | | | CatBoost | 0.81 | 0.83 | 0.66 | 0.97 | 0.25 | 0.89 | 0.36 | 0.61 | 0.73 |
| | | | ExtraTreesEntr | 0.78 | 0.82 | 0.49 | 0.93 | 0.24 | 0.87 | 0.32 | 0.58 | 0.70 |
| | | | ExtraTreesGini | 0.79 | 0.82 | 0.50 | 0.93 | 0.25 | 0.87 | 0.34 | 0.59 | 0.70 |
| | | | LightGBM | 0.80 | 0.83 | 0.59 | 0.95 | 0.25 | 0.88 | 0.35 | 0.60 | 0.71 |
| | | | LightGBMLarge | 0.81 | 0.82 | 0.65 | 0.97 | 0.22 | 0.89 | 0.33 | 0.60 | 0.70 |
| | | | LightGBMXT | 0.80 | 0.83 | 0.59 | 0.95 | 0.25 | 0.88 | 0.35 | 0.60 | 0.71 |
| | | | NeuralNetFastAI | 0.78 | 0.85 | 0.49 | 0.88 | 0.43 | 0.86 | 0.45 | 0.66 | 0.76 |
| | | | NeuralNetTorch | 0.80 | 0.84 | 0.56 | 0.93 | 0.34 | 0.88 | 0.42 | 0.64 | 0.76 |
| | | | RandomForestEntr | 0.81 | 0.82 | 0.64 | 0.97 | 0.21 | 0.89 | 0.32 | 0.59 | 0.71 |
| | | | RandomForestGini | 0.81 | 0.82 | 0.63 | 0.97 | 0.21 | 0.89 | 0.32 | 0.59 | 0.71 |
| | | | XGBoost | 0.81 | 0.82 | 0.62 | 0.96 | 0.24 | 0.89 | 0.34 | 0.60 | 0.71 |
| 4 | Feature Importance Using Random Forest | None | BML | 0.81 | 0.82 | 0.68 | 0.98 | 0.19 | 0.89 | 0.30 | 0.59 | 0.71 |
| | | | CatBoost | 0.81 | 0.83 | 0.61 | 0.95 | 0.28 | 0.89 | 0.38 | 0.62 | 0.71 |
| | | | ExtraTreesEntr | 0.81 | 0.82 | 0.64 | 0.97 | 0.23 | 0.89 | 0.34 | 0.60 | 0.73 |
| | | | ExtraTreesGini | 0.81 | 0.83 | 0.65 | 0.96 | 0.26 | 0.89 | 0.37 | 0.61 | 0.73 |
| | | | LightGBM | 0.80 | 0.83 | 0.53 | 0.93 | 0.29 | 0.88 | 0.38 | 0.61 | 0.71 |
| | | | LightGBMLarge | 0.81 | 0.83 | 0.66 | 0.96 | 0.26 | 0.89 | 0.37 | 0.61 | 0.71 |
| | | | LightGBMXT | 0.80 | 0.83 | 0.56 | 0.94 | 0.26 | 0.88 | 0.36 | 0.60 | 0.71 |
| | | | NeuralNetFastAI | 0.69 | 0.86 | 0.36 | 0.73 | 0.56 | 0.79 | 0.44 | 0.64 | 0.71 |
| | | | NeuralNetTorch | 0.79 | 0.85 | 0.50 | 0.88 | 0.43 | 0.87 | 0.46 | 0.66 | 0.71 |
| | | | RandomForestEntr | 0.81 | 0.82 | 0.65 | 0.97 | 0.24 | 0.89 | 0.35 | 0.60 | 0.73 |
| | | | RandomForestGini | 0.81 | 0.82 | 0.65 | 0.97 | 0.23 | 0.89 | 0.34 | 0.60 | 0.73 |
| | | | XGBoost | 0.81 | 0.83 | 0.62 | 0.96 | 0.26 | 0.89 | 0.36 | 0.61 | 0.71 |

***Supplemental Table 4: Summary of Model Results for Different Training Datasets***

| Training Dataset Number | Feature Selection Technique | Data Balancing Technique | Model | Accuarcy | LS Precision | HS Precision | LS Recall | HS Recall | LS F1 Score | HS F1 Score | Balanced Accuracy | ROC AUC |
|---|---|---|---|---|---|---|---|---|---|---|---|---|
| 5 | Recursive Feature Elimination Using Logistic Regression | None | BML | 0.82 | 0.82 | 0.76 | 0.98 | 0.20 | 0.89 | 0.32 | 0.59 | 0.73 |
| | | | CatBoost | 0.79 | 0.85 | 0.51 | 0.89 | 0.41 | 0.87 | 0.45 | 0.65 | 0.72 |
| | | | ExtraTreesEntr | 0.82 | 0.82 | 0.72 | 0.98 | 0.21 | 0.89 | 0.33 | 0.60 | 0.72 |
| | | | ExtraTreesGini | 0.82 | 0.82 | 0.72 | 0.98 | 0.21 | 0.89 | 0.33 | 0.60 | 0.72 |
| | | | LightGBM | 0.81 | 0.82 | 0.72 | 0.98 | 0.21 | 0.89 | 0.32 | 0.60 | 0.72 |
| | | | LightGBMLarge | 0.81 | 0.82 | 0.72 | 0.98 | 0.20 | 0.89 | 0.31 | 0.59 | 0.71 |
| | | | LightGBMXT | 0.81 | 0.82 | 0.71 | 0.98 | 0.18 | 0.89 | 0.29 | 0.58 | 0.71 |
| | | | NeuralNetFastAI | 0.78 | 0.85 | 0.48 | 0.88 | 0.43 | 0.86 | 0.45 | 0.66 | 0.72 |
| | | | NeuralNetTorch | 0.79 | 0.85 | 0.51 | 0.89 | 0.42 | 0.87 | 0.46 | 0.66 | 0.72 |
| | | | RandomForestEntr | 0.82 | 0.82 | 0.72 | 0.98 | 0.21 | 0.89 | 0.33 | 0.60 | 0.71 |
| | | | RandomForestGini | 0.82 | 0.82 | 0.72 | 0.98 | 0.21 | 0.89 | 0.33 | 0.60 | 0.71 |
| | | | XGBoost | 0.82 | 0.82 | 0.72 | 0.98 | 0.21 | 0.89 | 0.32 | 0.60 | 0.72 |
| 6 | Chi-squared Test Statistics | None | BML | 0.82 | 0.82 | 0.74 | 0.98 | 0.21 | 0.89 | 0.33 | 0.60 | 0.76 |
| | | | CatBoost | 0.81 | 0.83 | 0.66 | 0.96 | 0.26 | 0.89 | 0.37 | 0.61 | 0.75 |
| | | | ExtraTreesEntr | 0.80 | 0.83 | 0.56 | 0.94 | 0.28 | 0.88 | 0.38 | 0.61 | 0.70 |
| | | | ExtraTreesGini | 0.80 | 0.83 | 0.57 | 0.94 | 0.29 | 0.88 | 0.39 | 0.62 | 0.70 |
| | | | LightGBM | 0.80 | 0.83 | 0.58 | 0.95 | 0.27 | 0.88 | 0.36 | 0.61 | 0.72 |
| | | | LightGBMLarge | 0.80 | 0.83 | 0.55 | 0.94 | 0.27 | 0.88 | 0.36 | 0.60 | 0.71 |
| | | | LightGBMXT | 0.80 | 0.82 | 0.60 | 0.96 | 0.23 | 0.89 | 0.33 | 0.60 | 0.73 |
| | | | NeuralNetFastAI | 0.81 | 0.83 | 0.59 | 0.94 | 0.30 | 0.89 | 0.40 | 0.62 | 0.76 |
| | | | NeuralNetTorch | 0.78 | 0.86 | 0.47 | 0.86 | 0.46 | 0.86 | 0.47 | 0.66 | 0.76 |
| | | | RandomForestEntr | 0.80 | 0.83 | 0.56 | 0.94 | 0.28 | 0.88 | 0.37 | 0.61 | 0.70 |
| | | | RandomForestGini | 0.80 | 0.83 | 0.55 | 0.94 | 0.29 | 0.88 | 0.38 | 0.62 | 0.70 |
| | | | XGBoost | 0.81 | 0.83 | 0.63 | 0.96 | 0.26 | 0.89 | 0.37 | 0.61 | 0.74 |

*Supplemental Table 4: Summary of Model Results for Different Training Datasets*

| Training Dataset Number | Feature Selection Technique | Data Balancing Technique | Model | Acuarcy | LS Precision | HS Precision | LS Recall | HS Recall | LS F1 Score | HS F1 Score | Balanced Accuracy | ROC AUC |
|---|---|---|---|---|---|---|---|---|---|---|---|---|
| 7 | Discriminative Mutual Information | None | BML | 0.82 | 0.82 | 0.73 | 0.98 | 0.21 | 0.89 | 0.33 | 0.59 | 0.75 |
| | | | CatBoost | 0.58 | 0.92 | 0.32 | 0.52 | 0.83 | 0.66 | 0.46 | 0.68 | 0.76 |
| | | | ExtraTreesEntr | 0.64 | 0.88 | 0.33 | 0.63 | 0.68 | 0.73 | 0.44 | 0.66 | 0.71 |
| | | | ExtraTreesGini | 0.63 | 0.88 | 0.32 | 0.62 | 0.68 | 0.72 | 0.44 | 0.65 | 0.71 |
| | | | LightGBM | 0.59 | 0.91 | 0.31 | 0.53 | 0.80 | 0.67 | 0.45 | 0.66 | 0.73 |
| | | | LightGBMLarge | 0.59 | 0.89 | 0.31 | 0.55 | 0.75 | 0.68 | 0.44 | 0.65 | 0.72 |
| | | | LightGBMXT | 0.59 | 0.90 | 0.31 | 0.54 | 0.77 | 0.67 | 0.44 | 0.66 | 0.73 |
| | | | NeuralNetFastAI | 0.65 | 0.90 | 0.35 | 0.63 | 0.73 | 0.74 | 0.47 | 0.68 | 0.75 |
| | | | NeuralNetTorch | 0.60 | 0.92 | 0.33 | 0.54 | 0.82 | 0.68 | 0.47 | 0.68 | 0.75 |
| | | | RandomForestEntr | 0.64 | 0.88 | 0.33 | 0.64 | 0.67 | 0.74 | 0.44 | 0.66 | 0.71 |
| | | | RandomForestGini | 0.63 | 0.87 | 0.32 | 0.62 | 0.67 | 0.73 | 0.43 | 0.64 | 0.71 |
| | | | XGBoost | 0.58 | 0.91 | 0.31 | 0.52 | 0.81 | 0.67 | 0.45 | 0.66 | 0.75 |
| 8 | Top Ranked Features Merged from Various Feature Selection Methods | None | BML | 0.81 | 0.82 | 0.66 | 0.97 | 0.23 | 0.89 | 0.35 | 0.60 | 0.75 |
| | | | CatBoost | 0.75 | 0.87 | 0.43 | 0.79 | 0.57 | 0.83 | 0.49 | 0.68 | 0.76 |
| | | | ExtraTreesEntr | 0.81 | 0.84 | 0.59 | 0.94 | 0.33 | 0.89 | 0.42 | 0.64 | 0.76 |
| | | | ExtraTreesGini | 0.80 | 0.84 | 0.57 | 0.93 | 0.34 | 0.88 | 0.42 | 0.64 | 0.76 |
| | | | LightGBM | 0.80 | 0.84 | 0.56 | 0.92 | 0.35 | 0.88 | 0.43 | 0.64 | 0.75 |
| | | | LightGBMLarge | 0.81 | 0.84 | 0.61 | 0.95 | 0.32 | 0.89 | 0.42 | 0.64 | 0.75 |
| | | | LightGBMXT | 0.81 | 0.84 | 0.59 | 0.94 | 0.31 | 0.89 | 0.41 | 0.62 | 0.75 |
| | | | NeuralNetFastAI | 0.70 | 0.88 | 0.38 | 0.71 | 0.65 | 0.79 | 0.48 | 0.68 | 0.74 |
| | | | NeuralNetTorch | 0.78 | 0.86 | 0.49 | 0.86 | 0.48 | 0.86 | 0.48 | 0.67 | 0.75 |
| | | | RandomForestEntr | 0.81 | 0.84 | 0.59 | 0.93 | 0.36 | 0.89 | 0.45 | 0.64 | 0.75 |
| | | | RandomForestGini | 0.80 | 0.84 | 0.56 | 0.93 | 0.35 | 0.88 | 0.43 | 0.64 | 0.76 |
| | | | XGBoost | 0.80 | 0.84 | 0.52 | 0.91 | 0.37 | 0.87 | 0.44 | 0.64 | 0.74 |

## Crash Severity Risk Modeling Strategies under Data Imbalance

### *Supplemental Table 4: Summary of Model Results for Different Training Datasets*

| Training Dataset Number | Feature Selection Technique | Data Balancing Technique | Model | Accuarcy | LS Precision | HS Precision | LS Recall | HS Recall | LS F1 Score | HS F1 Score | Balanced Accuracy | ROC AUC |
|---|---|---|---|---|---|---|---|---|---|---|---|---|
| 9 | Top Ranked Features Merged from Various Feature Selection Methods | High Severity Weight = 4 | BML | 0.71 | 0.88 | 0.39 | 0.74 | 0.61 | 0.80 | 0.48 | 0.68 | 0.75 |
| | | | CatBoost | 0.75 | 0.88 | 0.43 | 0.79 | 0.60 | 0.83 | 0.50 | 0.70 | 0.77 |
| | | | ExtraTreesEntr | 0.81 | 0.82 | 0.75 | 0.98 | 0.19 | 0.89 | 0.30 | 0.58 | 0.76 |
| | | | ExtraTreesGini | 0.82 | 0.82 | 0.78 | 0.99 | 0.18 | 0.89 | 0.29 | 0.58 | 0.77 |
| | | | LightGBM | 0.74 | 0.88 | 0.42 | 0.77 | 0.62 | 0.82 | 0.50 | 0.70 | 0.77 |
| | | | LightGBMLarge | 0.78 | 0.86 | 0.49 | 0.87 | 0.46 | 0.86 | 0.47 | 0.66 | 0.75 |
| | | | LightGBMXT | 0.74 | 0.88 | 0.42 | 0.77 | 0.62 | 0.82 | 0.50 | 0.70 | 0.76 |
| | | | NeuralNetFastAI | 0.79 | 0.86 | 0.50 | 0.88 | 0.46 | 0.87 | 0.48 | 0.67 | 0.76 |
| | | | NeuralNetTorch | 0.80 | 0.85 | 0.54 | 0.91 | 0.40 | 0.88 | 0.46 | 0.66 | 0.76 |
| | | | RandomForestEntr | 0.82 | 0.82 | 0.74 | 0.98 | 0.21 | 0.89 | 0.32 | 0.60 | 0.76 |
| | | | RandomForestGini | 0.81 | 0.82 | 0.74 | 0.98 | 0.20 | 0.89 | 0.31 | 0.59 | 0.76 |
| | | | XGBoost | 0.76 | 0.88 | 0.45 | 0.80 | 0.60 | 0.84 | 0.51 | 0.70 | 0.76 |
| 10 | Top Ranked Features Merged from Various Feature Selection Methods | Oversampling Using SMOTE | BML | 0.82 | 0.84 | 0.65 | 0.95 | 0.33 | 0.89 | 0.43 | 0.64 | 0.75 |
| | | | CatBoost | 0.81 | 0.84 | 0.61 | 0.94 | 0.34 | 0.89 | 0.43 | 0.64 | 0.75 |
| | | | ExtraTreesEntr | 0.80 | 0.83 | 0.56 | 0.93 | 0.31 | 0.88 | 0.40 | 0.62 | 0.74 |
| | | | ExtraTreesGini | 0.80 | 0.83 | 0.56 | 0.94 | 0.30 | 0.88 | 0.39 | 0.62 | 0.75 |
| | | | LightGBM | 0.81 | 0.84 | 0.60 | 0.94 | 0.33 | 0.89 | 0.43 | 0.64 | 0.75 |
| | | | LightGBMLarge | 0.81 | 0.84 | 0.60 | 0.95 | 0.31 | 0.89 | 0.41 | 0.63 | 0.74 |
| | | | LightGBMXT | 0.81 | 0.84 | 0.59 | 0.94 | 0.32 | 0.89 | 0.42 | 0.63 | 0.74 |
| | | | NeuralNetFastAI | 0.80 | 0.85 | 0.53 | 0.90 | 0.41 | 0.88 | 0.46 | 0.66 | 0.76 |
| | | | NeuralNetTorch | 0.82 | 0.84 | 0.64 | 0.95 | 0.31 | 0.89 | 0.42 | 0.63 | 0.75 |
| | | | RandomForestEntr | 0.80 | 0.83 | 0.57 | 0.94 | 0.30 | 0.88 | 0.40 | 0.62 | 0.75 |
| | | | RandomForestGini | 0.80 | 0.83 | 0.57 | 0.94 | 0.29 | 0.88 | 0.38 | 0.62 | 0.74 |
| | | | XGBoost | 0.82 | 0.84 | 0.62 | 0.94 | 0.34 | 0.89 | 0.44 | 0.64 | 0.74 |

***Supplemental Table 4: Summary of Model Results for Different Training Datasets***

| Training Dataset Number | Feature Selection Technique | Data Balancing Technique | Model | Accuarcy | LS Precision | HS Precision | LS Recall | HS Recall | LS F1 Score | HS F1 Score | Balanced Accuracy | ROC AUC |
|---|---|---|---|---|---|---|---|---|---|---|---|---|
| 11 | Top Ranked Features Merged from Various Feature Selection Methods | Oversampling Using Kmeans-SMOTE | BML | 0.82 | 0.84 | 0.66 | 0.96 | 0.33 | 0.89 | 0.44 | 0.64 | 0.75 |
| | | | CatBoost | 0.82 | 0.84 | 0.62 | 0.95 | 0.32 | 0.89 | 0.43 | 0.64 | 0.75 |
| | | | ExtraTreesEntr | 0.80 | 0.82 | 0.56 | 0.95 | 0.23 | 0.88 | 0.33 | 0.59 | 0.74 |
| | | | ExtraTreesGini | 0.80 | 0.82 | 0.57 | 0.95 | 0.23 | 0.88 | 0.33 | 0.59 | 0.74 |
| | | | LightGBM | 0.81 | 0.84 | 0.62 | 0.95 | 0.33 | 0.89 | 0.43 | 0.64 | 0.75 |
| | | | LightGBMLarge | 0.81 | 0.83 | 0.65 | 0.96 | 0.27 | 0.89 | 0.38 | 0.62 | 0.74 |
| | | | LightGBMXT | 0.81 | 0.83 | 0.61 | 0.95 | 0.30 | 0.89 | 0.40 | 0.62 | 0.74 |
| | | | NeuralNetFastAI | 0.80 | 0.85 | 0.54 | 0.91 | 0.39 | 0.88 | 0.45 | 0.65 | 0.75 |
| | | | NeuralNetTorch | 0.82 | 0.84 | 0.65 | 0.96 | 0.30 | 0.89 | 0.42 | 0.63 | 0.74 |
| | | | RandomForestEntr | 0.81 | 0.82 | 0.63 | 0.96 | 0.24 | 0.89 | 0.34 | 0.60 | 0.75 |
| | | | RandomForestGini | 0.81 | 0.82 | 0.60 | 0.96 | 0.24 | 0.89 | 0.34 | 0.60 | 0.74 |
| | | | XGBoost | 0.81 | 0.84 | 0.59 | 0.94 | 0.34 | 0.89 | 0.43 | 0.64 | 0.75 |
| 12 | Top Ranked Features Merged from Various Feature Selection Methods | Oversampling Using ADASYN | BML | 0.82 | 0.84 | 0.67 | 0.96 | 0.33 | 0.89 | 0.44 | 0.64 | 0.76 |
| | | | CatBoost | 0.81 | 0.84 | 0.61 | 0.94 | 0.33 | 0.89 | 0.43 | 0.64 | 0.75 |
| | | | ExtraTreesEntr | 0.81 | 0.84 | 0.58 | 0.93 | 0.33 | 0.88 | 0.42 | 0.63 | 0.75 |
| | | | ExtraTreesGini | 0.79 | 0.83 | 0.52 | 0.93 | 0.30 | 0.88 | 0.38 | 0.62 | 0.74 |
| | | | LightGBM | 0.81 | 0.84 | 0.61 | 0.94 | 0.32 | 0.89 | 0.42 | 0.63 | 0.75 |
| | | | LightGBMLarge | 0.82 | 0.84 | 0.63 | 0.95 | 0.30 | 0.89 | 0.41 | 0.62 | 0.74 |
| | | | LightGBMXT | 0.81 | 0.84 | 0.57 | 0.94 | 0.32 | 0.88 | 0.41 | 0.63 | 0.74 |
| | | | NeuralNetFastAI | 0.79 | 0.86 | 0.50 | 0.88 | 0.45 | 0.87 | 0.47 | 0.66 | 0.76 |
| | | | NeuralNetTorch | 0.82 | 0.84 | 0.66 | 0.96 | 0.32 | 0.89 | 0.43 | 0.64 | 0.74 |
| | | | RandomForestEntr | 0.80 | 0.84 | 0.57 | 0.93 | 0.32 | 0.88 | 0.41 | 0.62 | 0.75 |
| | | | RandomForestGini | 0.80 | 0.83 | 0.55 | 0.93 | 0.30 | 0.88 | 0.39 | 0.62 | 0.74 |
| | | | XGBoost | 0.81 | 0.84 | 0.58 | 0.94 | 0.33 | 0.88 | 0.42 | 0.64 | 0.74 |

***Supplemental Table 4: Summary of Model Results for Different Training Datasets***

| Training Dataset Number | Feature Selection Technique | Data Balancing Technique | Model | Accuarcy | LS Precision | HS Precision | LS Recall | HS Recall | LS F1 Score | HS F1 Score | Balanced Accuracy | ROC AUC |
|---|---|---|---|---|---|---|---|---|---|---|---|---|
| 13 | Top Ranked Features Merged from Various Feature Selection Methods | Oversampling Using Random-OverSampler | BML | 0.71 | 0.88 | 0.39 | 0.74 | 0.64 | 0.80 | 0.49 | 0.69 | 0.75 |
| | | | CatBoost | 0.77 | 0.85 | 0.45 | 0.86 | 0.43 | 0.85 | 0.44 | 0.64 | 0.70 |
| | | | ExtraTreesEntr | 0.78 | 0.83 | 0.46 | 0.90 | 0.30 | 0.86 | 0.36 | 0.60 | 0.71 |
| | | | ExtraTreesGini | 0.78 | 0.83 | 0.47 | 0.91 | 0.30 | 0.87 | 0.37 | 0.60 | 0.71 |
| | | | LightGBM | 0.74 | 0.85 | 0.41 | 0.82 | 0.45 | 0.83 | 0.43 | 0.64 | 0.70 |
| | | | LightGBMLarge | 0.78 | 0.84 | 0.47 | 0.89 | 0.35 | 0.86 | 0.40 | 0.62 | 0.70 |
| | | | LightGBMXT | 0.76 | 0.84 | 0.43 | 0.87 | 0.37 | 0.85 | 0.40 | 0.62 | 0.68 |
| | | | NeuralNetFastAI | 0.77 | 0.84 | 0.45 | 0.88 | 0.37 | 0.86 | 0.41 | 0.62 | 0.70 |
| | | | NeuralNetTorch | 0.77 | 0.83 | 0.44 | 0.88 | 0.35 | 0.86 | 0.39 | 0.62 | 0.70 |
| | | | RandomForestEntr | 0.78 | 0.83 | 0.48 | 0.91 | 0.31 | 0.87 | 0.38 | 0.61 | 0.71 |
| | | | RandomForestGini | 0.78 | 0.83 | 0.48 | 0.91 | 0.31 | 0.87 | 0.38 | 0.61 | 0.71 |
| | | | XGBoost | 0.76 | 0.84 | 0.43 | 0.86 | 0.39 | 0.85 | 0.41 | 0.62 | 0.69 |
| 14 | Top Ranked Features Merged from Various Feature Selection Methods | Oversampling Using Kernel-based Synthetic Minority Oversampling (K-SMOTE) | BML | 0.69 | 0.88 | 0.37 | 0.71 | 0.64 | 0.78 | 0.47 | 0.67 | 0.74 |
| | | | CatBoost | 0.79 | 0.79 | 1.00 | 1.00 | 0.01 | 0.88 | 0.02 | 0.50 | 0.54 |
| | | | ExtraTreesEntr | 0.80 | 0.80 | 1.00 | 1.00 | 0.06 | 0.89 | 0.12 | 0.53 | 0.70 |
| | | | ExtraTreesGini | 0.80 | 0.80 | 0.96 | 1.00 | 0.07 | 0.89 | 0.13 | 0.54 | 0.71 |
| | | | LightGBM | 0.79 | 0.79 | 0.92 | 1.00 | 0.04 | 0.88 | 0.07 | 0.52 | 0.54 |
| | | | LightGBMLarge | 0.79 | 0.79 | 1.00 | 1.00 | 0.01 | 0.88 | 0.02 | 0.50 | 0.51 |
| | | | LightGBMXT | 0.79 | 0.79 | 1.00 | 1.00 | 0.01 | 0.88 | 0.03 | 0.50 | 0.66 |
| | | | NeuralNetFastAI | 0.81 | 0.83 | 0.62 | 0.95 | 0.30 | 0.89 | 0.41 | 0.62 | 0.74 |
| | | | NeuralNetTorch | 0.80 | 0.80 | 0.88 | 1.00 | 0.09 | 0.89 | 0.16 | 0.55 | 0.71 |
| | | | RandomForestEntr | 0.79 | 0.79 | 1.00 | 1.00 | 0.01 | 0.88 | 0.01 | 0.50 | 0.59 |
| | | | RandomForestGini | 0.79 | 0.79 | 1.00 | 1.00 | 0.01 | 0.88 | 0.01 | 0.50 | 0.63 |
| | | | XGBoost | 0.79 | 0.79 | 1.00 | 1.00 | 0.01 | 0.88 | 0.03 | 0.50 | 0.54 |

***Supplemental Table 4: Summary of Model Results for Different Training Datasets***

| Training Dataset Number | Feature Selection Technique | Data Balancing Technique | Model | Accuarcy | LS Precision | HS Precision | LS Recall | HS Recall | LS F1 Score | HS F1 Score | Balanced Accuracy | ROC AUC |
|---|---|---|---|---|---|---|---|---|---|---|---|---|
| 15 | Top Ranked Features Merged from Various Feature Selection Methods | Oversampling Using WGAN-GP | BML | 0.82 | 0.83 | 0.67 | 0.97 | 0.25 | 0.89 | 0.36 | 0.61 | 0.76 |
| | | | CatBoost | 0.82 | 0.82 | 0.76 | 0.98 | 0.19 | 0.89 | 0.30 | 0.58 | 0.76 |
| | | | ExtraTreesEntr | 0.81 | 0.82 | 0.73 | 0.98 | 0.18 | 0.89 | 0.28 | 0.58 | 0.75 |
| | | | ExtraTreesGini | 0.80 | 0.82 | 0.60 | 0.96 | 0.20 | 0.88 | 0.30 | 0.58 | 0.73 |
| | | | LightGBM | 0.81 | 0.82 | 0.71 | 0.98 | 0.21 | 0.89 | 0.33 | 0.60 | 0.75 |
| | | | LightGBMLarge | 0.80 | 0.82 | 0.60 | 0.96 | 0.22 | 0.88 | 0.32 | 0.59 | 0.73 |
| | | | LightGBMXT | 0.81 | 0.82 | 0.68 | 0.97 | 0.20 | 0.89 | 0.31 | 0.58 | 0.75 |
| | | | NeuralNetFastAI | 0.82 | 0.83 | 0.66 | 0.96 | 0.27 | 0.89 | 0.38 | 0.62 | 0.75 |
| | | | NeuralNetTorch | 0.81 | 0.82 | 0.78 | 0.99 | 0.17 | 0.89 | 0.28 | 0.58 | 0.75 |
| | | | RandomForestEntr | 0.80 | 0.82 | 0.60 | 0.96 | 0.20 | 0.88 | 0.30 | 0.58 | 0.73 |
| | | | RandomForestGini | 0.80 | 0.82 | 0.60 | 0.96 | 0.20 | 0.88 | 0.30 | 0.58 | 0.73 |
| | | | XGBoost | 0.81 | 0.82 | 0.65 | 0.97 | 0.20 | 0.89 | 0.31 | 0.58 | 0.75 |
| 16 | Top Ranked Features Merged from Various Feature Selection Methods | Oversampling Using Conditional WGAN-GP | BML | 0.68 | 0.84 | 0.33 | 0.72 | 0.50 | 0.78 | 0.39 | 0.61 | 0.66 |
| | | | CatBoost | 0.76 | 0.83 | 0.42 | 0.87 | 0.35 | 0.85 | 0.39 | 0.61 | 0.71 |
| | | | ExtraTreesEntr | 0.78 | 0.82 | 0.46 | 0.92 | 0.25 | 0.87 | 0.32 | 0.58 | 0.71 |
| | | | ExtraTreesGini | 0.73 | 0.82 | 0.36 | 0.84 | 0.34 | 0.83 | 0.35 | 0.59 | 0.68 |
| | | | LightGBM | 0.75 | 0.83 | 0.41 | 0.86 | 0.36 | 0.85 | 0.38 | 0.61 | 0.69 |
| | | | LightGBMLarge | 0.76 | 0.83 | 0.43 | 0.88 | 0.32 | 0.86 | 0.37 | 0.60 | 0.70 |
| | | | LightGBMXT | 0.75 | 0.83 | 0.39 | 0.86 | 0.33 | 0.84 | 0.36 | 0.60 | 0.69 |
| | | | NeuralNetFastAI | 0.71 | 0.85 | 0.37 | 0.77 | 0.51 | 0.81 | 0.43 | 0.64 | 0.70 |
| | | | NeuralNetTorch | 0.76 | 0.84 | 0.43 | 0.86 | 0.39 | 0.85 | 0.41 | 0.62 | 0.71 |
| | | | RandomForestEntr | 0.75 | 0.82 | 0.38 | 0.87 | 0.30 | 0.84 | 0.34 | 0.58 | 0.68 |
| | | | RandomForestGini | 0.78 | 0.82 | 0.47 | 0.92 | 0.27 | 0.87 | 0.34 | 0.60 | 0.72 |
| | | | XGBoost | 0.78 | 0.83 | 0.48 | 0.91 | 0.30 | 0.87 | 0.37 | 0.60 | 0.72 |

***Supplemental Table 4: Summary of Model Results for Different Training Datasets***

| Training Dataset Number | Feature Selection Technique | Data Balancing Technique | Model | Accuarcy | LS Precision | HS Precision | LS Recall | HS Recall | LS F1 Score | HS F1 Score | Balanced Accuracy | ROC AUC |
|---|---|---|---|---|---|---|---|---|---|---|---|---|
| 17 | Top Ranked Features Merged from Various Feature Selection Methods | Undersampling Using NearMiss-1 | BML | 0.56 | 0.87 | 0.28 | 0.52 | 0.71 | 0.65 | 0.40 | 0.61 | 0.67 |
| | | | CatBoost | 0.59 | 0.87 | 0.30 | 0.56 | 0.69 | 0.68 | 0.42 | 0.62 | 0.68 |
| | | | ExtraTreesEntr | 0.59 | 0.86 | 0.29 | 0.56 | 0.67 | 0.68 | 0.41 | 0.62 | 0.67 |
| | | | ExtraTreesGini | 0.59 | 0.87 | 0.30 | 0.57 | 0.67 | 0.69 | 0.41 | 0.62 | 0.67 |
| | | | LightGBM | 0.59 | 0.87 | 0.30 | 0.56 | 0.70 | 0.68 | 0.42 | 0.63 | 0.67 |
| | | | LightGBMLarge | 0.56 | 0.87 | 0.29 | 0.53 | 0.70 | 0.66 | 0.40 | 0.62 | 0.67 |
| | | | LightGBMXT | 0.58 | 0.87 | 0.29 | 0.55 | 0.70 | 0.67 | 0.41 | 0.62 | 0.67 |
| | | | NeuralNetFastAI | 0.56 | 0.88 | 0.29 | 0.51 | 0.75 | 0.65 | 0.42 | 0.63 | 0.69 |
| | | | NeuralNetTorch | 0.61 | 0.87 | 0.31 | 0.59 | 0.68 | 0.70 | 0.42 | 0.64 | 0.69 |
| | | | RandomForestEntr | 0.57 | 0.87 | 0.29 | 0.55 | 0.69 | 0.67 | 0.41 | 0.62 | 0.67 |
| | | | RandomForestGini | 0.57 | 0.87 | 0.29 | 0.54 | 0.69 | 0.67 | 0.41 | 0.62 | 0.67 |
| | | | XGBoost | 0.61 | 0.87 | 0.31 | 0.60 | 0.65 | 0.71 | 0.42 | 0.62 | 0.68 |
| 18 | Top Ranked Features Merged from Various Feature Selection Methods | Undersampling Using Random-UnderSampler | BML | 0.70 | 0.88 | 0.38 | 0.72 | 0.64 | 0.79 | 0.48 | 0.68 | 0.74 |
| | | | CatBoost | 0.71 | 0.88 | 0.39 | 0.72 | 0.65 | 0.80 | 0.48 | 0.68 | 0.74 |
| | | | ExtraTreesEntr | 0.68 | 0.87 | 0.35 | 0.70 | 0.61 | 0.77 | 0.45 | 0.66 | 0.71 |
| | | | ExtraTreesGini | 0.68 | 0.87 | 0.35 | 0.70 | 0.61 | 0.77 | 0.45 | 0.66 | 0.71 |
| | | | LightGBM | 0.70 | 0.88 | 0.38 | 0.71 | 0.65 | 0.79 | 0.48 | 0.68 | 0.73 |
| | | | LightGBMLarge | 0.67 | 0.87 | 0.35 | 0.69 | 0.60 | 0.77 | 0.44 | 0.64 | 0.71 |
| | | | LightGBMXT | 0.69 | 0.88 | 0.37 | 0.71 | 0.65 | 0.78 | 0.47 | 0.68 | 0.72 |
| | | | NeuralNetFastAI | 0.70 | 0.89 | 0.38 | 0.71 | 0.67 | 0.79 | 0.49 | 0.69 | 0.75 |
| | | | NeuralNetTorch | 0.71 | 0.89 | 0.39 | 0.71 | 0.69 | 0.79 | 0.50 | 0.70 | 0.76 |
| | | | RandomForestEntr | 0.68 | 0.87 | 0.35 | 0.70 | 0.61 | 0.78 | 0.45 | 0.66 | 0.71 |
| | | | RandomForestGini | 0.67 | 0.87 | 0.35 | 0.69 | 0.62 | 0.77 | 0.44 | 0.66 | 0.71 |
| | | | XGBoost | 0.71 | 0.88 | 0.39 | 0.73 | 0.63 | 0.80 | 0.48 | 0.68 | 0.73 |

## Crash Severity Risk Modeling Strategies under Data Imbalance

*Supplemental Table 4: Summary of Model Results for Different Training Datasets*

| Training Dataset Number | Feature Selection Technique | Data Balancing Technique | Model | Accuarcy | LS Precision | HS Precision | LS Recall | HS Recall | LS F1 Score | HS F1 Score | Balanced Accuracy | ROC AUC |
|---|---|---|---|---|---|---|---|---|---|---|---|---|
| 19 | Top Ranked Features Merged from Various Feature Selection Methods | Combination of Random-OverSampler and Random-UnderSampler | BML | 0.71 | 0.87 | 0.39 | 0.74 | 0.60 | 0.80 | 0.47 | 0.67 | 0.74 |
| | | | CatBoost | 0.73 | 0.85 | 0.39 | 0.79 | 0.50 | 0.82 | 0.44 | 0.64 | 0.70 |
| | | | ExtraTreesEntr | 0.75 | 0.85 | 0.41 | 0.83 | 0.44 | 0.84 | 0.42 | 0.64 | 0.71 |
| | | | ExtraTreesGini | 0.74 | 0.84 | 0.40 | 0.82 | 0.43 | 0.83 | 0.41 | 0.62 | 0.70 |
| | | | LightGBM | 0.72 | 0.86 | 0.38 | 0.77 | 0.54 | 0.81 | 0.45 | 0.66 | 0.71 |
| | | | LightGBMLarge | 0.74 | 0.85 | 0.40 | 0.82 | 0.44 | 0.83 | 0.42 | 0.63 | 0.70 |
| | | | LightGBMXT | 0.72 | 0.84 | 0.37 | 0.79 | 0.46 | 0.82 | 0.41 | 0.62 | 0.68 |
| | | | NeuralNetFastAI | 0.74 | 0.85 | 0.41 | 0.81 | 0.49 | 0.83 | 0.44 | 0.65 | 0.72 |
| | | | NeuralNetTorch | 0.75 | 0.85 | 0.42 | 0.84 | 0.44 | 0.84 | 0.43 | 0.64 | 0.71 |
| | | | RandomForestEntr | 0.75 | 0.85 | 0.42 | 0.83 | 0.45 | 0.84 | 0.43 | 0.64 | 0.70 |
| | | | RandomForestGini | 0.74 | 0.84 | 0.40 | 0.83 | 0.43 | 0.83 | 0.41 | 0.63 | 0.71 |
| | | | XGBoost | 0.73 | 0.85 | 0.39 | 0.80 | 0.46 | 0.83 | 0.42 | 0.63 | 0.69 |
| 20 | Discriminative Mutual Information | High Severity Weight = 4 | BML | 0.71 | 0.88 | 0.38 | 0.73 | 0.63 | 0.80 | 0.48 | 0.68 | 0.75 |
| | | | CatBoost | 0.72 | 0.88 | 0.40 | 0.75 | 0.62 | 0.81 | 0.49 | 0.68 | 0.76 |
| | | | ExtraTreesEntr | 0.73 | 0.84 | 0.38 | 0.82 | 0.40 | 0.83 | 0.39 | 0.61 | 0.68 |
| | | | ExtraTreesGini | 0.73 | 0.83 | 0.37 | 0.82 | 0.40 | 0.83 | 0.38 | 0.61 | 0.68 |
| | | | LightGBM | 0.72 | 0.88 | 0.39 | 0.74 | 0.61 | 0.81 | 0.48 | 0.68 | 0.75 |
| | | | LightGBMLarge | 0.72 | 0.87 | 0.39 | 0.75 | 0.58 | 0.81 | 0.47 | 0.66 | 0.72 |
| | | | LightGBMXT | 0.72 | 0.87 | 0.39 | 0.75 | 0.60 | 0.81 | 0.47 | 0.68 | 0.74 |
| | | | NeuralNetFastAI | 0.78 | 0.85 | 0.49 | 0.88 | 0.42 | 0.87 | 0.45 | 0.65 | 0.75 |
| | | | NeuralNetTorch | 0.79 | 0.85 | 0.50 | 0.88 | 0.44 | 0.87 | 0.47 | 0.66 | 0.76 |
| | | | RandomForestEntr | 0.73 | 0.83 | 0.37 | 0.82 | 0.40 | 0.82 | 0.38 | 0.61 | 0.68 |
| | | | RandomForestGini | 0.73 | 0.83 | 0.37 | 0.82 | 0.40 | 0.82 | 0.38 | 0.61 | 0.68 |
| | | | XGBoost | 0.74 | 0.87 | 0.42 | 0.79 | 0.57 | 0.83 | 0.49 | 0.68 | 0.75 |

## Crash Severity Risk Modeling Strategies under Data Imbalance

*Supplemental Table 4: Summary of Model Results for Different Training Datasets*

| Training Dataset Number | Feature Selection Technique | Data Balancing Technique | Model | Accuarcy | LS Precision | HS Precision | LS Recall | HS Recall | LS F1 Score | HS F1 Score | Balanced Accuracy | ROC AUC |
|---|---|---|---|---|---|---|---|---|---|---|---|---|
| 21 | Discriminative Mutual Information | Oversampling Using Random-OverSampler | BML | 0.71 | 0.87 | 0.39 | 0.75 | 0.60 | 0.80 | 0.47 | 0.67 | 0.74 |
| | | | CatBoost | 0.73 | 0.86 | 0.40 | 0.79 | 0.52 | 0.82 | 0.45 | 0.66 | 0.73 |
| | | | ExtraTreesEntr | 0.72 | 0.84 | 0.36 | 0.79 | 0.44 | 0.82 | 0.40 | 0.62 | 0.68 |
| | | | ExtraTreesGini | 0.72 | 0.84 | 0.37 | 0.80 | 0.44 | 0.82 | 0.40 | 0.62 | 0.67 |
| | | | LightGBM | 0.73 | 0.87 | 0.40 | 0.78 | 0.56 | 0.82 | 0.47 | 0.67 | 0.75 |
| | | | LightGBMLarge | 0.72 | 0.85 | 0.38 | 0.79 | 0.48 | 0.82 | 0.42 | 0.64 | 0.70 |
| | | | LightGBMXT | 0.72 | 0.85 | 0.37 | 0.78 | 0.47 | 0.81 | 0.41 | 0.62 | 0.68 |
| | | | NeuralNetFastAI | 0.74 | 0.86 | 0.41 | 0.80 | 0.52 | 0.83 | 0.46 | 0.66 | 0.74 |
| | | | NeuralNetTorch | 0.73 | 0.85 | 0.39 | 0.80 | 0.48 | 0.83 | 0.43 | 0.64 | 0.70 |
| | | | RandomForestEntr | 0.72 | 0.84 | 0.36 | 0.79 | 0.44 | 0.82 | 0.40 | 0.62 | 0.68 |
| | | | RandomForestGini | 0.72 | 0.84 | 0.37 | 0.80 | 0.43 | 0.82 | 0.40 | 0.62 | 0.67 |
| | | | XGBoost | 0.74 | 0.86 | 0.41 | 0.79 | 0.53 | 0.83 | 0.46 | 0.66 | 0.74 |
| 22 | Discriminative Mutual Information | Undersampling Using NearMiss-1 | BML | 0.42 | 0.83 | 0.23 | 0.33 | 0.75 | 0.48 | 0.36 | 0.54 | 0.60 |
| | | | CatBoost | 0.41 | 0.83 | 0.23 | 0.31 | 0.76 | 0.46 | 0.35 | 0.54 | 0.58 |
| | | | ExtraTreesEntr | 0.36 | 0.81 | 0.22 | 0.26 | 0.77 | 0.39 | 0.34 | 0.52 | 0.56 |
| | | | ExtraTreesGini | 0.37 | 0.81 | 0.22 | 0.26 | 0.77 | 0.39 | 0.34 | 0.52 | 0.57 |
| | | | LightGBM | 0.44 | 0.84 | 0.24 | 0.35 | 0.75 | 0.49 | 0.36 | 0.55 | 0.60 |
| | | | LightGBMLarge | 0.37 | 0.81 | 0.22 | 0.26 | 0.78 | 0.39 | 0.34 | 0.52 | 0.56 |
| | | | LightGBMXT | 0.39 | 0.83 | 0.23 | 0.28 | 0.79 | 0.41 | 0.35 | 0.54 | 0.59 |
| | | | NeuralNetFastAI | 0.47 | 0.84 | 0.24 | 0.41 | 0.70 | 0.55 | 0.36 | 0.55 | 0.59 |
| | | | NeuralNetTorch | 0.44 | 0.83 | 0.23 | 0.37 | 0.72 | 0.51 | 0.35 | 0.54 | 0.59 |
| | | | RandomForestEntr | 0.37 | 0.81 | 0.22 | 0.26 | 0.77 | 0.39 | 0.34 | 0.52 | 0.57 |
| | | | RandomForestGini | 0.37 | 0.80 | 0.22 | 0.26 | 0.77 | 0.39 | 0.34 | 0.52 | 0.57 |
| | | | XGBoost | 0.46 | 0.84 | 0.24 | 0.38 | 0.72 | 0.53 | 0.36 | 0.55 | 0.61 |

***Supplemental Table 4: Summary of Model Results for Different Training Datasets***

| Training Dataset Number | Feature Selection Technique | Data Balancing Technique | Model | Accuarcy | LS Precision | HS Precision | LS Recall | HS Recall | LS F1 Score | HS F1 Score | Balanced Accuracy | ROC AUC |
|---|---|---|---|---|---|---|---|---|---|---|---|---|
| 23 | Discriminative Mutual Information | Undersampling Using Random-UnderSampler | BML | 0.76 | 0.86 | 0.44 | 0.83 | 0.50 | 0.84 | 0.47 | 0.66 | 0.74 |
| | | | CatBoost | 0.77 | 0.86 | 0.46 | 0.85 | 0.47 | 0.85 | 0.46 | 0.66 | 0.75 |
| | | | ExtraTreesEntr | 0.73 | 0.85 | 0.38 | 0.80 | 0.46 | 0.82 | 0.42 | 0.63 | 0.69 |
| | | | ExtraTreesGini | 0.73 | 0.85 | 0.38 | 0.80 | 0.47 | 0.82 | 0.42 | 0.64 | 0.69 |
| | | | LightGBM | 0.76 | 0.86 | 0.45 | 0.84 | 0.48 | 0.85 | 0.46 | 0.66 | 0.74 |
| | | | LightGBMLarge | 0.74 | 0.85 | 0.41 | 0.82 | 0.46 | 0.83 | 0.43 | 0.64 | 0.71 |
| | | | LightGBMXT | 0.75 | 0.85 | 0.42 | 0.84 | 0.45 | 0.84 | 0.44 | 0.64 | 0.74 |
| | | | NeuralNetFastAI | 0.74 | 0.87 | 0.42 | 0.79 | 0.57 | 0.83 | 0.48 | 0.68 | 0.75 |
| | | | NeuralNetTorch | 0.76 | 0.87 | 0.44 | 0.82 | 0.53 | 0.84 | 0.48 | 0.68 | 0.75 |
| | | | RandomForestEntr | 0.73 | 0.85 | 0.38 | 0.80 | 0.46 | 0.82 | 0.42 | 0.63 | 0.69 |
| | | | RandomForestGini | 0.73 | 0.85 | 0.39 | 0.80 | 0.47 | 0.82 | 0.42 | 0.64 | 0.69 |
| | | | XGBoost | 0.77 | 0.86 | 0.45 | 0.85 | 0.47 | 0.85 | 0.46 | 0.66 | 0.75 |
| 24 | Discriminative Mutual Information | Combination of Random-OverSampler and Random-UnderSampler | BML | 0.73 | 0.87 | 0.41 | 0.78 | 0.57 | 0.82 | 0.48 | 0.68 | 0.75 |
| | | | CatBoost | 0.74 | 0.86 | 0.41 | 0.80 | 0.52 | 0.83 | 0.46 | 0.66 | 0.72 |
| | | | ExtraTreesEntr | 0.69 | 0.84 | 0.34 | 0.74 | 0.49 | 0.79 | 0.40 | 0.62 | 0.67 |
| | | | ExtraTreesGini | 0.69 | 0.84 | 0.34 | 0.74 | 0.49 | 0.79 | 0.40 | 0.62 | 0.67 |
| | | | LightGBM | 0.74 | 0.86 | 0.41 | 0.80 | 0.52 | 0.83 | 0.46 | 0.66 | 0.73 |
| | | | LightGBMLarge | 0.72 | 0.85 | 0.37 | 0.78 | 0.48 | 0.81 | 0.42 | 0.63 | 0.69 |
| | | | LightGBMXT | 0.70 | 0.84 | 0.35 | 0.76 | 0.48 | 0.80 | 0.41 | 0.62 | 0.68 |
| | | | NeuralNetFastAI | 0.72 | 0.86 | 0.38 | 0.76 | 0.55 | 0.81 | 0.45 | 0.66 | 0.72 |
| | | | NeuralNetTorch | 0.71 | 0.86 | 0.37 | 0.75 | 0.54 | 0.80 | 0.44 | 0.64 | 0.70 |
| | | | RandomForestEntr | 0.69 | 0.84 | 0.33 | 0.74 | 0.47 | 0.79 | 0.39 | 0.60 | 0.67 |
| | | | RandomForestGini | 0.69 | 0.84 | 0.34 | 0.74 | 0.49 | 0.79 | 0.40 | 0.62 | 0.67 |
| | | | XGBoost | 0.73 | 0.85 | 0.40 | 0.80 | 0.50 | 0.83 | 0.44 | 0.65 | 0.71 |